\documentclass[3p,sort & compress,final]{elsarticle}
\usepackage{amssymb}
\usepackage{latexsym}
\usepackage{amsfonts}
\usepackage{amsthm}
\usepackage{amsbsy}
\usepackage{amsmath}%
\usepackage{graphics,graphicx}
\usepackage{subfigure}%
\usepackage{multirow,multicol}
\usepackage{bm}
\usepackage{bbm}
\usepackage{color}
\usepackage{float}
\usepackage{calc}
\usepackage{caption}
\usepackage{dsfont}
\usepackage{ifthen}
\usepackage{mathrsfs}
\usepackage[colorlinks,linkcolor=blue,citecolor=blue]{hyperref}
\usepackage{epstopdf}
\usepackage{epsfig}
\usepackage{algorithm}
\usepackage{algorithmic}
\usepackage{pifont}
\usepackage{natbib}
\usepackage{overpic}
\usepackage{psfrag}
\usepackage{rotating}
\usepackage{stmaryrd}
\usepackage{verbatim}
\usepackage{setspace}
\usepackage{tikz}
\usepackage{appendix}

\newtheorem*{remark}{Remark}

\newdefinition{example}{Example}

\numberwithin{equation}{section}

\numberwithin{theorem}{section}
\journal{}

\begin{document}
	\begin{frontmatter}
		
		\title{Subspace Decomposition based DNN algorithm for elliptic type multi-scale PDEs }
		
		\author[shjtmath]{Xi-An Li} \ead{lixian9131@163.com,lixa0415@sjtu.edu.cn}
		\author[shjtmath,shjtnature,moelsc]{Zhi-Qin John Xu}\ead{xuzhiqin@sjtu.edu.cn} 
		\author[shjtmath,shjtnature,moelsc]{Lei Zhang\corref{lz}} \ead{Lzhang2012@sjtu.edu.cn}
		\cortext[lz]{Corresponding author.}
		
		\address[shjtmath]{School of Mathematical Sciences, Shanghai Jiao Tong University, Shanghai 200240, China}
		\address[shjtnature]{Institute of Natural Sciences, Shanghai Jiao Tong University, Shanghai 200240, China}
		\address[moelsc]{MOE-LSC, Shanghai Jiao Tong University, Shanghai 200240, China}

		\begin{abstract}
			While deep learning algorithms demonstrate a great potential in scientific computing, its application to multi-scale problems remains to be a big challenge. This is manifested by the ``frequency principle"  that neural networks tend to learn  low frequency components first. Novel architectures such as multi-scale deep neural network (MscaleDNN) were proposed to alleviate this problem to some extent. In this paper, we construct a subspace decomposition based DNN (dubbed SD$^2$NN) architecture for a class of multi-scale problems by combining traditional numerical analysis ideas and MscaleDNN algorithms. The proposed architecture includes one low frequency normal DNN submodule, and one (or a few) high frequency MscaleDNN submodule(s), which are designed to capture the smooth part and the oscillatory part of the multi-scale solutions, respectively. In addition, a novel trigonometric activation function is incorporated in the SD$^2$NN model.  We demonstrate the performance of  the SD$^2$NN architecture through several benchmark multi-scale problems in regular or irregular geometric domains. Numerical results show that the SD$^2$NN model is superior to existing models such as MscaleDNN. \\
			
		\end{abstract}
	\end{frontmatter}
	
	\section{Introduction}\label{sec:01}
	Machine learning algorithms, especially deep neural networks (DNNs), have not only achieved great success in traditional artificial intelligence tasks such as image recognition, natural language processing, and recommendation systems \cite{lecun2015deep,goodfellow2016deep}, but also attracted more and more attention in the field of scientific computation including the numerical solution of ordinary/partial differential equations, integral-differential equations and dynamical systems \cite{e2017a,weinan2017deep,han2018deep,raissi2019physics,weinan2020machine,shi2021comparative},
	with the fast development of novel computing devices, as well as the rapidly increasing volume and complexity of data. In this paper, we will introduce a new DNN-based algorithm to solve the following multi-scale equation,
	\begin{equation}\label{eq:nonlinear-ms}
	\left\{
	\begin{array}{cc}
	\mathcal{L}u(\bm{x})  = f(\bm{x}), &  \bm{x}\in \Omega,\\
	\mathcal{B}u(\bm{x})  = g(\bm{x}), & \bm{x}\in \partial\Omega,
	\end{array}
	\right.
	\end{equation}
	where $\Omega$ is a bounded subset of $\mathbb{R}^{d}$ with piecewise Lipschitz boundary which satisfies the interior cone condition. $\mathcal{L}$ is a linear or non-linear elliptic type differential operator on $\Omega$ which contains possibly non-separable multiple scales, and $\mathcal{B}$ is a boundary operator on $\partial\Omega$. We assume that $\mathcal{L}$ is uniformly strongly elliptic such that $\int_\Omega v' \mathcal{L} v \geq c\|v\|_{\mathcal{V}}$ for some admissible space $\mathcal{V}$, with a constant $c>0$. 
	
	The multi-scale equation \eqref{eq:nonlinear-ms} has many physical and engineering applications, such as heat conduction in composite materials, reservoir modeling in porous media, convection dominated flows and the Poisson-Boltzmann model for dielectric systems \cite{Owhadi2003,homeyer2012free,hagelaar2005solving,cohen2012adaptivity},  There has been amount of work concerning the design of multi-scale numerical methods to achieve the optimal balance of accuracy and complexity, such as homogenization \cite{papanicolau1978asymptotic,jikov2012homogenization}, heterogeneous multi-scale methods (HMM) \cite{ee03, ming2005analysis}, multi-scale network approximations \cite{berlyand2013introduction}, multi-scale finite element methods (MsFEM)\cite{Hou97}, variational multi-scale methods (VMS) \cite{hughes98, bazilevs2007variational}, flux norm homogenization \cite{berlyand2010flux,owhadi2008homogenization}, rough polyharmonic splines (RPS) \cite{OwhZhaBer:2014, NMTMA-14-862}, generalized multi-scale finite element methods (GMsFEM) \cite{egh12, chung2014adaptiveDG}, localized orthogonal decomposition (LOD) \cite{MalPet:2014,Henning2014}, etc. A common theme of the multi-scale methods is to construct coarse approximation spaces with optimal error control through the identification of low dimensional structures in the high dimensional multi-scale solution space. The solution space can be decomposed into a direct sum of a coarse space (with ``smooth'' components) and (possibly a few) fine spaces (with ``oscillatory'' components), which is a natural extension of the Fourier decomposition. The idea of subspace decomposition has been widely used in numerical analysis and multi-scale modeling \cite{hughes98,MalPet:2014,xu:1992,xie2019fast}, furthermore, it has surprisingly deep connections with Bayesian inference, kernel learning and probabilistic numerics \cite{Owhadi2015,OwhadiMultigrid:2017,owhadi2020kernel,owhadi2019kernel}. 
	
	Comparing with conventional numerical algorithms, DNN approximation can overcome the curse of dimensionality, therefore it is ideal for high dimensional PDEs. On the other hand, it can be used as a meshless method which is suitable for PDEs in complex domains. Various DNN based algorithms have been proposed in \cite{weinan2017deep,sirignano2018dgm,raissi2019physics,zang2020weak} to solve PDEs. Despite the above mentioned progress, the frequency principle (F-Principle) \cite{xu_training_2018,xu2020frequency,rahaman2018spectral,zhang2021linear} shows that general DNN-based algorithms often encounter a ``curse of high-frequency" as they are inefficient to learn high-frequency information of multiscale functions. A series of subsequent theoretical investigations further confirm such empirical observation \cite{xu2020frequency,luo2019theory,ronen2019the,e2019machine,cao2019towards,yang2019fine,bordelon2020spectrum,luo2020exact}. The study of the F-Principle has also been utilized to understand various phenomena emerging in applications of deep learning \cite{ma2020slow,sharma2020d,zhu2019dspnet,chakrabarty2019spectral,xu2020deep}. Inspired by the F-Principle, a series of algorithms are developed to solve multi-scale PDEs and to overcome the high-frequency curse of general DNNs  \citep{wang2020eigenvector,cai2020phase,jagtap2019adaptive,liu2020multi,li2020elliptic,wang2020multi}. For example, in MscaleDNN, high frequency components are shifted into low frequency ones by radial scaling such that they can be learnt more efficiently. In addition, some smooth and localized activation functions were proposed for MscaleDNN algorithm \citep{liu2020multi,li2020elliptic}.
	
	In this paper, motivated by the subspace decomposition technique in numerical analysis, and the MscaleDNN framework \citep{liu2020multi,li2020elliptic}, we propose a subspace-decomposition based DNN (called SD$^2$NN) architecture to solve the multi-scale equation \eqref{eq:nonlinear-ms}. The SD$^2$NN framework consists of two parts: one low-frequency or normal DNN submodule and one (or a few) MscaleDNN submodule(s), to capture the low-frequency and high-frequency components of the multi-scale solution, respectively. We enforce the orthogonality between subspaces in the SD$^2$NN architecture to enforce the stability of the decomposition, by adding a penalty term to the loss function.
	Such a framework enables simultaneous fast learning of both the smooth part and the oscillatory part of multiscale functions/solutions, thus avoids the curse of frequency. We also improve the activation function by adding a relaxation factor and using trigonometric functions \cite{Matthew2020Fourier,wang2020eigenvector}, which will be elaborated in the later sections. 
	
	The paper is organized as follows. In Section \ref{sec:02}, we briefly introduce the subspace decomposition based DNN framework to approximate multi-scale functions. In Section \ref{sec:03}, we use the SD$^2$NN framework to solve multi-scale problems in the variational formulation, and discuss various options to choose activation functions. Benchmark numerical experiments are carried out in Section \ref{sec:04} to evaluate the performance of SD$^2$NN. We conclude the paper in Section \ref{sec:05}.

	\section{Formulation of the Subspace Decomposition based DNN}
	\label{sec:02}
	
	\subsection{Multi-scale DNN (MscaleDNN)}
	
	We briefly review the formulation of MscaleDNN introduced in \cite{liu2020multi,li2020elliptic}, for convenience of readers. A deep neural network defines a function mapping: 
	$\bm{x}\in\mathbb{R}^{d}\rightarrow y_{\bm{\theta}}(\bm{x})\in\mathbb{R}$,
	where $d$ is the dimension of input. In particular, the DNN function is a nested composition of linear functions and nonlinear activation functions, which is of the form
	\begin{equation}
	\begin{cases}
	\bm{y}_{\bm{\theta}}^{[0]} = \bm{x}\\
	\bm{y}_{\bm{\theta}}^{[\ell]} = \sigma\circ(\bm{W}^{[\ell]}\bm{y}_{\bm{\theta}}^{[\ell-1]}+\bm{b}^{[\ell]}), ~~\text{for}~~\ell =1, 2, 3, \cdots\cdots, L
	\end{cases}
	\end{equation}
	where $\bm{W}^{[\ell]} \in  \mathbb{R}^{m_{\ell+1}\times m_{\ell}}, \bm{b}^{[\ell]}\in\mathbb{R}^{m_{\ell+1}}$ are the weight matrix and bias vector of the $\ell$-th hidden layer, respectively, $m_0=d$ and $m_{L+1}$ is the dimension of the output, $``\circ"$ stands for the elementary-wise operation. $\sigma(\cdot)$ is an element-wise activation function. We also denote the output of DNNs by $\bm{y}(\bm{x}; \bm{\theta})$ with $\bm{\theta}$ standing for the parameter set $\left(\bm{W}^{[1]},\cdots \bm{W}^{[L]}, \bm{b}^{[1]},\cdots \bm{b}^{[L]}\right)$.
	
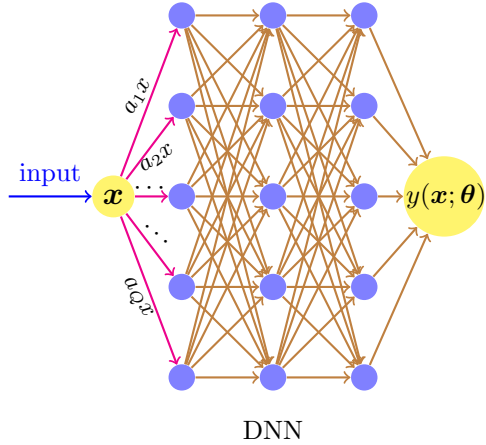
\begin{figure}[H]
		\begin{center}
			\begin{tikzpicture}[scale=0.6]	
			\node(in) at (-2.5,3){};
			\node[circle, fill=yellow!70,inner sep=2.25pt] (x) at  (0.0,3){\large$\bm{x}$};
			
			\draw[line width=0.8pt,color=blue,->] (in) --node[above]{input} (x);
			
			\node[circle, fill=blue!50,inner sep=3.5pt] (h01) at (1.5,-1){};
			\node[circle, fill=blue!50,inner sep=3.5pt] (h02) at (1.5,1){};
			\node[circle, fill=blue!50,inner sep=3.5pt] (h03) at (1.5,3){};
			\node[circle, fill=blue!50,inner sep=3.5pt] (h04) at (1.5,5){};
			\node[circle, fill=blue!50,inner sep=3.5pt] (h05) at (1.5,7){};
			
			\draw[line width=0.8pt,color=magenta,->] (x) --node[below, rotate=-60,yshift=0.5mm,color=black]{\small$a_{Q}x$} (h01);
			\draw[line width=0.8pt,color=magenta,->] (x) --node[above,rotate=-40,yshift=-0.5mm,color=black]{$\cdots$} (h02);
			\draw[line width=0.8pt,color=magenta,->] (x) --node[above,yshift=-1mm,color=black]{$\cdots$} (h03);
			\draw[line width=0.8pt,color=magenta,->] (x) --node[below,rotate=35,yshift=0.5mm,color=black]{\small$a_{2}x$} (h04);
			\draw[line width=0.8pt,color=magenta,->] (x) --node[above, xshift=0.5mm,yshift=-0.35mm,rotate=60,color=black]{\small $a_{1}x$} (h05);
			
			\node[circle, fill=blue!50,inner sep=3.5pt] (h11) at (3.5,-1.0){};
			\node[circle, fill=blue!50,inner sep=3.5pt] (h12) at (3.5,1){};
			\node[circle, fill=blue!50,inner sep=3.5pt] (h13) at (3.5,3){};
			\node[circle, fill=blue!50,inner sep=3.5pt] (h14) at (3.5,5){};
			\node[circle, fill=blue!50,inner sep=3.5pt] (h15) at (3.5,7){};
			\node[circle, fill=blue!0,inner sep=1.5pt] (DNN) at (3.5,-2.15){DNN};
			
			\draw[line width=0.8pt,color=brown,->] (h01) -- (h11);
			\draw[line width=0.8pt,color=brown,->] (h01) -- (h12);
			\draw[line width=0.8pt,color=brown,->] (h01) -- (h13);
			\draw[line width=0.8pt,color=brown,->] (h01) -- (h14);
			\draw[line width=0.8pt,color=brown,->] (h01) -- (h15);
			
			\draw[line width=0.8pt,color=brown,->] (h02) -- (h11);
			\draw[line width=0.8pt,color=brown,->] (h02) -- (h12);
			\draw[line width=0.8pt,color=brown,->] (h02) -- (h13);
			\draw[line width=0.8pt,color=brown,->] (h02) -- (h14);
			\draw[line width=0.8pt,color=brown,->] (h02) -- (h15);
			
			\draw[line width=0.8pt,color=brown,->] (h03) -- (h11);
			\draw[line width=0.8pt,color=brown,->] (h03) -- (h12);
			\draw[line width=0.8pt,color=brown,->] (h03) -- (h13);
			\draw[line width=0.8pt,color=brown,->] (h03) -- (h14);
			\draw[line width=0.8pt,color=brown,->] (h03) -- (h15);
			
			\draw[line width=0.8pt,color=brown,->] (h04) -- (h11);
			\draw[line width=0.8pt,color=brown,->] (h04) -- (h12);
			\draw[line width=0.8pt,color=brown,->] (h04) -- (h13);
			\draw[line width=0.8pt,color=brown,->] (h04) -- (h14);
			\draw[line width=0.8pt,color=brown,->] (h04) -- (h15);
			
			\draw[line width=0.8pt,color=brown,->] (h05) -- (h11);
			\draw[line width=0.8pt,color=brown,->] (h05) -- (h12);
			\draw[line width=0.8pt,color=brown,->] (h05) -- (h13);
			\draw[line width=0.8pt,color=brown,->] (h05) -- (h14);
			\draw[line width=0.8pt,color=brown,->] (h05) -- (h15);
			
			\node[circle, fill=blue!50,inner sep=3.5pt] (h21) at (5.5,-1.0){};
			\node[circle, fill=blue!50,inner sep=3.5pt] (h22) at (5.5,1){};
			\node[circle, fill=blue!50,inner sep=3.5pt] (h23) at (5.5,3){};
			\node[circle, fill=blue!50,inner sep=3.5pt] (h24) at (5.5,5){};
			\node[circle, fill=blue!50,inner sep=3.5pt] (h25) at (5.5,7){};
			\node[circle, fill=yellow!70,inner sep=0.1pt,text width=1.0cm] (u) at (7.25,3){$y(\bm{x};\bm{\theta})$};
			
			\draw[line width=0.8pt,color=brown,->] (h11) -- (h21);
			\draw[line width=0.8pt,color=brown,->] (h11) -- (h22);
			\draw[line width=0.8pt,color=brown,->] (h11) -- (h23);
			\draw[line width=0.8pt,color=brown,->] (h11) -- (h24);
			\draw[line width=0.8pt,color=brown,->] (h11) -- (h25);
			
			\draw[line width=0.8pt,color=brown,->] (h12) -- (h21);
			\draw[line width=0.8pt,color=brown,->] (h12) -- (h22);
			\draw[line width=0.8pt,color=brown,->] (h12) -- (h23);
			\draw[line width=0.8pt,color=brown,->] (h12) -- (h24);
			\draw[line width=0.8pt,color=brown,->] (h12) -- (h25);
			
			\draw[line width=0.8pt,color=brown,->] (h13) -- (h21);
			\draw[line width=0.8pt,color=brown,->] (h13) -- (h22);
			\draw[line width=0.8pt,color=brown,->] (h13) -- (h23);
			\draw[line width=0.8pt,color=brown,->] (h13) -- (h24);
			\draw[line width=0.8pt,color=brown,->] (h13) -- (h25);
			
			\draw[line width=0.8pt,color=brown,->] (h14) -- (h21);
			\draw[line width=0.8pt,color=brown,->] (h14) -- (h22);
			\draw[line width=0.8pt,color=brown,->] (h14) -- (h23);
			\draw[line width=0.8pt,color=brown,->] (h14) -- (h24);
			\draw[line width=0.8pt,color=brown,->] (h14) -- (h25);
			
			\draw[line width=0.8pt,color=brown,->] (h15) -- (h21);
			\draw[line width=0.8pt,color=brown,->] (h15) -- (h22);
			\draw[line width=0.8pt,color=brown,->] (h15) -- (h23);
			\draw[line width=0.8pt,color=brown,->] (h15) -- (h24);
			\draw[line width=0.8pt,color=brown,->] (h15) -- (h25);
			
			\draw[line width=0.8pt,color=brown,->] (h21) -- (u);
			\draw[line width=0.8pt,color=brown,->] (h22) -- (u);
			\draw[line width=0.8pt,color=brown,->] (h23) -- (u);
			\draw[line width=0.8pt,color=brown,->] (h24) -- (u);
			\draw[line width=0.8pt,color=brown,->] (h25) -- (u);
			\end{tikzpicture}
		\end{center}
		\caption{A schematic diagram for MScaleDNN with $N_i = 1$, $i=1,\cdots, Q$.}
		\label{fig2mscalednn}
	\end{figure}

	A normal DNN is usually inadequate to solve multi-scale PDEs as it tends to stagnate on low frequency components. The MscaleDNN architecture was proposed \cite{liu2020multi,li2020elliptic} based on the frequency principle to alleviate this problem. It is illustrated in Figure \ref{fig2mscalednn} and described in the following: We first divide the neurons in the first hidden-layer into $Q$ parts, and construct the following vector 
	\begin{equation}\label{separate}
	\Lambda = (\overbrace{a_1,\cdots,a_1}^{N_1}, \overbrace{a_2,\cdots,a_2}^{N_2},\cdots,\overbrace{a_Q,\cdots,a_Q}^{N_Q})^T
	\end{equation}
	where $\{a_q\}_{q=1}^{Q}$ are positive scale-factors (usually in ascending order), and $\{N_q\}_{q=1}^{Q}$ are the number of the duplicated $q$-th scale-factor. The input data $\bm{x}$ is scaled by $\Lambda$ such that $\tilde{\bm{x}} = \Lambda\odot (\bm{W}^{1}\bm{x}+\bm{b}^{1})$, then $\tilde{\bm{x}}$ is fed into the subsequent module of MscaleDNN, here and hereinafter $\odot$ stands for the element-wise product.
	
	From the viewpoint of function approximation, the first layer of the MscaleDNN model can be regarded as a series of basis functions with various scales and the output of MscaleDNN is the (nonlinear) combination of those basis functions \cite{liu2020multi,li2020elliptic,wang2020eigenvector}. 
	
	We use the residual neural network (ResNet) \cite{he2016deep} to overcome the vanishing gradient phenomenon in backpropagation by introducing skip connections between nonadjacent layers. For example, the ResNet unit with one-step connection produces a filtered version $\bm{y}^{[\ell+1]}(\bm{x};\bm{\theta})$ for the input $\bm{y}^{[\ell]}(\bm{x};\bm{\theta})$ is as follows
	\begin{equation*}
	\bm{y}^{[\ell+1]}(\bm{x};\bm{\theta}) = \bm{y}^{[\ell]}(\bm{x};\bm{\theta})+\sigma \circ\bigg{(}\bm{W}^{[\ell+1]}\bm{y}^{[\ell]}(\bm{x};\bm{\theta})+\bm{b}^{[\ell+1]}\bigg{)}.
	\end{equation*}
	ResNet can accelerate the training process and improve the performance of DNNs. For scientific computation tasks, it can help improve the capability of DNNs to approximate high-order derivatives and solutions of PDEs \cite{e2018the,zou2020deep}. 
	
	\subsection{Subspace decomposition}
	
	It is natural to decompose multi-scale solutions into smooth components and oscillatory components. For example, in periodic homogenization with $\mathcal{L}  = -\nabla \cdot a(x/\epsilon) \nabla  $, the solution $u^\epsilon$  of $\mathcal{L} u^\epsilon = f$ can be approximated by $u_0 + \epsilon u_1$, such that $u_0$ is the solution of the homogenized equation $\mathcal{L}^0 u_0 = f$, where $\mathcal{L}^0 = -\nabla \cdot A \nabla $ with $A$ being the homogenized coefficient. $u^0$ is much smoother than $u^\epsilon$, and $u_1 = \epsilon \sum_{i=1}^d\chi_i(x/\epsilon) \frac{\partial u_0}{\partial x_i}$ characterizes the oscillations. $\chi_i$ is the so-called corrector of the homogenization problem or solution of the cell problem \cite{bensoussan1978asymptotic, jikov2012homogenization}, such that
	\begin{equation}
	\left\{
	\begin{array}{c}
	-\nabla_y \cdot a(y) (\nabla_y \xi_i(y) + e_i) = 0,\\
	\xi_i \text{ periodic in }Y.
	\end{array}
	\right.
	\end{equation}
	where $y = x/\epsilon$, and $Y$ is the unit periodic cell.
	
	In many numerical homogenization type multi-scale methods such as VMS, RPS and LOD methods \cite{hughes98, bazilevs2007variational, OwhZhaBer:2014, NMTMA-14-862, MalPet:2014,Henning2014,Hughes1998b}, a coarse solution is computed in a low dimensional approximation space $V_c$ with quasi-optimal approximation, stability and localization properties. The bases in $V_c$ contain fine scale information, and they can be pre-computed in localized subdomains in parallel. For some multi-scale problems, numerical homogenization methods can be proved to achieve guaranteed quantitative error estimate with respect to coarse resolutions. Furthermore, multi-scale methods can be considered as two level numerical methods, and can be generalized as multilevel subspace decomposition to construct multigrid type preconditioners for the efficient resolution of fine scale problem, see \cite{owhadi2017multigrid, xie2019fast}.
	
	\section{SD$^2$NN model to multi-scale problems and the options for activation function}\label{sec:03}
	
	\subsection{Unified SD$^2$NN architecture to solve multi-scale problems}
	
	We introduce subspace-decomposed DNN (called SD$^2$NN) architecture in the following. For simplicity, we employ the deep Ritz method \cite{e2018the} for the solution of multi-scale PDEs \eqref{eq:nonlinear-ms}. Other architectures such as PINN \cite{Raissi2017}, deep Galerkin \cite{sirignano2018dgm} etc. can be adapted.
	
	\paragraph{Continuous variational formulation}
	
	The solution of \eqref{eq:nonlinear-ms} can be obtained by minimizing the following Dirichlet energy 
	\begin{equation}\label{original-variational}
	\mathcal{J}(v) = \frac{1}{2}\int_{\Omega} v'\mathcal{L}vd\bm{x} -\int_{\Omega}fvd\bm{x},
	\end{equation}
	where $v = v(\bm{x})\in \mathcal{V}$ is a trial function, where $\mathcal{V}$ is the admissible function space for $v$. We are looking for the solution 
	\begin{displaymath}
	u = \arg\min_{v\in \mathcal{V}} \mathcal{J}(v).
	\end{displaymath}
	
	Suppose that the solution $u$ has the following coarse/fine decomposition, $u=u_c+u_f$, in which $u_c$ contains the coarse shape and $u_f$ contains the fine details of the multi-scale solution $u$, respectively. Formally,  \eqref{original-variational} can be rewritten as
	\begin{equation}\label{splitVariational}
	\mathcal{J}(v_c, v_f) = \frac{1}{2}\int_{\Omega} (v_c+v_f)'\mathcal{L} (v_c+v_f)d\bm{x} -\int_{\Omega}f(v_c+v_f)d\bm{x}.
	\end{equation}
	To ensure the  well-posedness of the variational problem with respect to $v_c$ and $v_f$, we need to define the respective subspaces $\mathcal{V}_c$ and $\mathcal{V}_f$, in where $v_c$ and $v_f$ "live", which will be specified in the particular examples. The coarse part $v_c$ can be represented by a low-frequency or a normal DNN $y_1(\cdot,\bm{\theta}_1)$, and the fine-part (or high-frequency part) $u_f$ can be represented by a multi-scale DNN $y_2(\cdot,\bm{\theta}_2)$. Here, $\bm{\theta}_1 \in\Theta_1$ and $\bm{\theta}_2\in\Theta_2$ denote the parameters of the underlying DNNs. In this case, we have 
	\begin{equation}
	\begin{array}{cc}
	\mathcal{V}_c & = \bigg{\{} y_1(\bm{x};\bm{\theta}_1) \bigg{|} y_1(\bm{x}; \bm{\theta}_1) = g, \bm{x} \in \partial \Omega; \bm{\theta}_1 \in \Theta_1 \bigg{\}},\\
	\mathcal{V}_f & = \bigg{\{} y_2(\bm{x};\bm{\theta}_2) \bigg{|} y_2(\bm{x};     \bm{\theta}_2) = 0, \bm{x} \in \partial \Omega;\bm{\theta}_2 \in \Theta_2     \bigg{\}}.
	\end{array}
	\label{eqn:vcvf}
	\end{equation}
	\begin{remark}
		Here we present the SD$^2$NN model as a two level model, and it can be easily generalized to a multilevel model by adding more MscaleDNN submodules with appropriate frequency factors. We will illustrate this point through Example \ref{Diffusion1D3scale} in the numerics section.
	\end{remark}
	
	We further introduce a hyper-parameter $\alpha>0$ to control the contribution of fine-part, i.e., $y(\bm{x};\bm{\theta}_1,\bm{\theta}_2)=y_1(\bm{x};\bm{\theta}_1) + \alpha y_2(\bm{x};\bm{\theta}_2)$. Let 
	\begin{equation}\label{contionus_var}
	\begin{aligned}
	\mathcal{J}_\alpha\bigg{(}y_1(\bm{x};\bm{\theta}_1), y_2(\bm{x};\bm{\theta}_2)\bigg{)} &=\frac{1}{2}\int_{\Omega} \bigg{(}y_1(\bm{x};\bm{\theta}_1) + \alpha y_2(\bm{x};\bm{\theta}_2)\bigg{)}'\mathcal{L} \bigg{(}y_1(\bm{x};\bm{\theta}_1) + \alpha y_2(\bm{x};\bm{\theta}_2)\bigg{)}d\bm{x}\\ &-\int_{\Omega}f\bigg{(}y_1(\bm{x};\bm{\theta}_1) + \alpha y_2(\bm{x};\bm{\theta}_2)\bigg{)}d\bm{x},
	\end{aligned}
	\end{equation}
	we then obtain the following variational problem
	\begin{equation*}
	u_c,u_f = \underset{y_1(\bm{x}, \bm{\theta}_1)\in \mathcal{V}_c, y_2(\bm{x}, \bm{\theta}_2)\in \mathcal{V}_f}{\arg\min}~\mathcal{J}_\alpha\bigg{(}y_1(\bm{x};\bm{\theta}_1), y_2(\bm{x};\bm{\theta}_2)\bigg{)}
	\end{equation*}
	
	\paragraph{Discrete Variational Formulation}
	
	The integral in \eqref{contionus_var} can be discretized by Monte Carlo method \cite{robert1999monte}, namely, we define
	\begin{equation*}
	L_{\mathcal{J}_{\alpha}}(S_I;\bm{\theta}_1,\bm{\theta}_2) =\frac{1}{n_{in}}\sum_{i=1}^{n_{in}} \Bigg{[}\frac{1}{2} \bigg{(}y_1(\bm{x}_I^i;\bm{\theta}_1) + \alpha y_2(\bm{x}_I^i;\bm{\theta}_2)\bigg{)}'\mathcal{L} \bigg{(}y_1(\bm{x}_I^i;\bm{\theta}_1) + \alpha y_2(\bm{x}_I^i;\bm{\theta}_2)\bigg{)}-f\bigg{(}y_1(\bm{x}_I^i;\bm{\theta}_1) + \alpha y_2(\bm{x}_I^i;\bm{\theta}_2)\bigg{)}\Bigg{]},
	\end{equation*}
	here and hereinafter $S_I$ stands for the sampling points with uniform distribution in $\Omega$, and 
	\begin{equation*}
	\bm{\theta}_1^*, \bm{\theta}_2^*=
	\underset{\bm{\theta}_1\in{\Theta}_1,\bm{\theta}_2\in{\Theta}_2}{\arg\min}~L_{\mathcal{J}_{\alpha}}(S_I;\bm{\theta}_1,\bm{\theta}_2),
	\end{equation*}
	such that $u_c = y_1(\bm{x};\bm{\theta}_1^*)$, $u_f = \alpha y_2(\bm{x};\bm{\theta}_2^*)$.
	
	\paragraph{Orthogonality constraints}
	
	In order to separate the coarse-part and fine-part of solution, we add the $L^2$ orthogonality constraint   $\int_{\Omega}y_1(\bm{x};\bm{\theta})\cdot \alpha y_2(\bm{x};\bm{\theta})d\bm{x}= 0$ as a penalty term to the loss function, namely, let
	\begin{equation}\label{orthogonality}
	L_{orth}(S_I;\bm{\theta}_1,\bm{\theta}_2) = \bigg{|}\frac{1}{n_{in}}\sum_{i=1}^{n_{in}}y_1(\bm{x}_I^i;\bm{\theta}_1)\cdot \alpha y_2(\bm{x}_I^i;\bm{\theta}_2)\bigg{|}^2, \bm{x}_I^i\in S_I.
	\end{equation}
	
	\paragraph{Boundary condition}
	
	Boundary conditions are important constraints for numerical solution of PDEs. According to the definition of coarse and fine spaces \eqref{eqn:vcvf}, we have the following penalties for the coarse part  $y_1\big(\bm{x};\bm{\theta}_1\big)$ and the fine-part $y_2\big(\bm{x};\bm{\theta}_2\big)$, 
	\begin{equation}\label{bdc}
	L_{bdc}(S_B;\bm{\theta}_1)=\frac{1}{n_{bd}}\sum_{j=1}^{n_{bd}} \bigg{[}\mathcal{B}y_1\big(\bm{x}_B^j;\bm{\theta}_1\big)-g(\bm{x}_B^j)\bigg{]}^2   ~~\text{for}~~\bm{x}_B^j\in S_B,
	\end{equation}
	and 
	\begin{equation}\label{bdf}
	L_{bdf}(S_B;\bm{\theta}_2)=\frac{1}{n_{bd}}\sum_{\mathcal{}=1}^{n_{bd}} \bigg{[}\mathcal{B}\alpha y_2\big(\bm{x}_B^j;\bm{\theta}_2\big)-0\bigg{]}^2   ~~\text{for}~~\bm{x}_B^j\in S_B.
	\end{equation}
	
	For comparison, we also introduce the penalty for the "unified" boundary condition of $y\big(\bm{x};\bm{\theta}_1,\bm{\theta}_2\big)=y_1(\bm{x};\bm{\theta}_1) + \alpha y_2(\bm{x};\bm{\theta}_2)$, i.e., 
	\begin{equation}\label{bdu}
	L_{bdu}(S_B;\bm{\theta}_1)=\frac{1}{n_{bd}}\sum_{j=1}^{n_{bd}} \bigg{[}\mathcal{B}y\big(\bm{x}_B^j;\bm{\theta}_1, \bm{\theta}_2\big)-g(\bm{x}_B^j)\bigg{]}^2   ~~\text{for}~~\bm{x}_B^j\in S_B,
	\end{equation}
	with $S_B$ standing for the collection of sampling points with uniform distribution on $\partial \Omega$. 
	
	We conclude with the following loss function:
	\begin{equation}
	L_{\alpha}({S_I,S_B};\bm{\theta}_1,\bm{\theta}_2) = \underbrace{L_{\mathcal{J}_{\alpha}}(S_I;\bm{\theta}_1,\bm{\theta}_2)}_{loss\_in} + \underbrace{\gamma\bigg{(}L_{bdc}(S_B;\bm{\theta}_1)+ L_{bdf}(S_B;\bm{\theta}_2)\bigg{)}}_{loss\_bd} + \underbrace{\beta L_{orth}(S_I;\bm{\theta}_1,\bm{\theta}_2)}_{loss\_orth} \label{eq: orth-loss}
	\end{equation}
	with $S_I=\{\bm{x}_I^i\}_{i=1}^{n_{it}}$ and $S_B=\{\bm{x}_B^j\}_{j=1}^{n_{bd}}$ being the sets of uniformly distributed sample points on $\Omega$ and $\partial \Omega$, respectively. The first term $loss\_{in}$ minimizes the residual of the PDE, the second term $loss\_{bd}$ enforces the given boundary condition, and the third term $loss\_dot$ imposes the orthogonality constraint. In addition, we introduce two penalty parameter $\gamma$ and $\beta$ to control the contribution of $loss\_bd$ and $loss\_orth$ for loss function, in which $\gamma$ is increasing gradually during the training process \cite{2007Numerical}
	and $\beta$ is a fixed constant.

	Our goal is to find two sets of parameter $\bm{\theta}_1, \bm{\theta}_2$  which minimize the loss function $L_{\alpha}({S_I,S_B};\bm{\theta}_1, \bm{\theta}_2)$, i.e.,
	\begin{equation*}
	\bm{\theta}_1^*,\bm{\theta}_2^* = \arg\min L_{\alpha}(S_I,S_B;\bm{\theta}_1, \bm{\theta}_2)~~\Longrightarrow~~ u_c(\bm{x}), u_f(\bm{x})= y_1(\bm{x};\bm{\theta}_1^*), \alpha y_2(\bm{x};\bm{\theta}_2^*).
	\end{equation*}
	If $L_{\alpha}({S_I,S_B};\bm{\theta}_1, \bm{\theta}_2)$ is small enough, then $y(\bm{x};\bm{\theta}_1, \bm{\theta}_2)$ will be very close to the solution of \eqref{eq:nonlinear-ms}. The parameter $\bm{\theta}_1$ and $\bm{\theta}_2$ can be computed by stochastic gradient descent method with a pre-defined learning rate schedule. We use the Xavier initialization for the weights and biases \cite{glorot2010understanding,rotskoff2018parameters}. In other word, the weights and biases are sampled from normal distribution 
	$\mathcal{D}=\mathcal{N}\left(0,\frac{2}{m_{in}+m_{out}}\right)$,
	where $m_{in}$ and $m_{out}$ are the input and output dimensions of the corresponding layer, respectively.

	\subsection{Options for activation function}
	The choice of activation functions is crucial for the performance of DNN based algorithms. Two localized activation functions, namely, $\text{sReLU}(\bm{x}) = \text{ReLU}(\bm{x})*\text{ReLU}(1-\bm{x})$ and $\text{s2ReLU}(\bm{x}) = \sin(2\pi \bm{x})*\text{ReLU}(\bm{x})*\text{ReLU}(1-\bm{x})$ were proposed for MscaleDNN in \cite{liu2020multi, li2020elliptic}. Heuristic analysis and numerical results show that the latter one is smoother and more robust. However, these two activation functions are supported in range $[0,1]$, therefore the output will become zero if the input is outside the interval $[0,1]$. This will affect the performance of MscaleDNN in the training process. 
	
\begin{figure}[H]
		\centering
		\begin{minipage}{6cm}
			\includegraphics[scale=0.4]{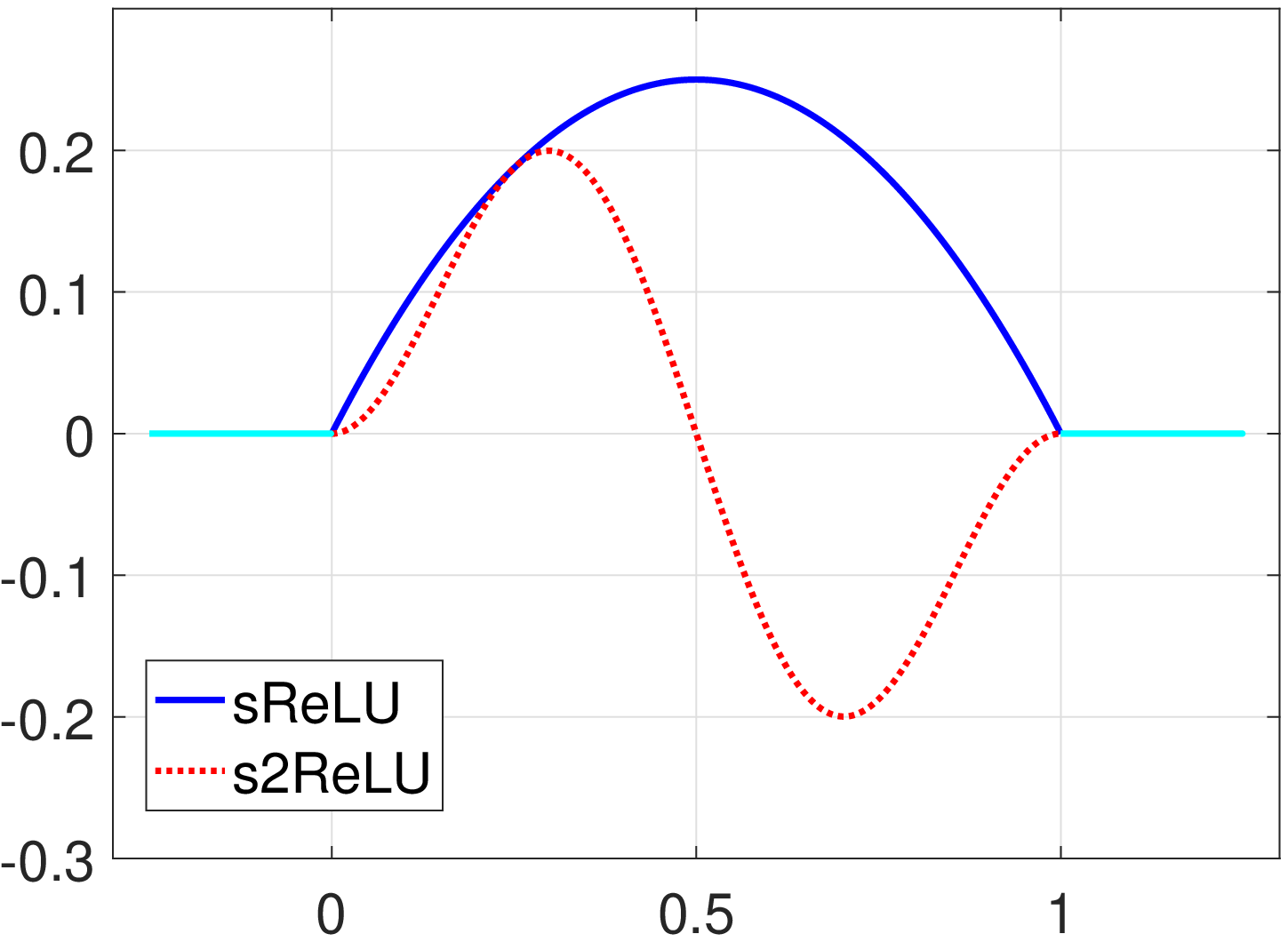}
		\end{minipage}
		\begin{minipage}{6cm}
			\includegraphics[scale=0.4]{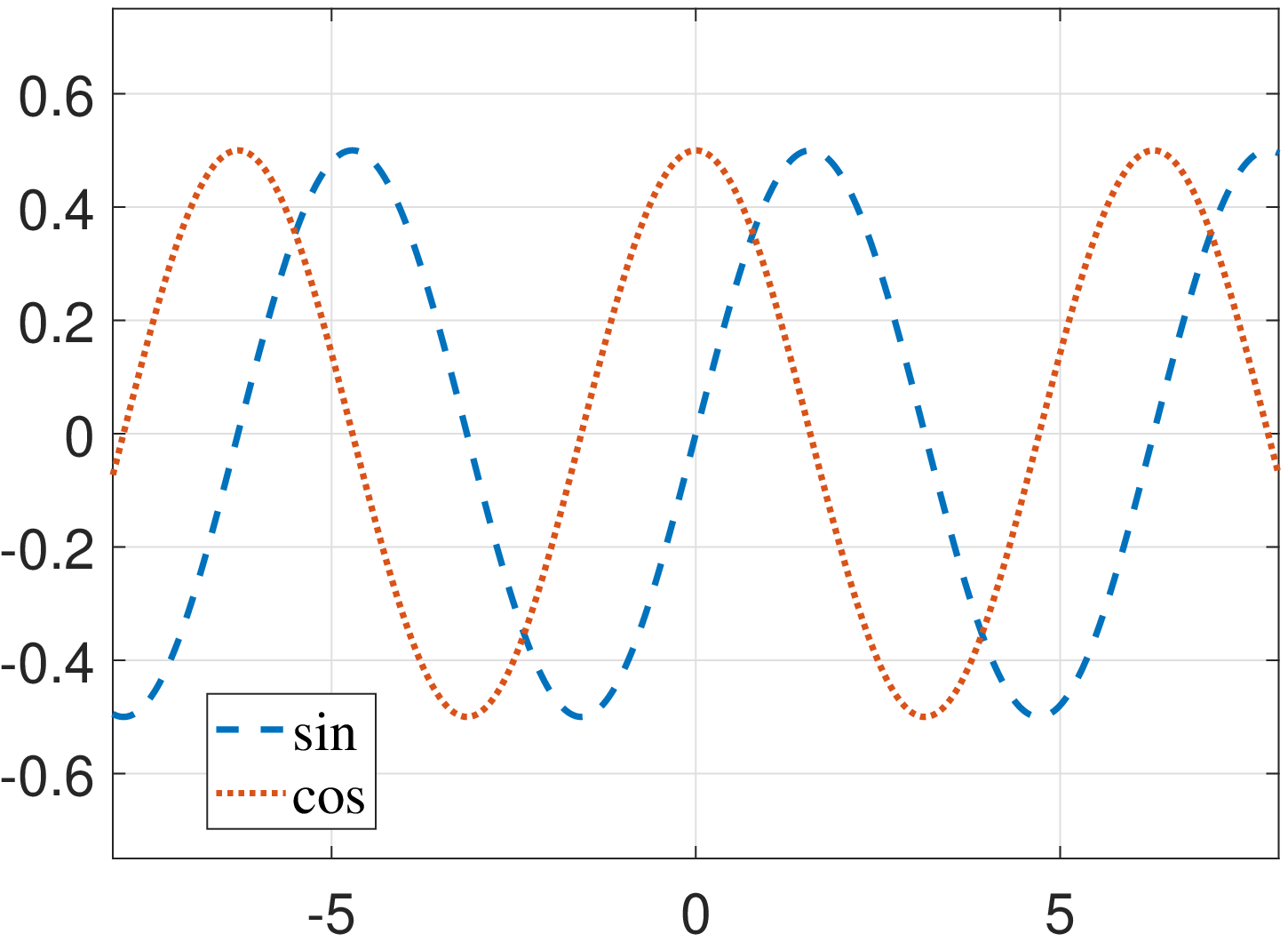}
		\end{minipage}
		\caption{sReLU and s2ReLU functions (left), SFM functions with $s=0.5$ (right).}
		\label{activation}
	\end{figure}

	In this work, we propose the following \emph{soften Fourier mapping} (\emph{SFM}) activation function inspired by the trigonometric activation function proposed in  \cite{Matthew2020Fourier,cai2020phase,wang2020eigenvector}, namely, 
	\begin{equation*}
	\sigma(\bm{z}) = 
	s\times\left[\begin{array}{c}
	\cos(\bm{z})\\
	\sin(\bm{z}) 
	\end{array}
	\right],
	\end{equation*}
	\noindent where the \emph{relaxation} parameter $s\in(0,1]$ is used to control the range of output. Empirically, we find that $s=0.5$ is a good choice, see Fig. \ref{activation}. With the SFM activation function, the MscaleDNN can be regard as a pipeline of Fourier-like transform. The first hidden layer of the MscaleDNN architecture mimic the Fourier expansion, and the remaining structure learns the approximate Fourier coefficients, which are often relatively less oscillate with respect to the input. In this sense, the learning can be effectively accelerated. 
	
\begin{figure}[H]
		\begin{center}
			\begin{tikzpicture}[scale=0.7]	
			\draw[thick, dashed, draw = purple](-2.0, 9.4)-- (9.25, 9.4) -- (9.25, -2.5) -- (-2.0, -2.5) -- (-2.0, 9.4);	
			\node[circle, fill=red!60,inner sep=2.25pt] (xit) at  (-2.75,6.5){\small$\bm{x}_{I}$};
			\node[circle, fill=gray!60,inner sep=2.25pt] (xbd) at  (-2.75,0.5){\small$\bm{x}_{B}$};
			
			\node[circle, fill=orange!70,inner sep=2.25pt] (x) at  (-1.5,3.5){\large$\bm{x}$};
			
			\draw[thick, ->, draw = purple](xit)-- (-2,5) -- (x);
			\draw[thick, ->, draw = purple](xbd)-- (-2,2) -- (x);
			
			\node[circle, fill=blue!50,inner sep=0.1pt] (h01) at (1.5,4.5){\small$\int$};
			\node[circle, fill=blue!50,inner sep=0.1pt] (h02) at (1.5,5.5){\small$\int$};
			\node[circle, fill=blue! 0,inner sep=2pt] (h03) at (1.5,6.5){\small$\vdots$};
			\node[circle, fill=blue!50,inner sep=0.1pt] (h04) at (1.5,7.5){\small$\int$};
			\node[circle, fill=blue!50,inner sep=0.1pt] (h05) at (1.5,8.5){\small$\int$};
			
			\draw[line width=0.8pt,color=red,->] (x) -- (h01);
			\draw[line width=0.8pt,color=red,->] (x) -- (h02);
			\draw[line width=0.8pt,color=red,->] (x) -- (h04);
			\draw[line width=0.8pt,color=red,->] (x) -- (h05);		
			
			\node[circle, fill=blue!50,inner sep=0.1pt] (h11) at (4.25,4.5){\small$\int$};
			\node[circle, fill=blue!50,inner sep=0.1pt] (h12) at (4.25,5.5){\small$\int$};
			\node[circle, fill=blue!0 ,inner sep=2pt  ] (h13) at (4.25,6.5){\small$\vdots$};
			\node[circle, fill=blue!50,inner sep=0.1pt] (h14) at (4.25,7.5){\small$\int$};
			\node[circle, fill=blue!50,inner sep=0.1pt] (h15) at (4.25,8.5){\small$\int$};
			
			\node[rectangle, fill=white!50,inner sep=0.1pt] (low) at (3.5,9.10){Low-freqency or normal NN};
			
			\draw[line width=0.8pt,color=brown,->] (h01) -- (h11);
			\draw[line width=0.8pt,color=brown,->] (h01) -- (h12);
			\draw[line width=0.8pt,color=brown,->] (h01) -- (h14);
			\draw[line width=0.8pt,color=brown,->] (h01) -- (h15);
			
			\draw[line width=0.8pt,color=brown,->] (h02) -- (h11);
			\draw[line width=0.8pt,color=brown,->] (h02) -- (h12);
			\draw[line width=0.8pt,color=brown,->] (h02) -- (h14);
			\draw[line width=0.8pt,color=brown,->] (h02) -- (h15);
			
			\draw[line width=0.8pt,color=brown,->] (h04) -- (h11);
			\draw[line width=0.8pt,color=brown,->] (h04) -- (h12);
			\draw[line width=0.8pt,color=brown,->] (h04) -- (h14);
			\draw[line width=0.8pt,color=brown,->] (h04) -- (h15);
			
			\draw[line width=0.8pt,color=brown,->] (h05) -- (h11);
			\draw[line width=0.8pt,color=brown,->] (h05) -- (h12);
			\draw[line width=0.8pt,color=brown,->] (h05) -- (h14);
			\draw[line width=0.8pt,color=brown,->] (h05) -- (h15);
			
			\node[circle, fill=blue!50,inner sep=0.1pt] (h21) at (7,4.5){\small$\int$};
			\node[circle, fill=blue!50,inner sep=0.1pt] (h22) at (7,5.5){\small$\int$};
			\node[circle, fill=blue!0,inner sep=2pt] (h23) at (7,6.5){\small$\vdots$};
			\node[circle, fill=blue!50,inner sep=0.1pt] (h24) at (7,7.5){\small$\int$};
			\node[circle, fill=blue!50,inner sep=0.1pt] (h25) at (7,8.5){\small$\int$};
			
			\draw[line width=0.8pt,color=brown,->] (h11) -- (h21);
			\draw[line width=0.8pt,color=brown,->] (h11) -- (h22);
			\draw[line width=0.8pt,color=brown,->] (h11) -- (h24);
			\draw[line width=0.8pt,color=brown,->] (h11) -- (h25);
			
			\draw[line width=0.8pt,color=brown,->] (h12) -- (h21);
			\draw[line width=0.8pt,color=brown,->] (h12) -- (h22);
			\draw[line width=0.8pt,color=brown,->] (h12) -- (h24);
			\draw[line width=0.8pt,color=brown,->] (h12) -- (h25);
			
			\draw[line width=0.8pt,color=brown,->] (h14) -- (h21);
			\draw[line width=0.8pt,color=brown,->] (h14) -- (h22);
			\draw[line width=0.8pt,color=brown,->] (h14) -- (h24);
			\draw[line width=0.8pt,color=brown,->] (h14) -- (h25);
			
			\draw[line width=0.8pt,color=brown,->] (h15) -- (h21);
			\draw[line width=0.8pt,color=brown,->] (h15) -- (h22);
			\draw[line width=0.8pt,color=brown,->] (h15) -- (h24);
			\draw[line width=0.8pt,color=brown,->] (h15) -- (h25);
			
			\node[circle, fill=orange!70,inner sep=0.1pt,text width=1.0cm] (u1) at (8.25,6.5){\scriptsize$y_1(\bm{x};\bm{\theta}_1)$};
			
			\draw[line width=0.8pt,color=red,->] (h21) -- (u1);
			\draw[line width=0.8pt,color=red,->] (h22) -- (u1);
			\draw[line width=0.8pt,color=red,->] (h24) -- (u1);
			\draw[line width=0.8pt,color=red,->] (h25) -- (u1);
			
			\node[circle, fill=red!50,inner sep=0.1pt] (gradU1) at (10.75, 6.5) {\scriptsize$\nabla y_1(\bm{x};\bm{\theta}_1)$};
			
			\draw[line width=0.8pt,color=black,->] (u1) -- (gradU1);
			
			\node[circle, fill=red!50,inner sep=1.25pt] (fx) at (10.75, 3.5) {$f(\bm{x})$};
			\draw[line width=0.8pt,,color=black, ->] (x)--(fx);
			
			\node[rectangle, fill=red!60, inner sep=2.5pt](loss-it) at (14.5,3.5) {$L_{\mathcal{J}_{\alpha}}\bigg{(}y_1(\bm{x};\bm{\theta}_1),  y_2(\bm{x};\bm{\theta}_2)\bigg{)}$};
			
			\draw[line width=0.8pt,color=black,->] (gradU1) -- (loss-it);
			\draw[line width=0.8pt,color=black,->] (fx) -- (loss-it);
			\draw[line width=0.8pt,color=black, ->] (u1)..controls(9,3.75) and (11.2,5.5)..(loss-it);
			
			\node[rectangle, fill=gray!60, inner sep=3pt](lossUbd1) at (13.0,7.75) {\small$|y_1(\bm{x};\bm{\theta}_1)-g(\bm{x})|^2~$ };
			
			\draw[line width=0.8pt,color=black, ->] (u1) ..controls(9.0,8.85) and (11,7.75).. (lossUbd1);
			\node[rectangle, rounded corners=1mm,fill=blue!50,inner sep=3.5pt] (k01) at (1.5,2.5){\small$\begin{matrix} s\cdot\sin(W^{[1]}_1 \tilde{\bm{x}})\\ s\cdot\cos(W^{[1]}_1 \tilde{\bm{x}})\end{matrix}$};
			\node[circle, fill=blue!0,inner sep=2pt] (k03) at (1.5,0.5){$\vdots$};
			\node[rectangle, rounded corners=1mm,fill=blue!50,inner sep=3.5pt] (k05) at (1.5,-1.5){\small$\begin{matrix} s\cdot\sin(W^{[1]}_Q \tilde{\bm{x}})\\ s\cdot\cos(W^{[1]}_Q \tilde{\bm{x}})\end{matrix}$};
			
			\draw[line width=0.8pt,color=red,->] (x) -- node[below, rotate=-30,yshift=0.5mm,color=black]{\small $\tilde{\bm{x}}=a_1\bm{x}$}(0.1,2.5);
			
			\draw[line width=0.8pt,color=red,->] (x) -- node[below, rotate=-70,yshift=0.5mm,color=black]{\small $\tilde{\bm{x}}=a_Q\bm{x}$}(0.1,-1.5);
			
			\node[circle, fill=blue!50,inner sep=0.1pt] (k11) at (4.25,-1.5){\small$\int$};
			\node[circle, fill=blue!50,inner sep=0.1pt] (k12) at (4.25,-0.5){\small$\int$};
			\node[circle, fill=blue!0,inner sep=2pt] (k13) at (4.25,0.5){$\vdots$};
			\node[circle, fill=blue!50,inner sep=0.1pt] (k14) at (4.25,1.5){\small$\int$};
			\node[circle, fill=blue!50,inner sep=0.1pt] (k15) at (4.25,2.5){\small$\int$};

			\node[rectangle, fill=white!50,inner sep=0.1pt] (low) at (5.5,-2.25){High-freqency NN};
			
			\draw[line width=0.8pt,color=brown,->] (k01) -- (k11);
			\draw[line width=0.8pt,color=brown,->] (k01) -- (k12);
			\draw[line width=0.8pt,color=brown,->] (k01) -- (k14);
			\draw[line width=0.8pt,color=brown,->] (k01) -- (k15);
			
			\draw[line width=0.8pt,color=brown,->] (k05) -- (k11);
			\draw[line width=0.8pt,color=brown,->] (k05) -- (k12);
			\draw[line width=0.8pt,color=brown,->] (k05) -- (k14);
			\draw[line width=0.8pt,color=brown,->] (k05) -- (k15);
			
			\node[circle, fill=blue!50,inner sep=0.1pt] (k21) at (7,-1.5){\small$\int$};
			\node[circle, fill=blue!50,inner sep=0.1pt] (k22) at (7,-0.5){\small$\int$};
			\node[circle, fill=blue! 0,inner sep=2pt] (k23) at (7,0.5){$\vdots$};
			\node[circle, fill=blue!50,inner sep=0.1pt] (k24) at (7,1.5){\small$\int$};
			\node[circle, fill=blue!50,inner sep=0.1pt] (k25) at (7,2.5){\small$\int$};
			
			\draw[line width=0.8pt,color=brown,->] (k11) -- (k21);
			\draw[line width=0.8pt,color=brown,->] (k11) -- (k22);
			\draw[line width=0.8pt,color=brown,->] (k11) -- (k24);
			\draw[line width=0.8pt,color=brown,->] (k11) -- (k25);
			
			\draw[line width=0.8pt,color=brown,->] (k12) -- (k21);
			\draw[line width=0.8pt,color=brown,->] (k12) -- (k22);
			\draw[line width=0.8pt,color=brown,->] (k12) -- (k24);
			\draw[line width=0.8pt,color=brown,->] (k12) -- (k25);
			
			\draw[line width=0.8pt,color=brown,->] (k14) -- (k21);
			\draw[line width=0.8pt,color=brown,->] (k14) -- (k22);
			\draw[line width=0.8pt,color=brown,->] (k14) -- (k24);
			\draw[line width=0.8pt,color=brown,->] (k14) -- (k25);
			
			\draw[line width=0.8pt,color=brown,->] (k15) -- (k21);
			\draw[line width=0.8pt,color=brown,->] (k15) -- (k22);
			\draw[line width=0.8pt,color=brown,->] (k15) -- (k24);
			\draw[line width=0.8pt,color=brown,->] (k15) -- (k25);
			
			\node[circle, fill=orange!70,inner sep=0.1pt,text width=1.0cm] (u2) at (8.25,0.75){\scriptsize$y_2(\bm{x};\bm{\theta}_2)$};
			
			\draw[line width=0.8pt,color=red,->] (k21) -- (u2);
			\draw[line width=0.8pt,color=red,->] (k22) -- (u2);
			\draw[line width=0.8pt,color=red,->] (k24) -- (u2);
			\draw[line width=0.8pt,color=red,->] (k25) -- (u2);
			
			\node[circle, fill=red!50,inner sep=0.1pt] (gradU2) at (10.75,0.75) {\scriptsize$\nabla y_2(\bm{x};\bm{\theta}_2)$};
			\draw[line width=0.8pt,color=black,->] (u2) -- (gradU2);
			\draw[line width=0.8pt,color=black, ->] (u2)..controls(9.0,3) and (11.3,1.75)..(loss-it);
			\draw[line width=0.8pt,color=black,->] (gradU2) -- (loss-it);
			
			\node[rectangle, fill=gray!60, inner sep=3pt](lossUbd2) at (13.0,-0.5) {\small$|\alpha y_2(\bm{x};\bm{\theta}_2)-0|^2~$ };
			
			\draw[line width=0.8pt,color=black, ->] (u2) ..controls(9.0,-1.75) and (11,-0.5).. (lossUbd2);	
			
			\node[circle, fill=cyan!60, inner sep=2pt](loss) at (19.0,3.5){loss};
			\node[circle, fill=cyan!60, inner sep=2pt](theta) at (19.0,0.0){$\bm{\theta}_1^*,\bm{\theta}_2^*$};	
			
			\draw[line width=0.8pt,color=black, ->] (loss-it) -- (loss);
			\draw[line width=0.8pt,color=black, ->] (lossUbd1) -- (loss);
			\draw[line width=0.8pt,color=black, ->] (lossUbd2) -- (loss);
			
			\draw[line width=0.8pt,color=black,->] (loss) -- node[rotate=-90,right,xshift=-8mm,yshift=2mm]{\small minimize} (theta);
			
			\node[rectangle, fill=red!60, inner sep=3pt](lossUdU) at (17.75,7.75) {\small$\alpha^2 \big{|}y_1(\bm{x};\bm{\theta}_1)\cdot y_2(\bm{x};\bm{\theta}_2)\big{|}^2$};	
			
			\draw[line width=0.8pt,color=black, -] (u2) ..controls(9.0,-1.75) and (11,-1.25).. (20.5, -1.25);
			\draw[thick, ->,draw = black](18.0, -1.25)--(20.5, -1.25) -- (20.5, 7.75)--(lossUdU);
			\draw[line width=0.8pt,color=black, ->] (u1) ..controls(9.0,9) and (14,8.5).. (lossUdU);
			\draw[line width=0.8pt,color=black, ->] (18.5, 7.3) -- (loss);
			\end{tikzpicture}
		\end{center}
		\caption{Schematic of SD$^2$NN for solving multi-scale problems}
		\label{SD$^2$NN}
	\end{figure}
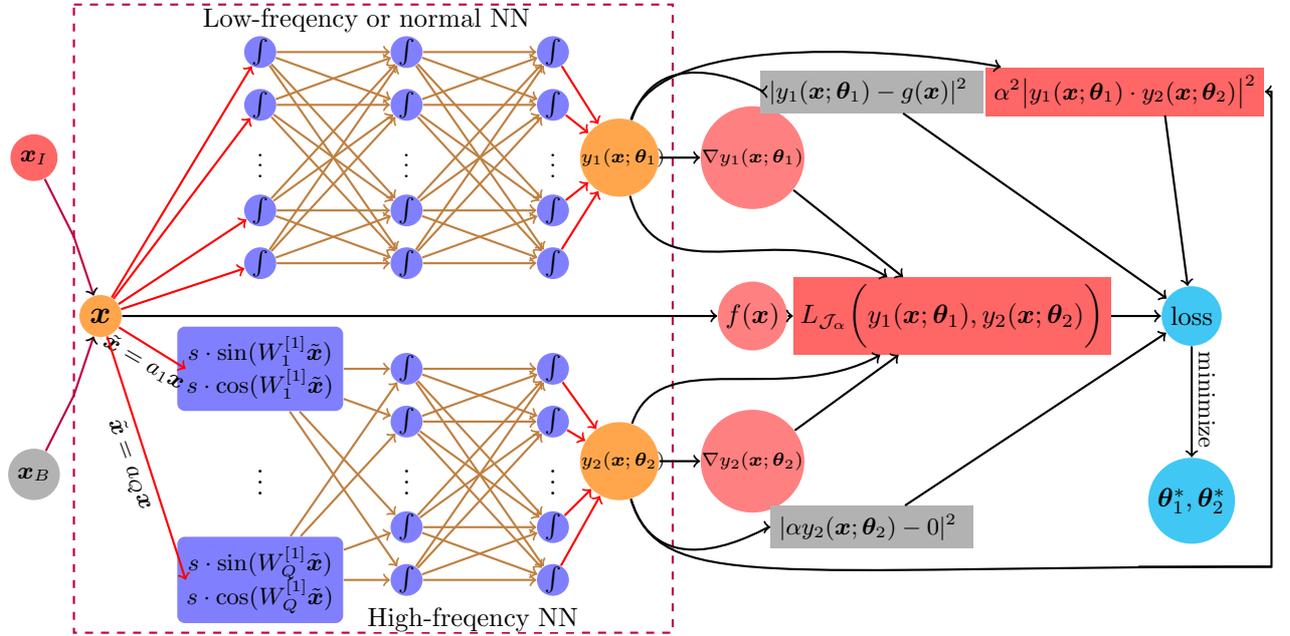

	The above mentioned activation functions are illustrated in Fig.  \ref{SD$^2$NN}.
	
	We conclude the section with the schematic of the SD$^2$NN architecture in Figure \ref{SD$^2$NN}. The output of SD$^2$NN is obtained by a linear layer with no activation function.
	
	\section{Numerical experiments}\label{sec:04}
	In this section, we test the performance of  SD$^2$NN method for multi-scale problems with highly oscillatory coefficients and Poisson-Boltzmann equation in perforated domains. We demonstrate that the SD$^2$NN method can efficiently and simultaneously solve the coarse part and the fine part of the multi-scale solution. We also investigated the effects of different activation functions (for example, SFM and s2ReLU). We show that SD$^2$NN has better performance compared with existing methods such as MscaleDNN in \cite{li2020elliptic, wang2020eigenvector}. 
	
	\subsection{Model and training setup}
	
	\subsubsection{Model setup}
	The details of all the models in the numerical experiments are elaborated in the following.
	\begin{itemize}
		\item 
		\emph{SD$^2$NN1}: We use a normal fully-connected DNN as the coarse submodule and a high-frequency MscaleDNN as the fine submodule. The first layer of the fine submodule has scaling factors $\Lambda=(21,22,23, \cdots,120)$ with $Q=100$, as in \eqref{separate}. The activation function for each hidden layer in the normal DNN is chosen as $\rm{ tanh}(x)$, the activation functions for the first hidden layer and the subsequent layers of MscaleDNN are chosen as SFM with $s=0.5$ and s2ReLU, respectively. 
		\item 
		\emph{SD$^2$NN2}: We use a low-frequency MscaleDNN as the coarse submodule and a high-frequency MscaleDNN as the fine submodule. The first layers of coarse submodule and fine submodule have scaling factors $\Lambda_c=(0.5,1, 1.5, 2,\cdots,19.5, 20)$ with $Q = 40$, and $\Lambda_f=(21,22,23, \cdots,120)$ with $Q = 100$, respectively. The activation function for their first hidden layers are all chosen as SFM with $s=1.0$ and $s=0.5$(denoted as SFM(1.0) and SFM(0.5)), but the activation functions for their subsequent layers are chosen as ${\rm tanh}(x)$ and s2ReLU, respectively. 
		\item 
		\emph{SD$^2$NN3}: The activation functions and scaling factors are same as \emph{SD$^2$NN2}. But we take away the orthogonality penalty term \eqref{eq: orth-loss} from the loss function, and the boundary condition is the unified one in \eqref{bdu}. 
		\item 
		\emph{Mscale}: A MscaleDNN model \cite{li2020elliptic} with s2ReLU activation function for all layers. The neurons of the first layer are divided into $Q = 120$ groups with scaling factors $\Lambda=(1,2,3,\cdots,119, 120)$ in \eqref{separate}. 
		\item 
		\emph{WWP}: A MscaleDNN model \cite{wang2020eigenvector} with hybrid activation function including SFM with $s=1.0$ (denoted as SFM(1.0)) and ${\rm tanh}(x)$. The scaling factors of the first layer are divided as four subgroups. Each subgroup has 30 samples from Gaussian distribution $\mathcal{N}(0;\tau_i^2)$ with $\tau_1=1$, $\tau_1=20$, $\tau_1=50$ and $\tau_4=100$. This model is proposed by Wang, Wang, and Perdikaris in \cite{wang2020eigenvector}, and we denote it as \emph{WWP}.
	\end{itemize}
	\begin{table}[H]
		\centering
		\caption{ Comparisons for the above models}
		\label{comparison}
		\begin{tabular}{|l|c|c|c|c|c|c|}
			\hline
			Model          &Submodules &Activation &Scale factor                           & Boundary                          & Orthogonality \\ \hline
			\multirow{2}*{SD$^2$NN1}&Normal     &tanh       &-----                                  &individual boundary                &\multirow{2}*{Yes}\\\cline{2-4}
			&MscaleDNN  &SFM(0.5)+s2ReLU &$(21,22,23, \cdots,120)$          & \eqref{bdc} and \eqref{bdf}       &   \\  \hline
			\multirow{2}*{SD$^2$NN2}&MscaleDNN  &SFM(1.0)+tanh   &$(0.5,1, 1.5, 2,\cdots,19.5, 20)$ &{individual boundary}              &\multirow{2}*{Yes}   \\ \cline{2-4}
			&MscaleDNN  &SFM(0.5)+s2ReLU &$(21,22,23, \cdots,120)$          & \eqref{bdc} and \eqref{bdf}       &         \\ \hline
			\multirow{2}*{SD$^2$NN3}&MscaleDNN  &SFM(1.0)+tanh   &$(0.5,1, 1.5, 2,\cdots,19.5, 20)$ &\multirow{2}*{unified boundary \eqref{bdu}}     &\multirow{2}*{No}   \\ \cline{2-4}
			&MscaleDNN  &SFM(0.5)+s2ReLU &$(21,22,23, \cdots,120)$          &     &   \\\hline
			Mscale          & -----     &s2ReLU     &$(1,2,3,\cdots,119, 120)$         &unified boundary \eqref{bdu}                  &-----   \\ \hline
			WWP             & -----     &SFM(1.0)+tanh   &$\begin{matrix}\Lambda=[\Lambda_1;\Lambda_2;\Lambda_3;\Lambda_4]~~\textup{with}\\\Lambda_i\sim \mathcal{N}(0,\tau_i),~~\tau_1=1,\\  \tau_2=20, \tau_3=50, \tau_4=100\end{matrix}$         &unified boundary \eqref{bdu}                   &-----   \\ \hline
		\end{tabular}
	\end{table}
	
	\subsubsection{Training setup}
	We use the relative square error to evaluate the accuracy of different models:
	\begin{equation*}
	REL = \sum_{i=1}^{N'}\frac{|\tilde{u}(\bm{x}^i)-u^*(\bm{x}^i)|^2}{|u^*(\bm{x}^i)|^2}
	\end{equation*}
	where $\tilde{u}(\bm{x}^i)$ and $u^*(\bm{x}^i)$ are the approximate DNN solution and the exact solution, respectively, $\{\bm{x}^i\}_{i=1}^{N'}$ are testing points, and $N'$ is the number of testing points. 
	
	In our numerical experiments, all training and testing data are sampled uniformly in $\Omega$ (or $\partial\Omega$), and all networks are trained by Adam optimizer. The initial learning rate is set as $2\times 10^{-4}$ with a decay rate $5 \times 10^{-5}$ for each training epoch. For visualization of the training process, we test our model every 1000 epochs in the training process. The penalty parameter $\beta$ for the orthogonality constraint \eqref{orthogonality} is set as 20, $\gamma$ for the boundary constraint \eqref{bdc} and \eqref{bdf} is set as
	\begin{equation}
	\gamma=\left\{
	\begin{aligned}
	\gamma_0, \quad &\textup{if}~~i_{\textup{epoch}}<0.1T_{\max}\\
	10\gamma_0,\quad &\textup{if}~~0.1T_{\max}<=i_{\textup{epoch}}<0.2T_{\max}\\
	50\gamma_0, \quad&\textup{if}~~ 0.2T_{\max}<=i_{\textup{epoch}}<0.25T_{\max}\\
	100\gamma_0, \quad&\textup{if}~~ 0.25T_{\max}<=i_{\textup{epoch}}<0.5T_{\max}\\
	200\gamma_0, \quad&\textup{if}~~ 0.5T_{\max}<=i_{\textup{epoch}}<0.75T_{\max}\\
	500\gamma_0, \quad&\textup{otherwise}
	\end{aligned}
	\right.
	\end{equation}
	where $\gamma_0=100$ in all our tests and $T_{\max}$ represents the total epoch number. We implement our code in TensorFlow (version 1.14.0) on a work station (256-GB RAM, single NVIDIA GeForce GTX 2080Ti 12-GB).
	
	\subsection{Numerical examples and results}
	In this section, we test the SD$^2$NN models for multi-scale diffusion and Poisson-Boltzmann equations. 
	
	\begin{example}[1D multi-scale elliptic problem]\label{DiffusionEq_1d_01}
		We use this problem to benchmark our models. Let us consider the following multi-scale elliptic equation in domain $\Omega=[a,b]^d$
		\begin{equation}\label{DiffusionEq}
		\begin{cases}
		-\textrm{div}\bigg{(}A(\bm{x})|\nabla u(\bm{x})|^{p-2}\nabla u(\bm{x})\bigg{)} = f(\bm{x}), ~~\bm{x}\in \Omega,\\
		~~~~~~~~~~~u(\bm{x}) = g(\bm{x}),~~~~~~~~~~~~~~~~~\bm{x}\in \partial\Omega.
		\end{cases}
		\end{equation}
		which has the following Dirichlet energy 
		\begin{equation}\label{pLaplacian-variational}
		\mathcal{J}(v) = \frac{1}{p}\int_{\Omega} A|\nabla v|^pd\bm{x} -\int_{\Omega}fvd\bm{x}.
		\end{equation}
		
		\textbf{Linear Case:}	
		
		We first consider the linear case $p=2$ with $\Omega = [0,1]$ and boundary condition $u(0)=u(1)=0$,
		\begin{equation}\label{DiffusionEq_1d_01_aeps}
		A(x)=\left(2+\cos\left(2\pi\frac{x}{\epsilon}\right)\right)^{-1}
		\end{equation}
		with a small parameter $\epsilon>0$ such that $\epsilon^{-1}\in\mathbb{N}^+$, the unique solution is given by
		\begin{equation}\label{DiffusionEq_1d_01_ueps}
		u(x) = x-x^2+\epsilon\left(\frac{1}{4\pi}\sin\left(2\pi\frac{x}{\epsilon}\right)-\frac{1}{2\pi}x\sin\left(2\pi\frac{x}{\epsilon}\right)-\frac{\epsilon}{4\pi^2}\cos\left(2\pi\frac{x}{\epsilon}\right)+\frac{\epsilon}{4\pi^2}\right).
		\end{equation}
		for $f(x)=1$.
	\end{example}
	
\begin{figure}[H]
		\centering
		\subfigure[difference of exact and DNN solutions]{
			\label{pLaplaceE101c}
			\includegraphics[scale=0.325]{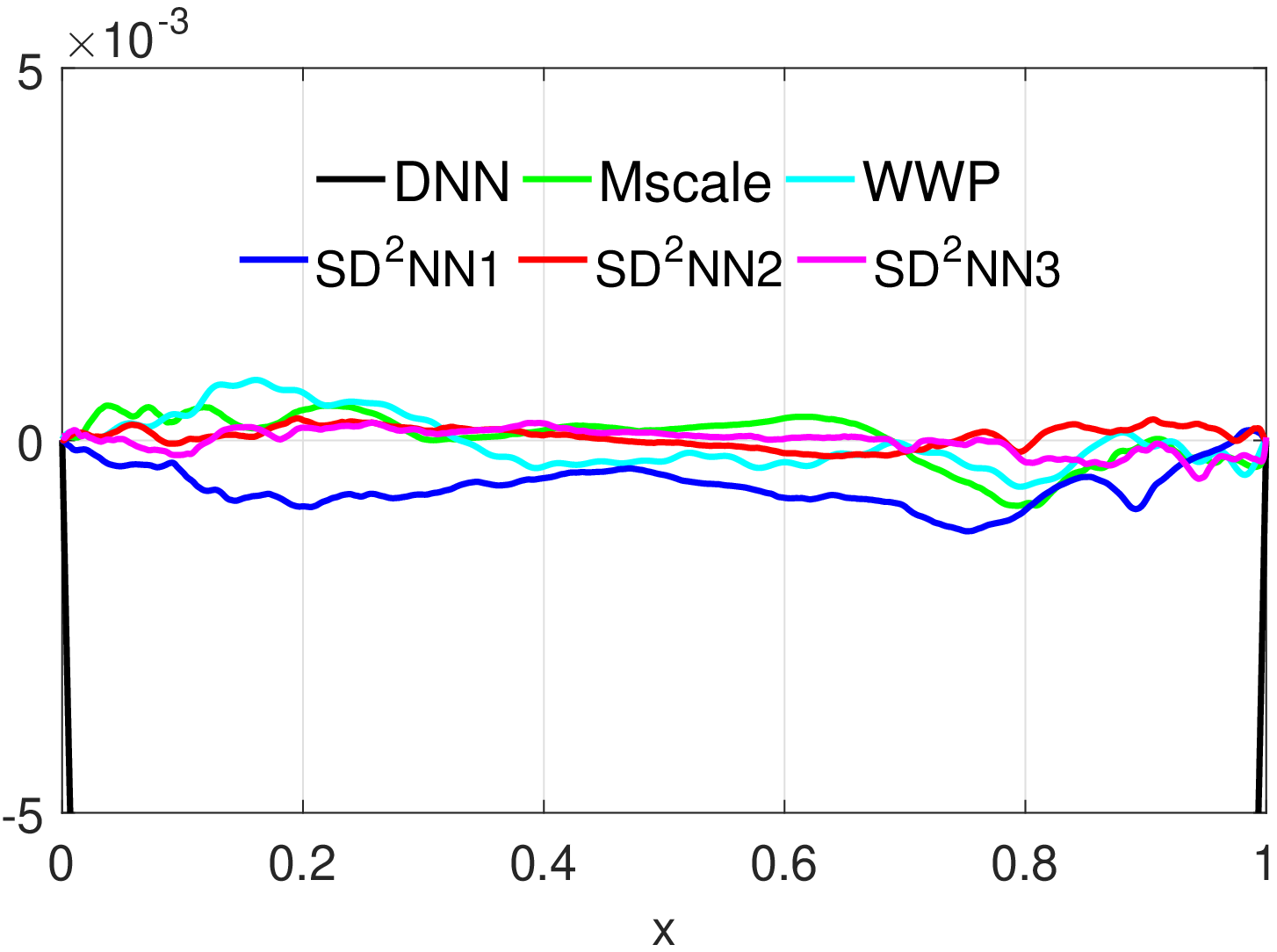}
		}
		\,
		\subfigure[relative error of DNN, Mscale and WWP]{
			\label{pLaplaceE101d}
			\includegraphics[scale=0.325]{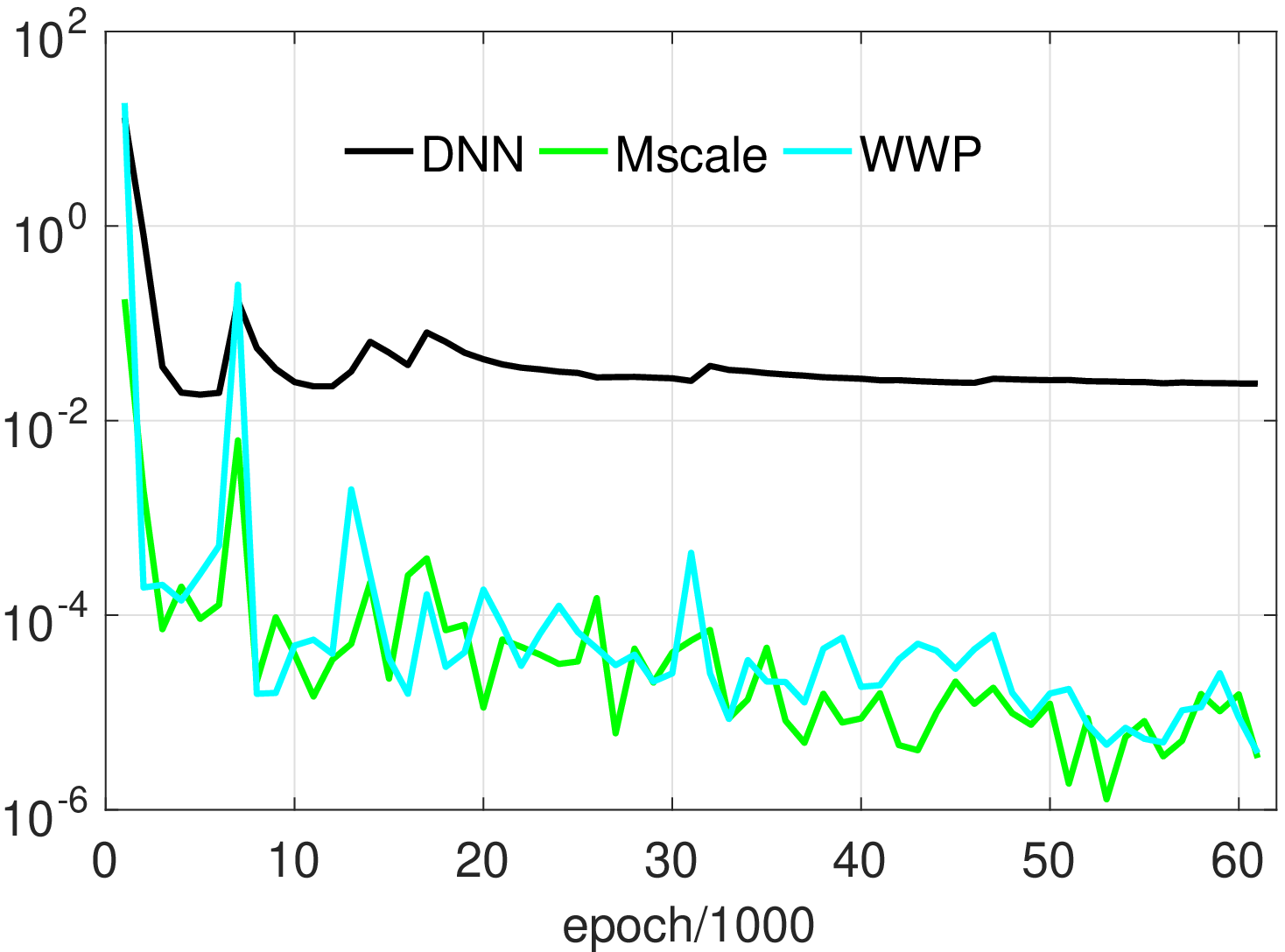}
		}
		\,
		\subfigure[relative error of WWP and SD$^2$NNs]{
			\label{pLaplaceE101e}
			\includegraphics[scale=0.325]{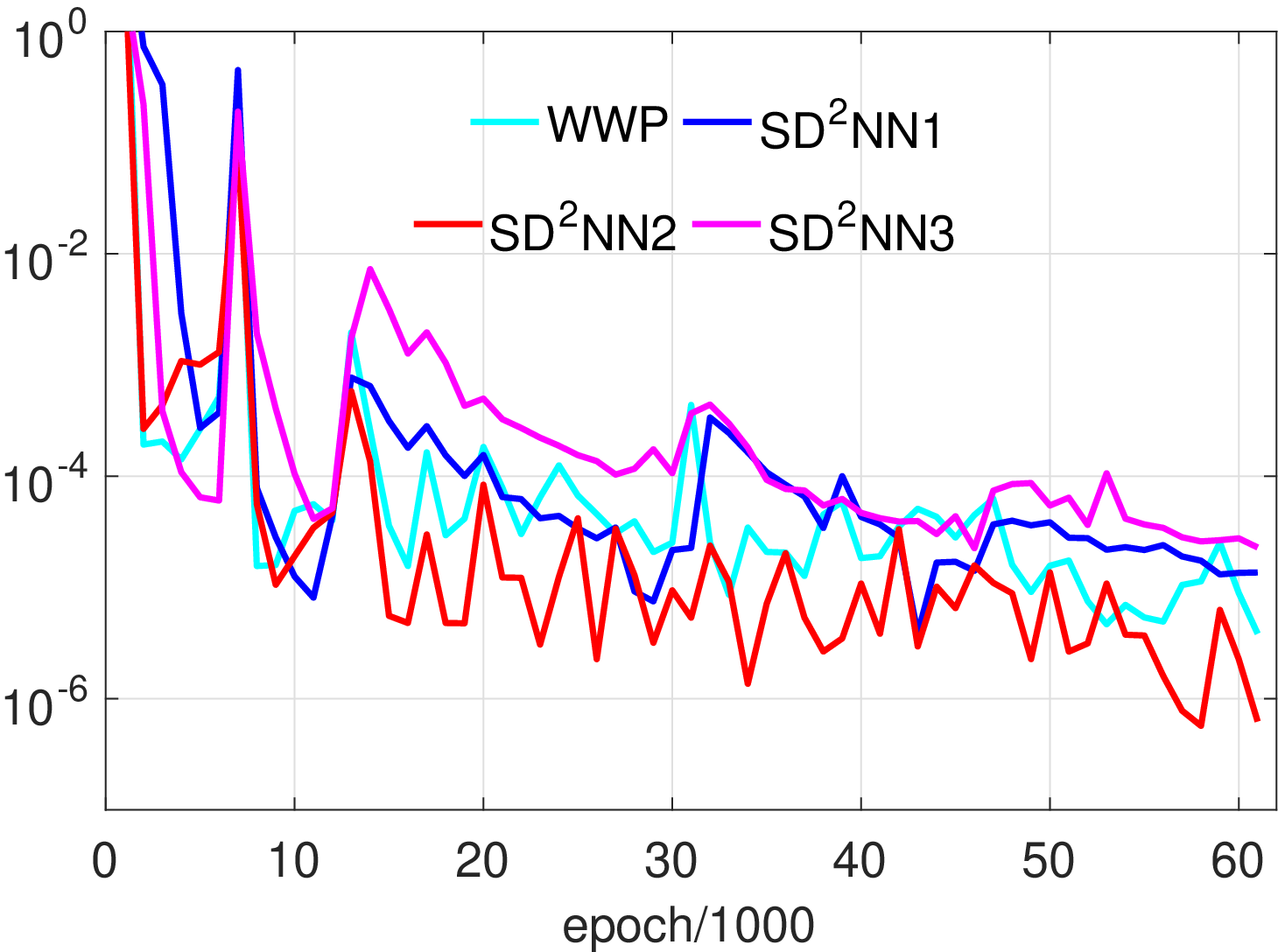}
		}
		\subfigure[difference of coarse solutions]{
			\label{pLaplaceE101a}
			\includegraphics[scale=0.325]{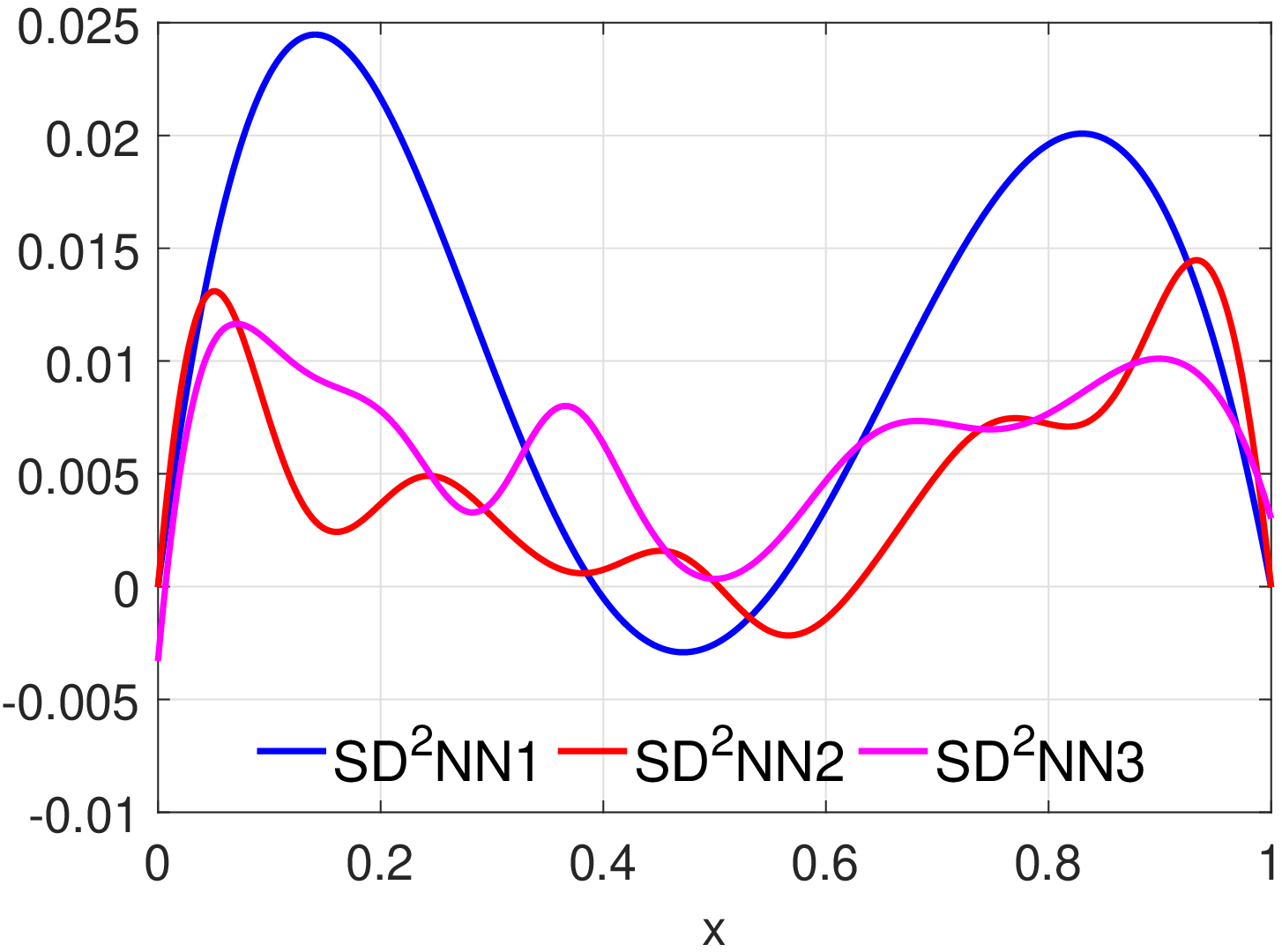}
		}
		\,
		\subfigure[difference of fine solutions]{
			\label{pLaplaceE101b}
			\includegraphics[scale=0.325]{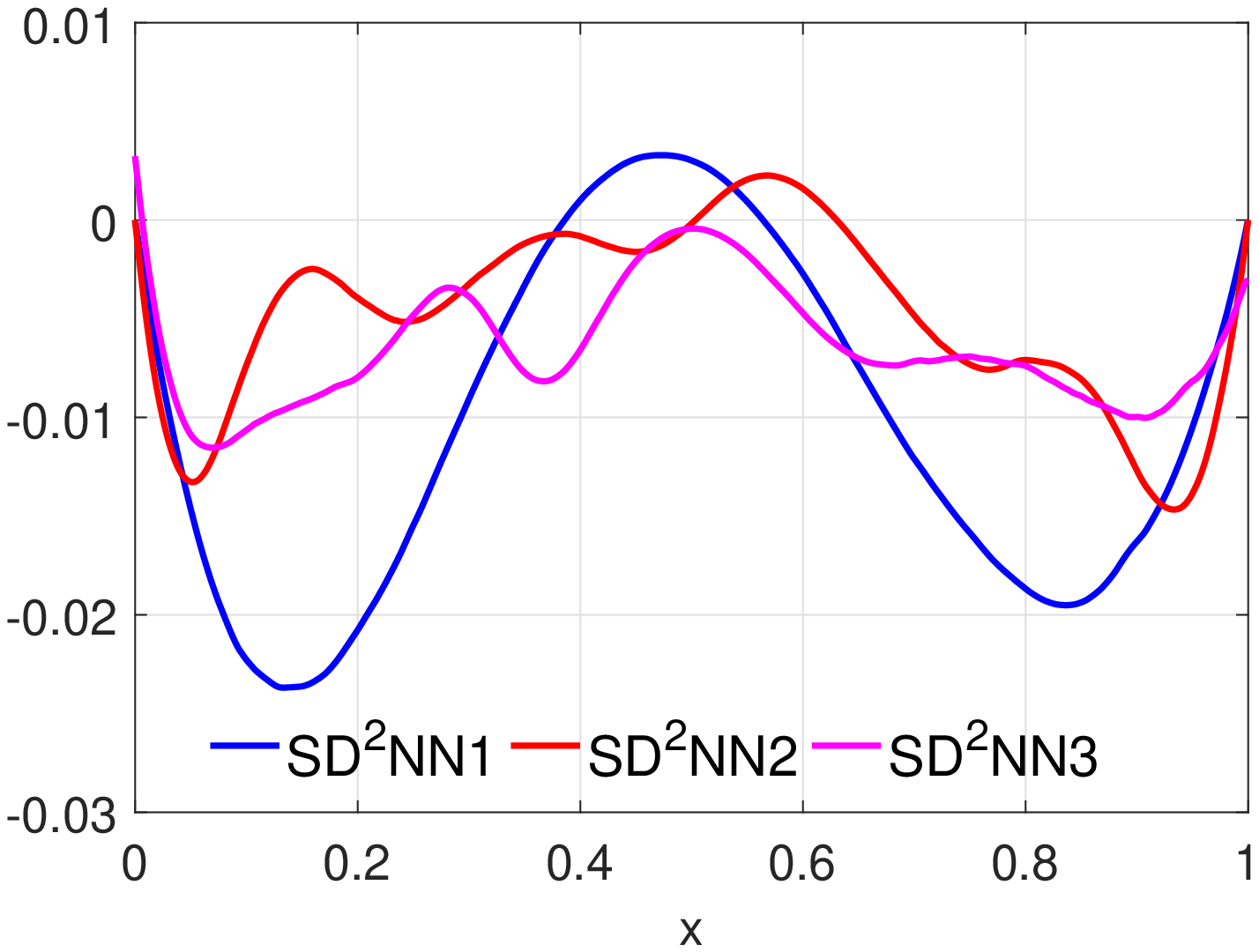}
		}
		\caption{Testing results with $\epsilon=0.1$ for Example \ref{DiffusionEq_1d_01}.}
		\label{pLaplaceE101}
	\end{figure}

\begin{figure}[H]
		\centering
		\subfigure[difference of exact and DNN solutions]{
			\label{pLaplaceE1001c}
			\includegraphics[scale=0.325]{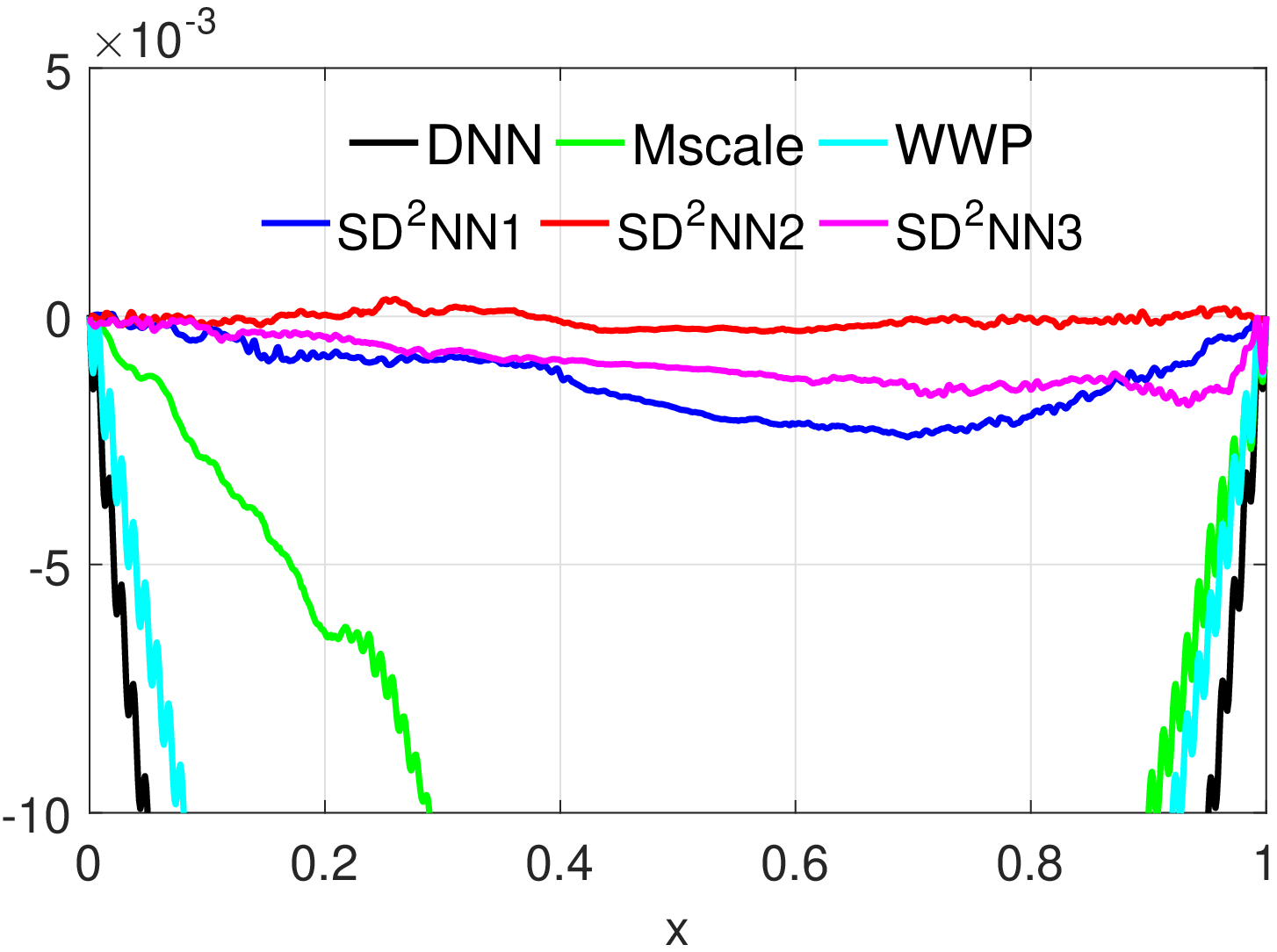}
		}
		\subfigure[relative error of DNN, Mscale and WWP]{
			\label{pLaplaceE1001d}
			\includegraphics[scale=0.325]{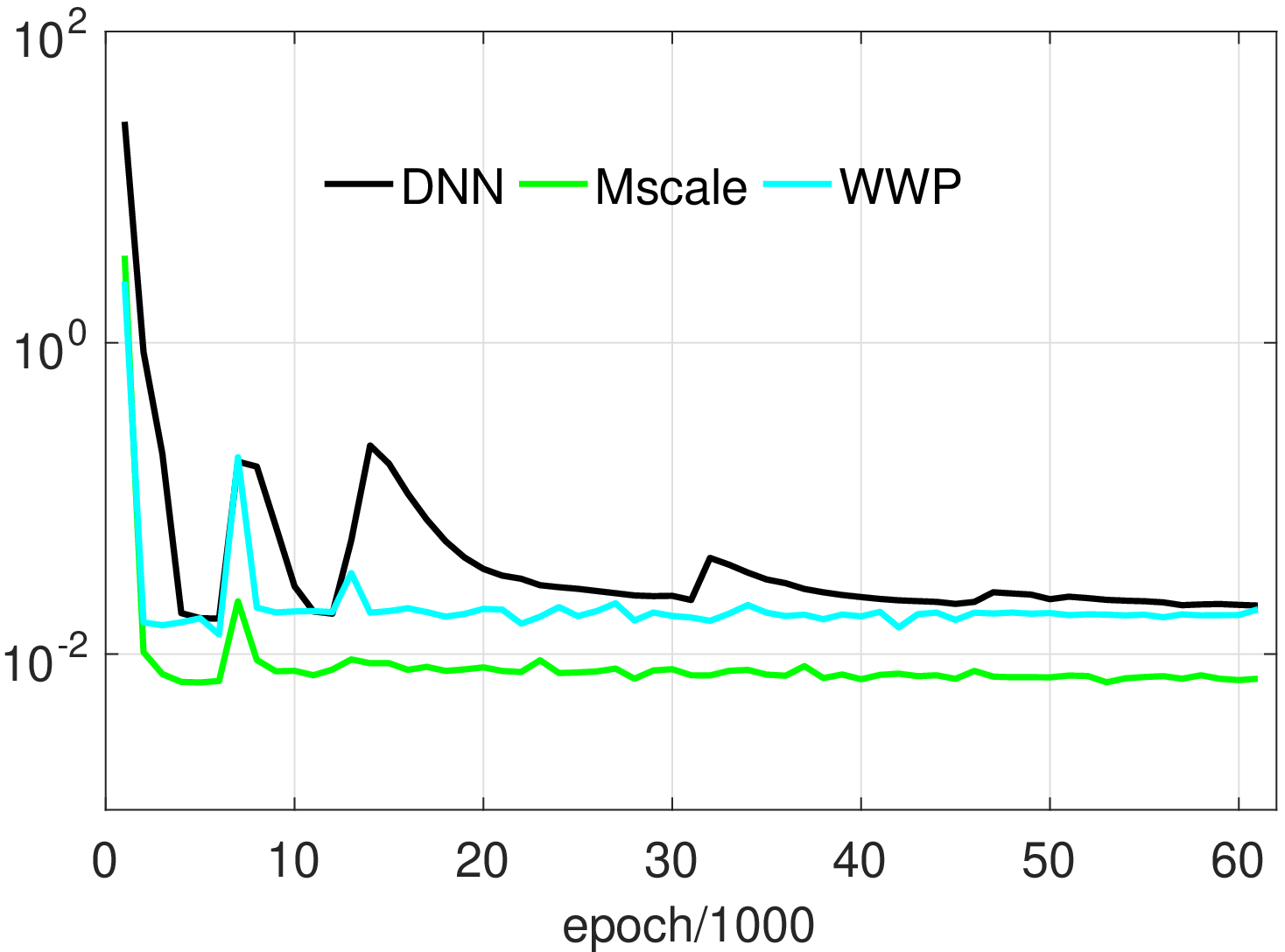}
		}
		\subfigure[relative error of WWP and SD$^2$NNs ]{
			\label{pLaplaceE1001e}
			\includegraphics[scale=0.325]{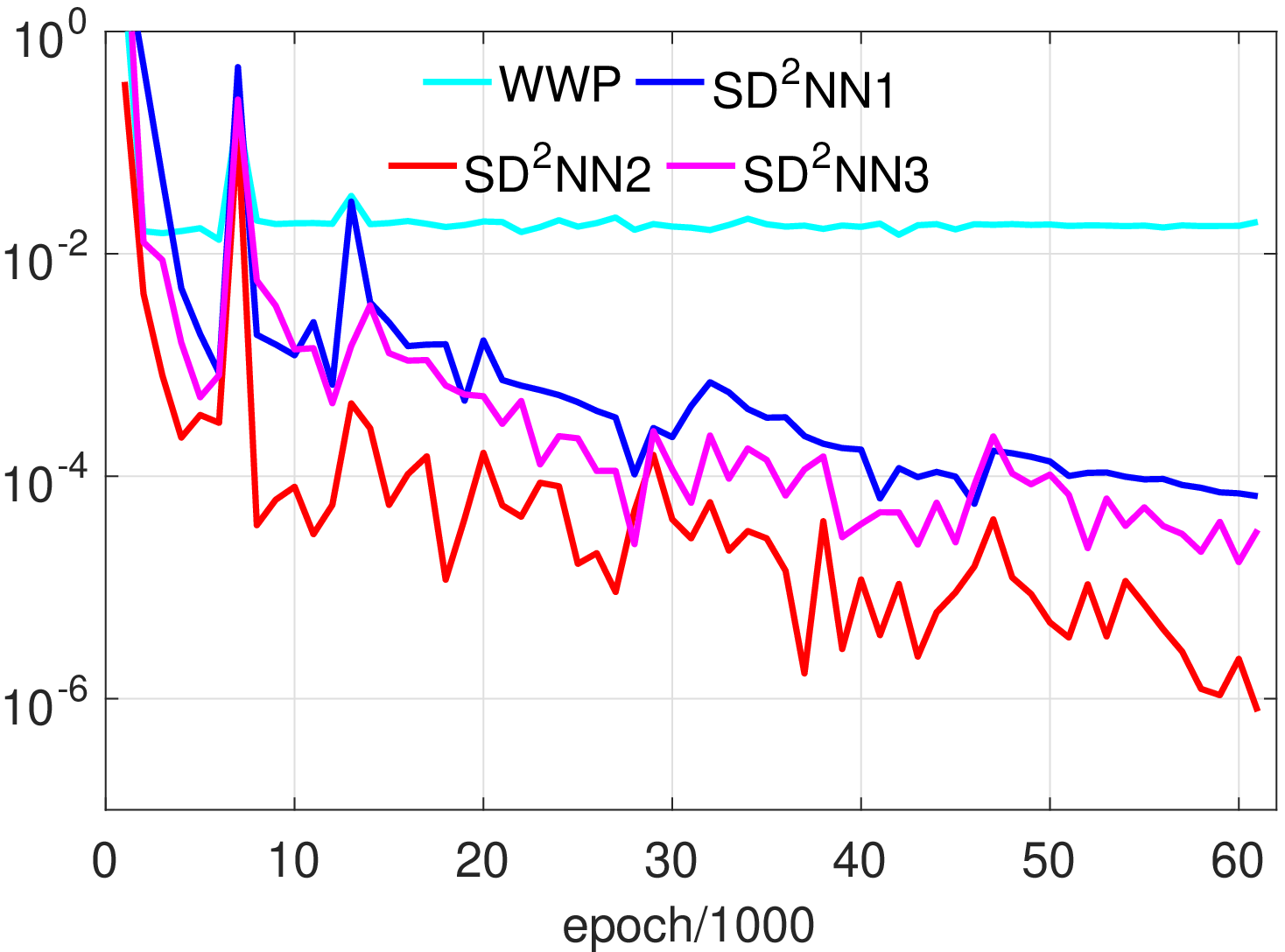}
		}
		\subfigure[difference of coarse solutions]{
			\label{pLaplaceE1001a}
			\includegraphics[scale=0.325]{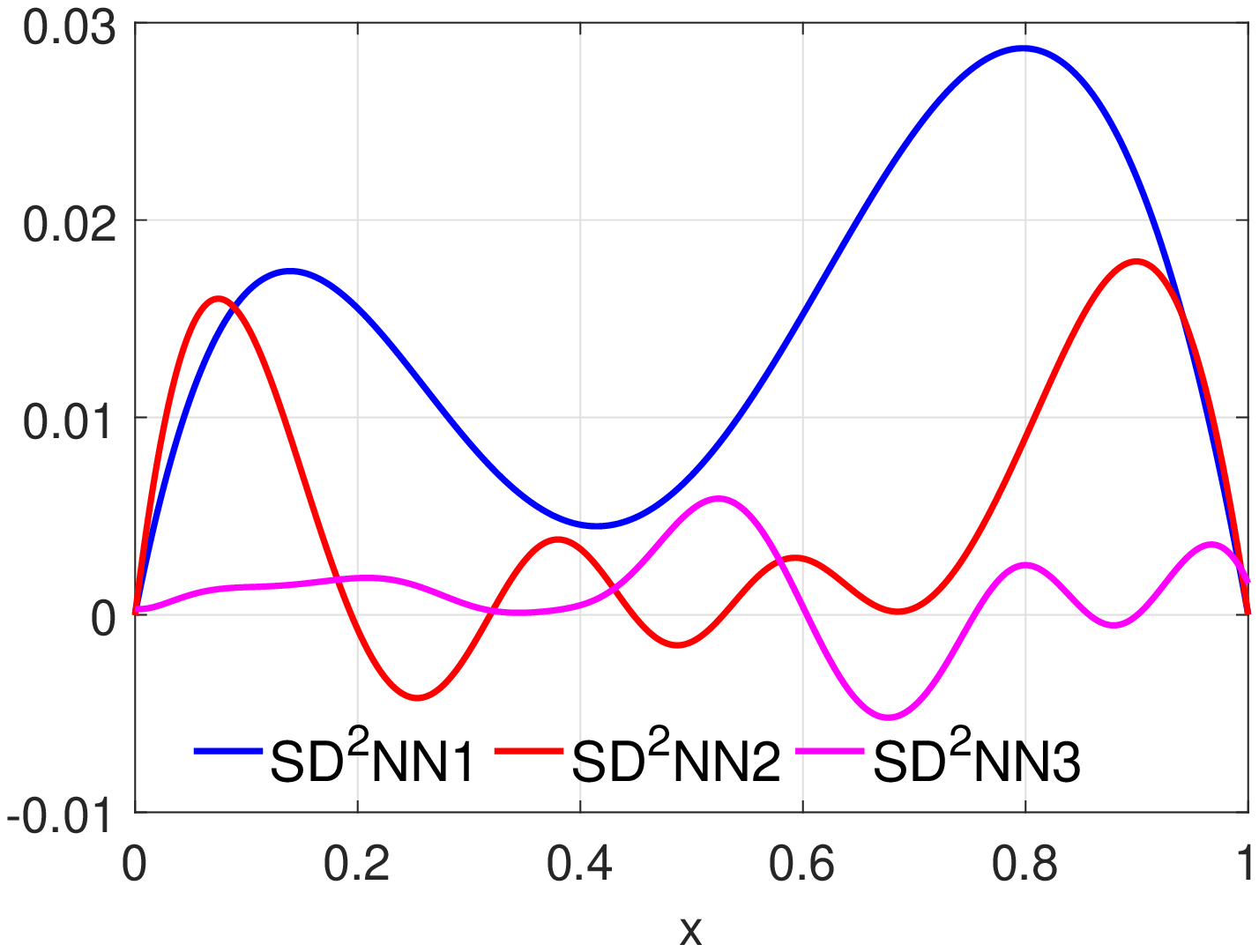}
		}
		\subfigure[difference of fine solutions]{
			\label{pLaplaceE1001b}
			\includegraphics[scale=0.325]{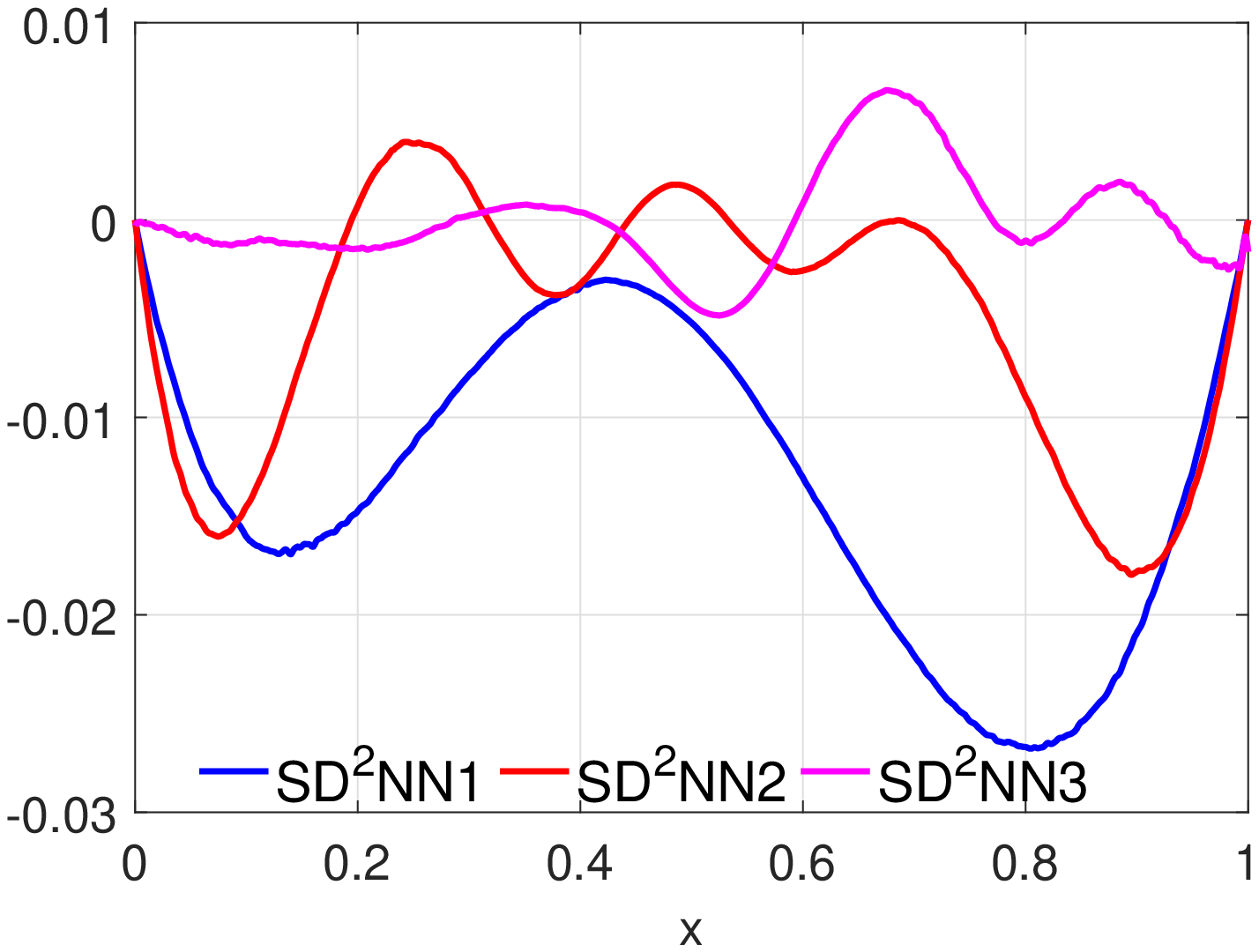}
		}
		\caption{Testing results when $\epsilon=0.01$  for Example \ref{DiffusionEq_1d_01}. }
		\label{pLaplaceE1001}
	\end{figure}

	We use MscaleDNNs and SD$^2$NNs with aforementioned setups to solve \eqref{DiffusionEq} when $\epsilon=0.1$ and $\epsilon=0.01$, respectively. For comparison, a normal DNN model with \emph{tanh} activation function (denoted by DNN) is also employed to solve this multi-scale problem.  In SD$^2$NN models, the balance parameter $\alpha$ for fine-part is set as 0.01. The number of parameters are comparable for different models (see Table \ref{Network_size2DiffusionEq_1d} in Appendix \ref{appendixA}). At each training step, we randomly sample 3000 interior points and 500 boundary points to evaluate the loss function. All models are trained for 60000 epochs. In the testing step, we uniformly sample 1000 points in [0, 1]. We show the testing results for $\epsilon=0.1$ and $\epsilon=0.01$ in Figures \ref{pLaplaceE101} and \ref{pLaplaceE1001}, respectively.
	
	\begin{table}[H]
		\centering
		\caption{ The relative error of different models for Example \ref{DiffusionEq_1d_01}when $\alpha=0.01$.}
		\label{Table_1DpLaplace}
		\begin{tabular}{|l|c|c|c|c|c|c|}
			\hline
			&DNN      &Mscale      &WWP     &SD$^2$NN1      &SD$^2$NN2       & SD$^2$NN3    \\  \hline
			{$\epsilon=0.1$} &2.40e-2  &3.42e-6     &3.84e-6      &1.36e-5    &6.22e-7     &7.81e-7   \\  \hline
			$\epsilon=0.01$  &2.05e-2  &6.95e-3     &1.94e-2      &2.80e-4    &7.69e-7     &3.30e-5   \\
			\hline
		\end{tabular}
	\end{table}
	
	From Figures \ref{pLaplaceE101c} and \ref{pLaplaceE1001c}, we observe that all three SD$^2$NN models capture the exact solutions with $\epsilon=0.1$ and $\epsilon=0.01$ pretty well. By comparison, the performances of Mscale and WWP are okay for the case $\epsilon=0.1$, but deteriorate for the more oscillatory case $\epsilon=0.01$. The normal DNN model fails in both cases. Figures \ref{pLaplaceE101d}, \ref{pLaplaceE101e}, \ref{pLaplaceE1001d} and \ref{pLaplaceE1001e} and the relative errors in Table \ref{Table_1DpLaplace} further demonstrate that SD$^2$NN2 is the best method. For the $\epsilon=0.01$ case, the error of SD$^2$NN2 is smaller than other methods by at least two orders of magnitude. 
	
	We choose the coarse part of the exact solution $u(x)$ as $u^c(x):= x-x^2$, and the fine part $u^f(x)$ as the remainder. Though it is not unique to choose the fine (and coarse) parts, they are ``equivalent" if the difference is smooth. We draw the differences of coarse solutions of SD$^2$NN1, SD$^2$NN2 and SD$^2$NN3 with $u^c(x)$ in Figures \ref{pLaplaceE101a} and \ref{pLaplaceE1001a},  and the differences of fine solutions with $u^f(x)$ in Figures \ref{pLaplaceE101b} and \ref{pLaplaceE1001b}. 
	The differences are smooth, which shows that SD$^2$NN models can capture the correct coarse and fine components of the solution.
	
\begin{figure}[H]
		\centering
		\subfigure[difference of coarse solutions]{
			\label{pLaplaceE1001a0.05}
			\includegraphics[scale=0.325]{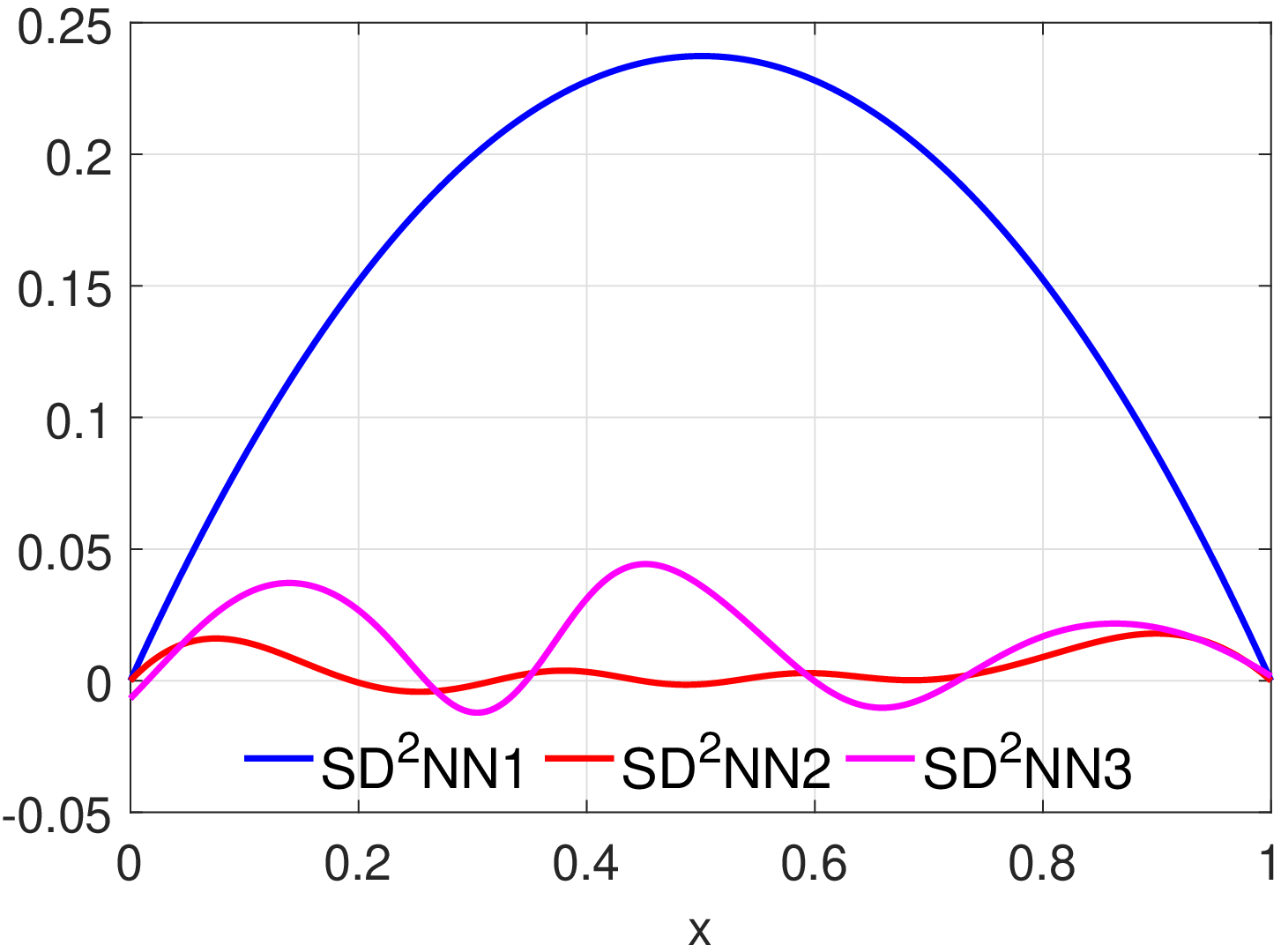}
		}
		\subfigure[difference of fine solutions]{
			\label{pLaplaceE1001b0.05}
			\includegraphics[scale=0.325]{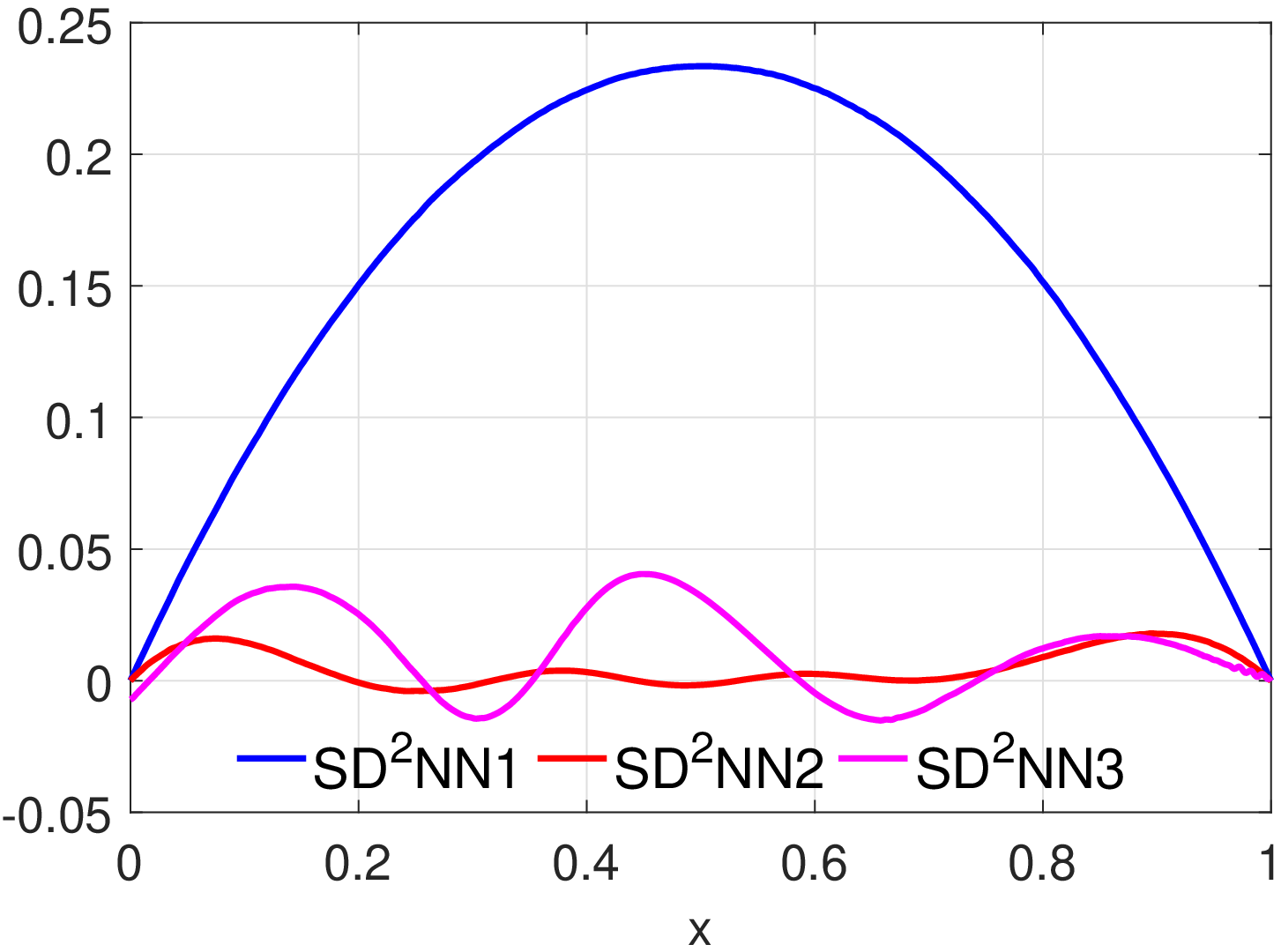}
		}
		\subfigure[difference of exact and DNN solutions]{
			\label{pLaplaceE1001c.05}
			\includegraphics[scale=0.325]{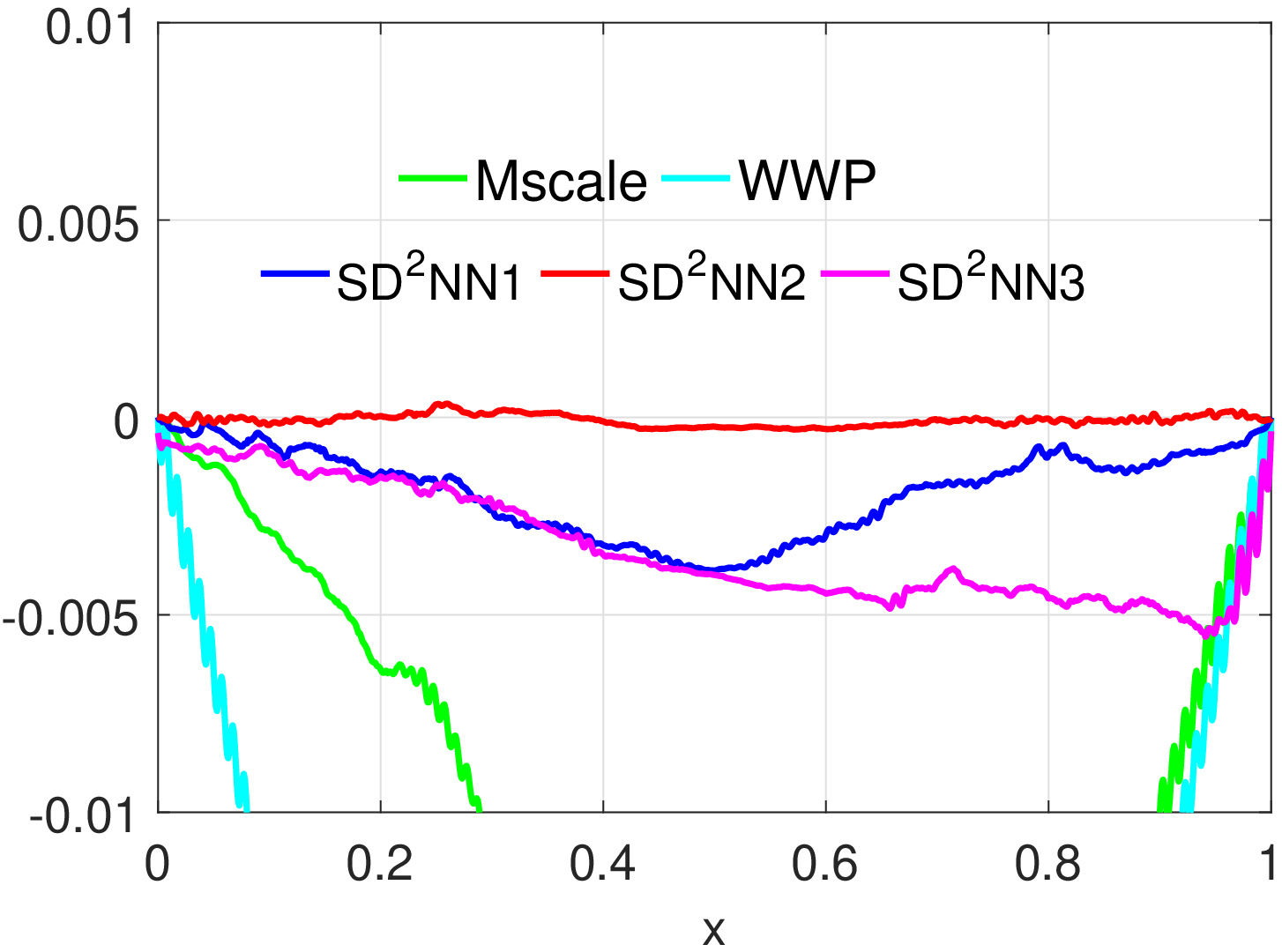}
		}
		\caption{Testing results when $\epsilon=0.01$ for Example \ref{DiffusionEq_1d_01}. }
		\label{pLaplaceE10010.05}
	\end{figure}

	\emph{Influence of hyper-parameter $\alpha$:} we study the influence of $\alpha$ for SD$^2$NN model. In the test, we set $\alpha=0.05$, $\epsilon=0.01$, keep all other parameters fixed, then train all models for 60000 epochs. Based on the results in Figure \ref{pLaplaceE10010.05}, we observe that SD$^2$NN2 is more stable and accurate compared to all other models. 
	
\begin{figure}
		\centering
		\includegraphics[scale=0.55]{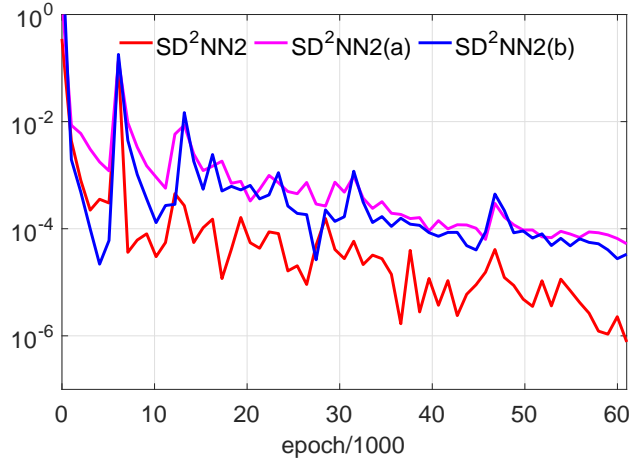}
		\caption{Relative error for SD$^2$NN2 when $\epsilon=0.01$ for Example \ref{DiffusionEq_1d_01}. }
		\label{Study2s}
	\end{figure}

	\emph{Influence of relaxation-parameter $s$:} we study the influence of relaxation-parameter $s$ for activation function \emph{SFM} in \emph{SD$^2$NN} model, especially for \emph{SD$^2$NN2}. Here, we list the setup of \emph{SD$^2$NN2} in Table \ref{comparison}, as well as two alternative setups: 
	\begin{itemize}
		\item \emph{SD$^2$NN2}:  SFM(1.0)+tanh for coarse submodule and  SFM(0.5)+S2ReLU for fine submodule;
		\item \emph{SD$^2$NN2}(a): SFM(0.5)+tanh for coarse submodule and  SFM(0.5)+S2ReLU for fine submodule; 
		\item \emph{SD$^2$NN2}(b): SFM(1.0)+tanh for coarse submodule and  SFM(1.0)+S2ReLU for fine submodule.
	\end{itemize}
	we set $\alpha=0.05$, $\epsilon=0.01$, and train both models for 60000 epochs. From the results in Figure \ref{Study2s}, it seems for the fine submodule of SD$^2$NN2, a relaxed parameter $s=0.5$ offers better performance.
	
	\textbf{Nonlinear Case:} 
	
	We further consider the nonlinear case with $p=8$ for \eqref{DiffusionEq} with $\Omega =[0, 1]$ and $A(x)=\left(2+\cos\left(2\pi\frac{x}{\epsilon}\right)\right)^{-1}$. By choosing appropriate $f(x)$, the (unique) solution is given by 
	\begin{equation}\label{Example1d_01_3}
	u(x) = x-x^2+\epsilon\left(\frac{1}{4\pi}\sin\left(2\pi\frac{x}{\epsilon}\right)-\frac{1}{2\pi}x\sin\left(2\pi\frac{x}{\epsilon}\right)-\frac{\epsilon}{4\pi^2}\cos\left(2\pi\frac{x}{\epsilon}\right)+\frac{\epsilon}{4\pi^2}\right).
	\end{equation}
	with $u_\epsilon(0)=u_\epsilon(1)=0$. 
	
	\begin{figure}[H]
		\centering
		\subfigure[difference of exact and DNN solutions]{
			\label{Boltzm_02a}
			\includegraphics[scale=0.4]{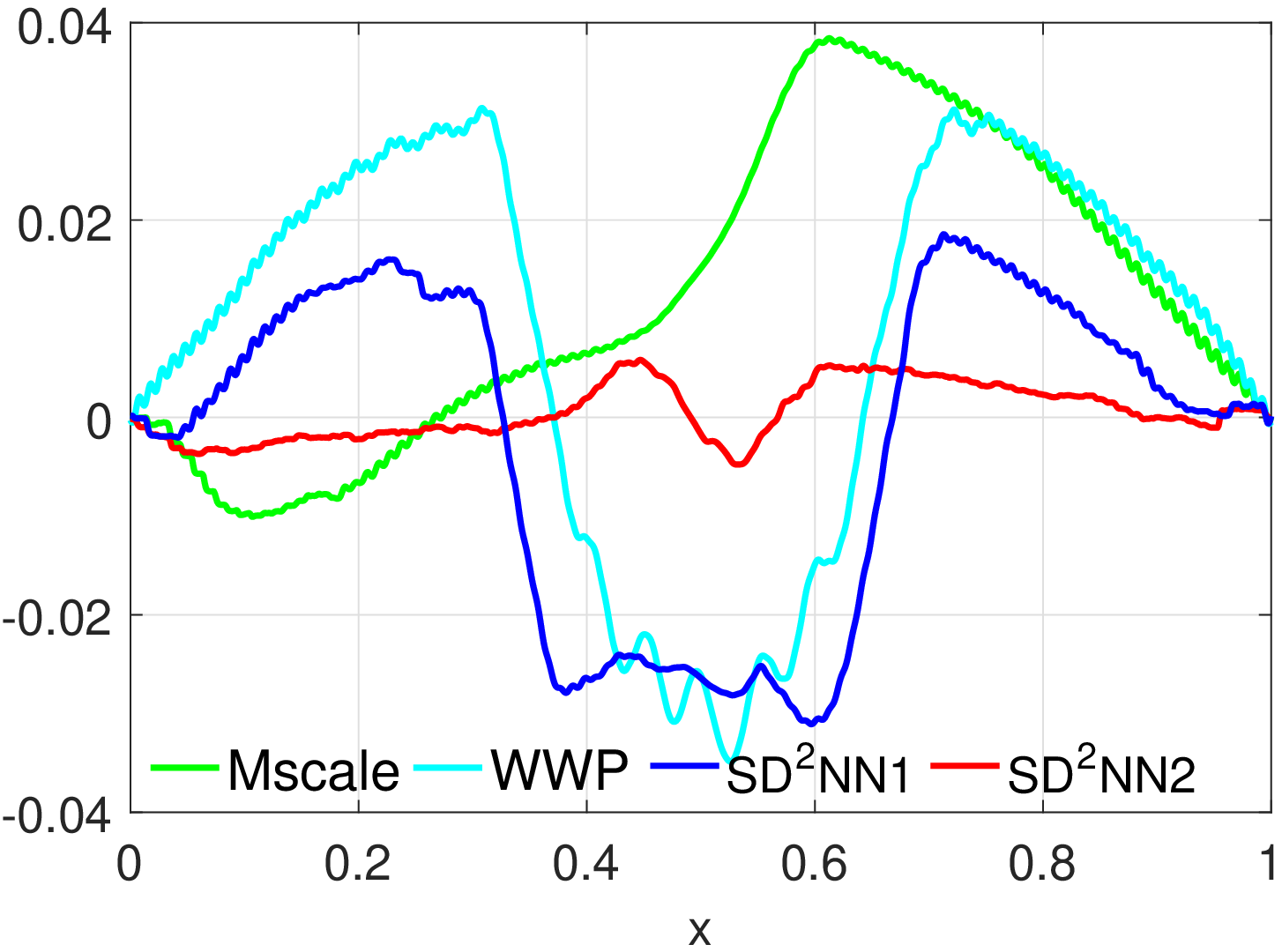}
		}
		\subfigure[relative error for different models]{
			\label{Boltzm_02d}
			\includegraphics[scale=0.4]{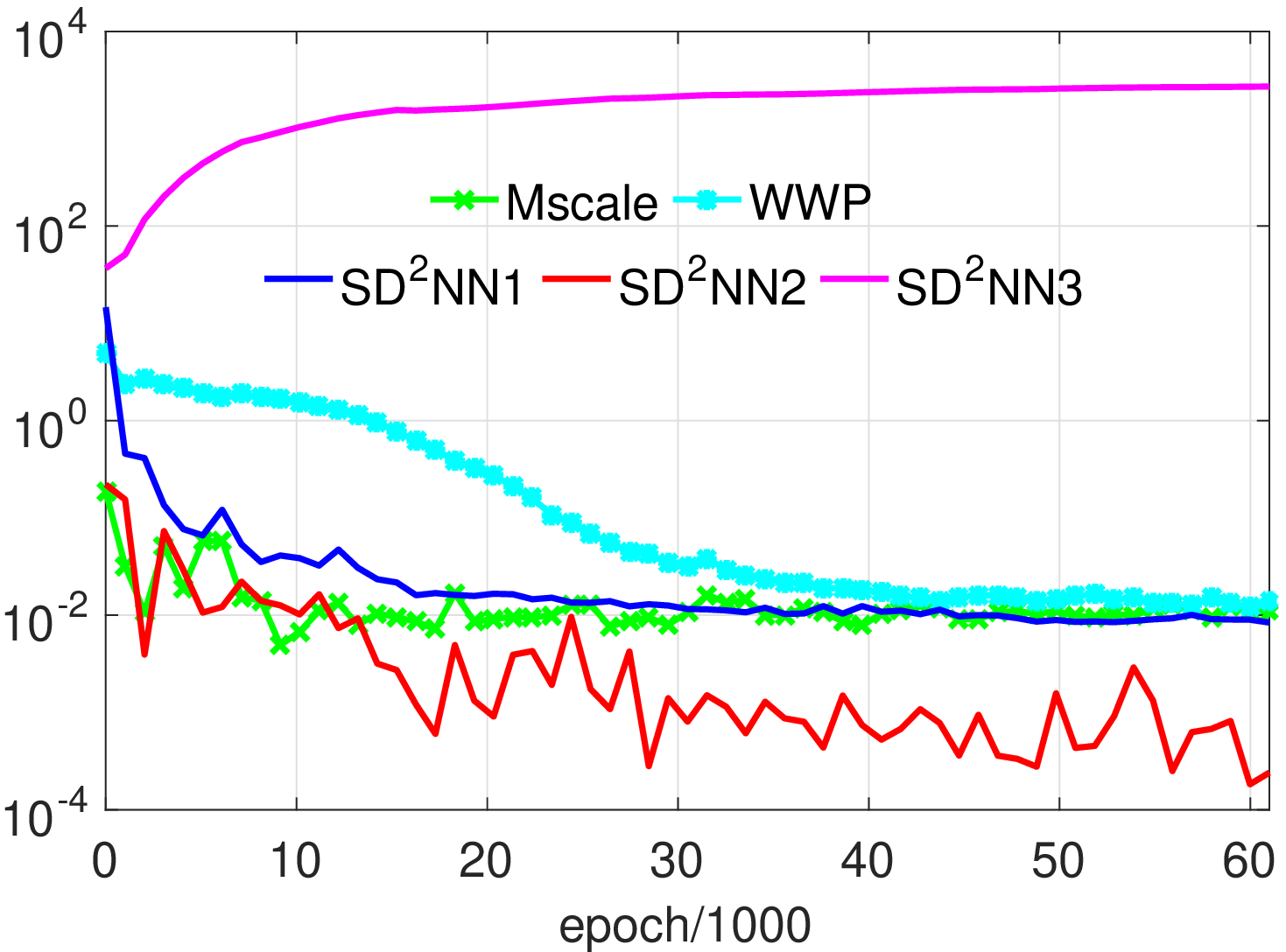}
		}
		\subfigure[difference of coarse solution]{
			\label{Boltzm_02e}
			\includegraphics[scale=0.4]{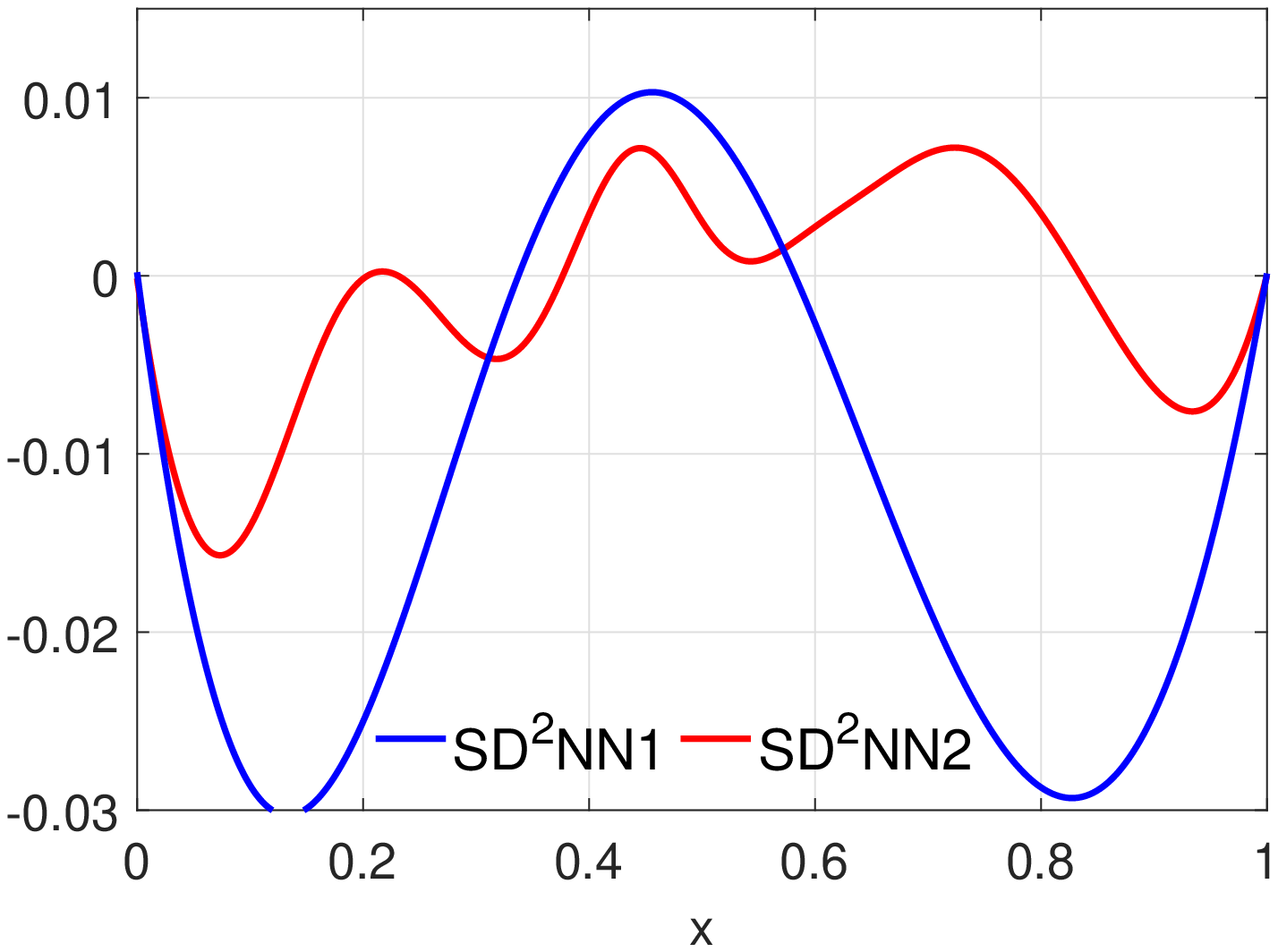}
		}
		\subfigure[difference of fine solution]{
			\label{Boltzm_02f}
			\includegraphics[scale=0.4]{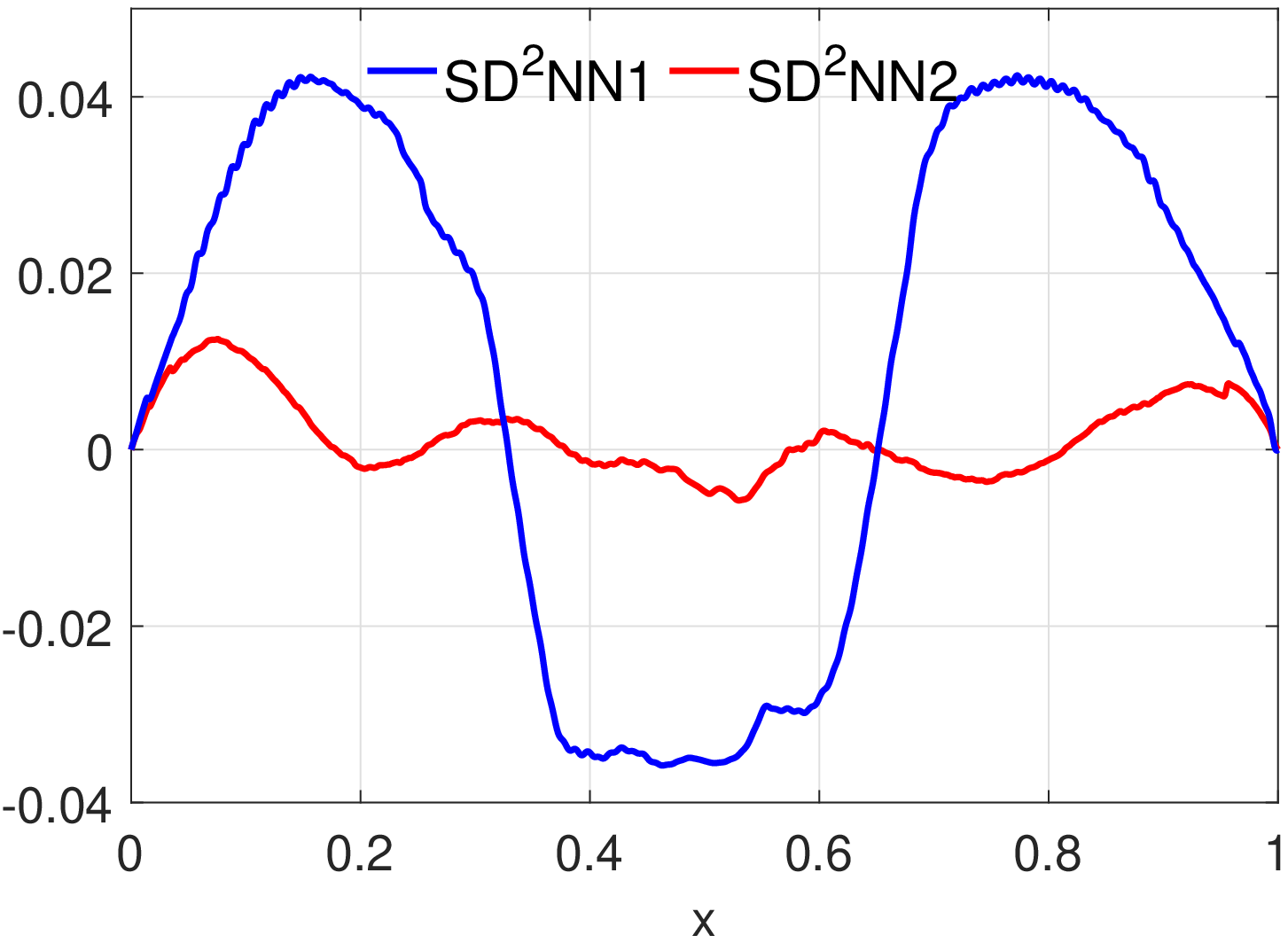}
		}
		\caption{Testing results for $\epsilon=0.01$ and $p=8$. }
		\label{BoltzmE1001}
	\end{figure}
	
	We employ Mscale, WWP and SD$^2$NN2 models to solve \eqref{DiffusionEq}. Based on the results in Figure \ref{BoltzmE1001}, the SD$^2$NN2 model still outperform Mscale,  WWP, SD$^2$NN1 and SD$^2$NN3 (fails to converge) for nonlinear multi-scale problem. In addition, the differences of coarse and fine solutions in Figures \ref{Boltzm_02e} and \ref{Boltzm_02f}  show that the submodules of SD$^2$NN2 also can well capture the coarse and fine parts of the multi-scale solution.
	
	\begin{example}[Three-scale problem]\label{Diffusion1D3scale}
		We solve the problem \eqref{DiffusionEq} with the following three-scale coefficient
		\begin{equation}\label{3scaleA}
		A(x) = \left(2+\cos\left(2\pi\frac{x}{\epsilon_1}\right)\right)\left(2+\cos\left(2\pi\frac{x}{\epsilon_2}\right)\right)
		\end{equation}
		with two small parameter $1\gg\epsilon_1\gg\epsilon_2>0$, also for $p=2$ and $\Omega = [0,1]$. We impose the exact solution 
		\begin{equation}\label{Diffusion1D3scaleU}
		u(x) = x-x^2+\frac{\epsilon_1}{4\pi}\sin\left(2\pi\frac{x}{\epsilon_1}\right) +\frac{\epsilon_2}{4\pi}\sin\left(2\pi\frac{x}{\epsilon_2}\right).
		\end{equation}
		with $u(0)=u(1)=0$, such that $f(x)$ can be obtained by direct computation. 
	\end{example}
	
	We employ the aforementioned Mscale, WWP, SD$^2$NN1, SD$^2$NN2 and SD$^2$NN3 to solve \eqref{DiffusionEq} with \eqref{3scaleA} with $\epsilon_1=0.1$ and $\epsilon_2=0.01$. We use 2 MscaleDNN submodules in the SD$^2$NN models, and the balance parameters  $\alpha_1=0.1$ and $\alpha_2=0.01$ for the mesoscale submodule and the fine submodule, respectively. Table \ref{Network_size2Diffusion_3scale} in Appendix \ref{appendixA} shows that the number of parameters for different models are comparable. All models are trained for 60000 epochs. For each training step, we randomly sample 3000 interior points and 500 boundary points as the training data, and uniformly sample 1000 points in [0,1] as the testing data. 
	\begin{figure}[H]
		\centering
		\subfigure[difference for exact solution and approximate solutions]{
			\label{pLaplace3Scale:solu}
			\includegraphics[scale=0.45]{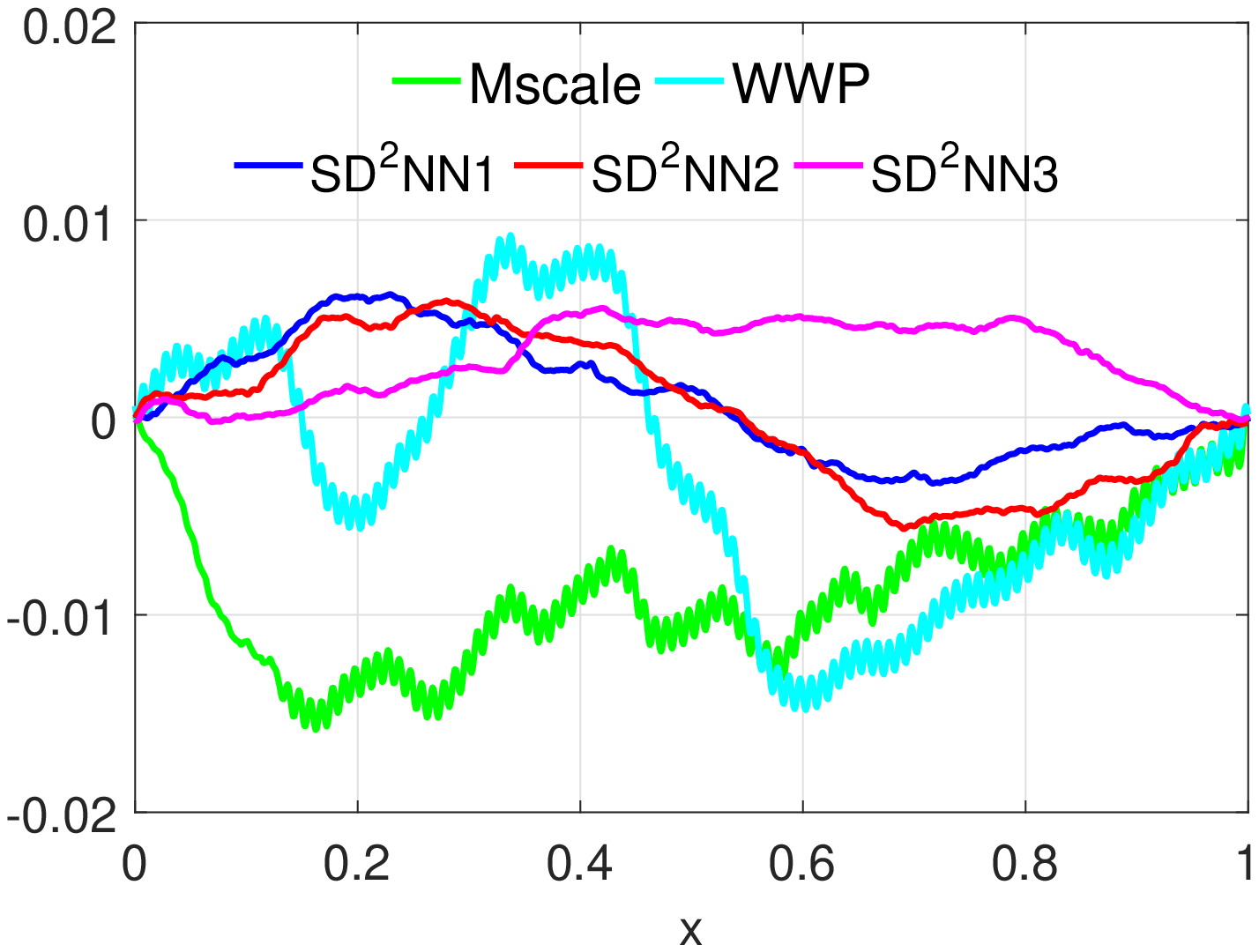}
		}
		\subfigure[coarse part in the frequency domain]{
			\label{pLaplace3Scale:fft2coarse}
			\includegraphics[scale=0.45]{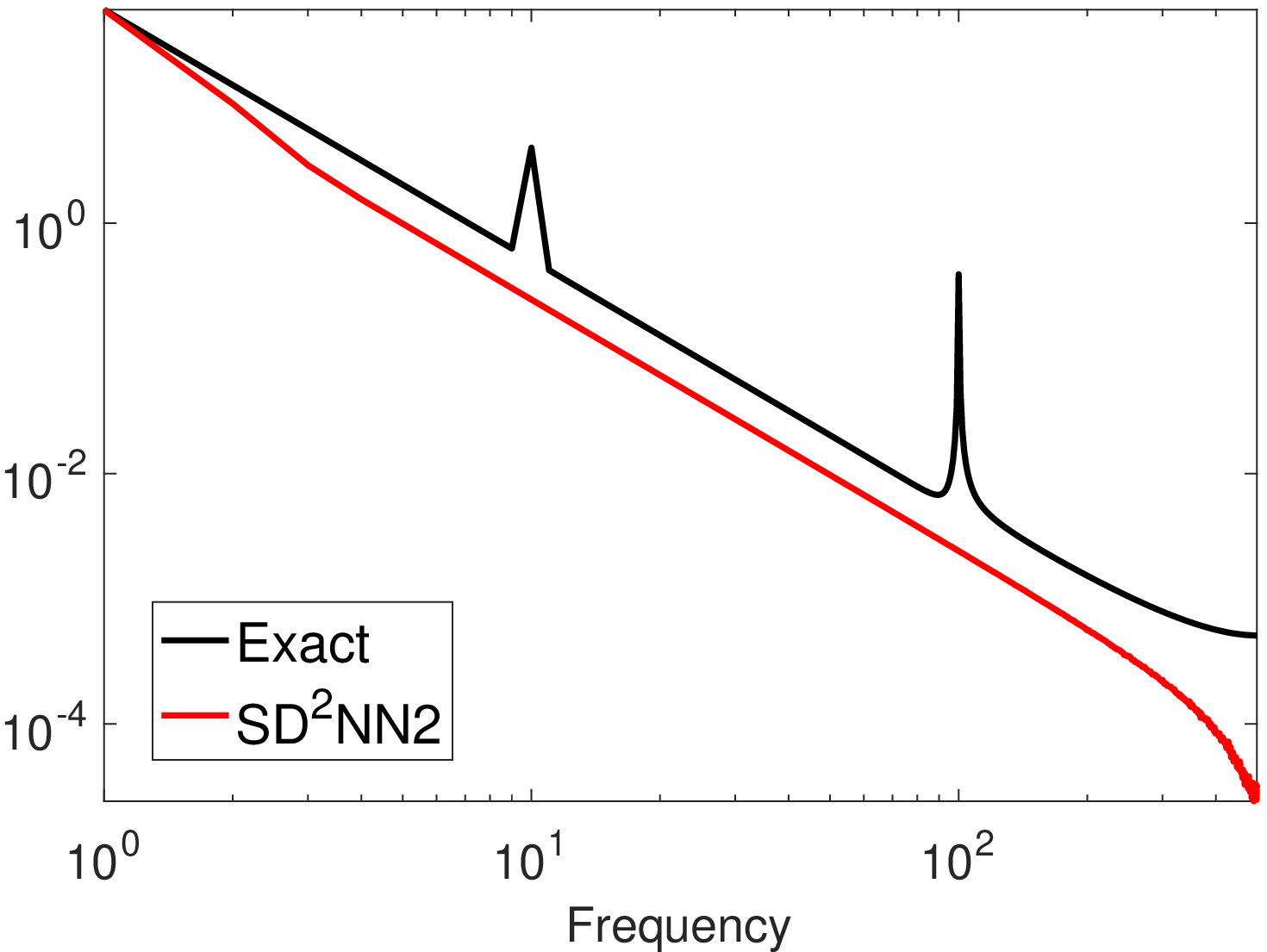}
		}
		\subfigure[mesoscale part in the frequency domain]{
			\label{pLaplace3Scale:fft2meso}
			\includegraphics[scale=0.45]{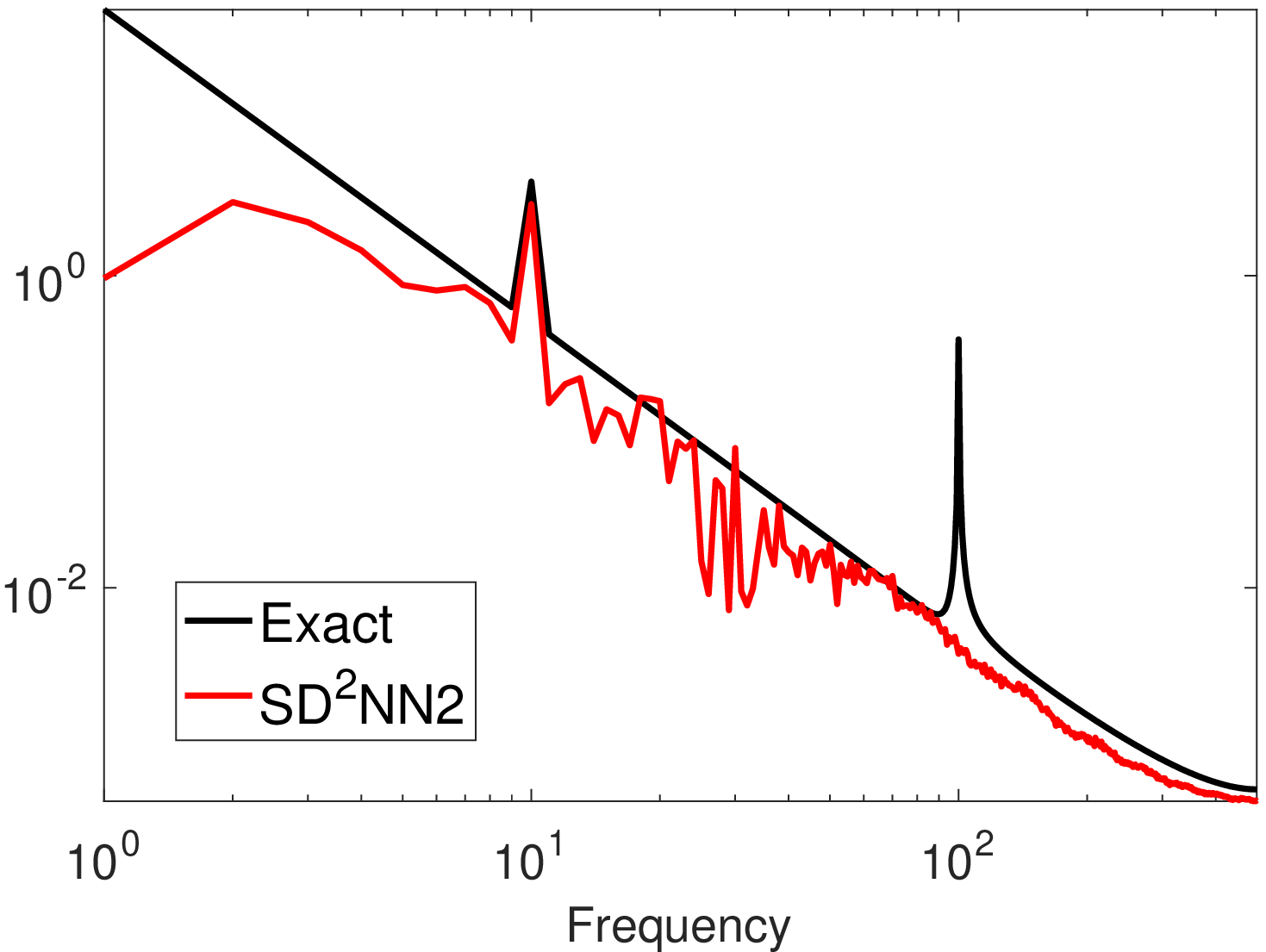}
		}
		\subfigure[finescale part in the frequnecy domain]{
			\label{pLaplace3Scale:fft2fine}
			\includegraphics[scale=0.45]{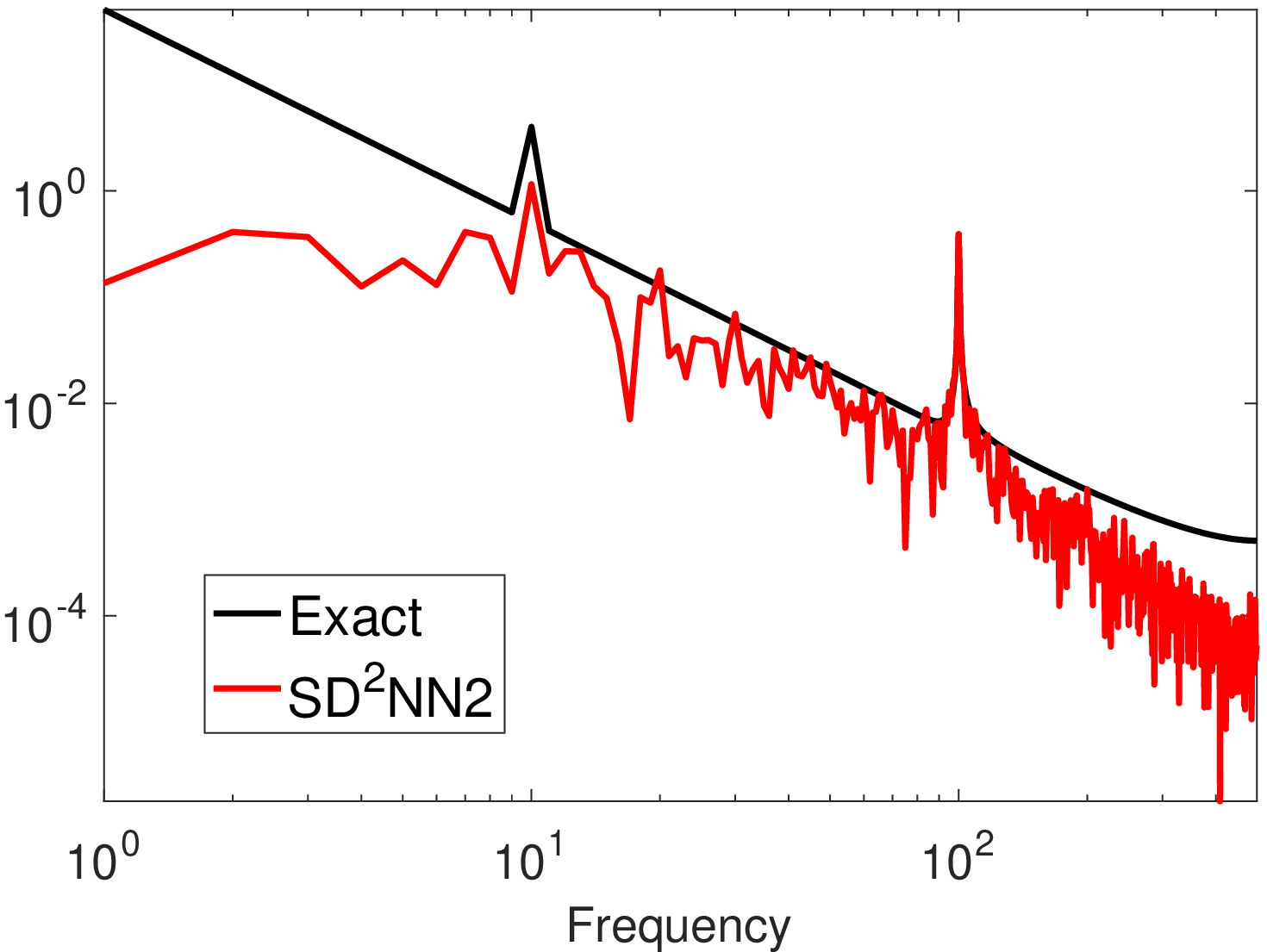}
		}
		\caption{Numerical results for Example \ref{Diffusion1D3scale}}
		\label{pLaplace3Scale}
	\end{figure}
	
	\begin{table}[H]
		\centering
		\caption{ The relative error of different models for Example \ref{Diffusion1D3scale}}
		\label{Table_1DpLaplace_3scale}
		\begin{tabular}{|c|c|c|c|c|}
			\hline
			Mscale     &WWP   &SD$^2$NN1    &SD$^2$NN2      &SD$^2$NN3           \\  \hline
			2.70e-3    &1.53e-3    &1.25e-3  &6.75e-5    &1.55e-4    \\  \hline
		\end{tabular}
	\end{table}
	In Figure \ref{pLaplace3Scale:solu}, we draw the difference of the approximate solution and the numerical solution, together with the relative errors from Table \ref{Table_1DpLaplace_3scale}, it shows that SD$^2$NN2 is the best method. In Figures \ref{pLaplace3Scale:fft2coarse} -- \ref{pLaplace3Scale:fft2fine} we compare the coarse, mesoscale, fine scale parts of the SD$^2$NN2 solution with the exact solution, in the frequency domain. We demonstrate that three submodules of the SD$^2$NN2 model can correctly capture the corresponding components of the solution as they are supposed to be.
	
	From now on, we will only consider the Mscale, WWP and SD$^2$NN2 models in the following examples.
	
	\begin{example}\label{Diffusion2D}
		We consider the following two-dimensional problem \eqref{DiffusionEq} for $p=2$ and $\Omega = [-1,1]\times [-1,1]$. In this example, we set $f=1$ and the multi-scale coefficient
		\begin{equation*}
		A(x_1,x_2) =\Pi_{i=1}^{6} \bigg{(}1+0.5\cos\left(2^i\pi(x_1+x_2)\right)\bigg{)}\bigg{(}1+0.5\sin\left(2^i\pi(x_2-3x_1)\right)\bigg{)}.
		\end{equation*}
		from \cite{owhadi2007homogenization,owhadi2008numerical,owhadi2014polyharmonic}. The reference solution $u(x_1,x_2)$ can be computed by the finite element method on a square grid of mesh-size $h=1/129$.
	\end{example}
	
	We compute the solution of \eqref{DiffusionEq} by Mscale, WWP and SD$^2$NN2, respectively. In the SD$^2$NN2 model, the balance parameter $\alpha$ for fine-submodule is 0.05. The network size and parameter number for different models are comparable and listed in Table \ref{Network_size2DiffusionEq_2d} in Appendix \ref{appendixA}. All models are trained for 100000 epochs. At each training step, we sample 3000 interior points and 500 boundary points. The testing points are taken from the finite element grid with $h=1/129$.
	\begin{figure}[H]
		\centering    
		\subfigure[point-wise error for Mscale] { 
			\label{Diffusion2d_E4:scale}     
			\includegraphics[scale=0.3]{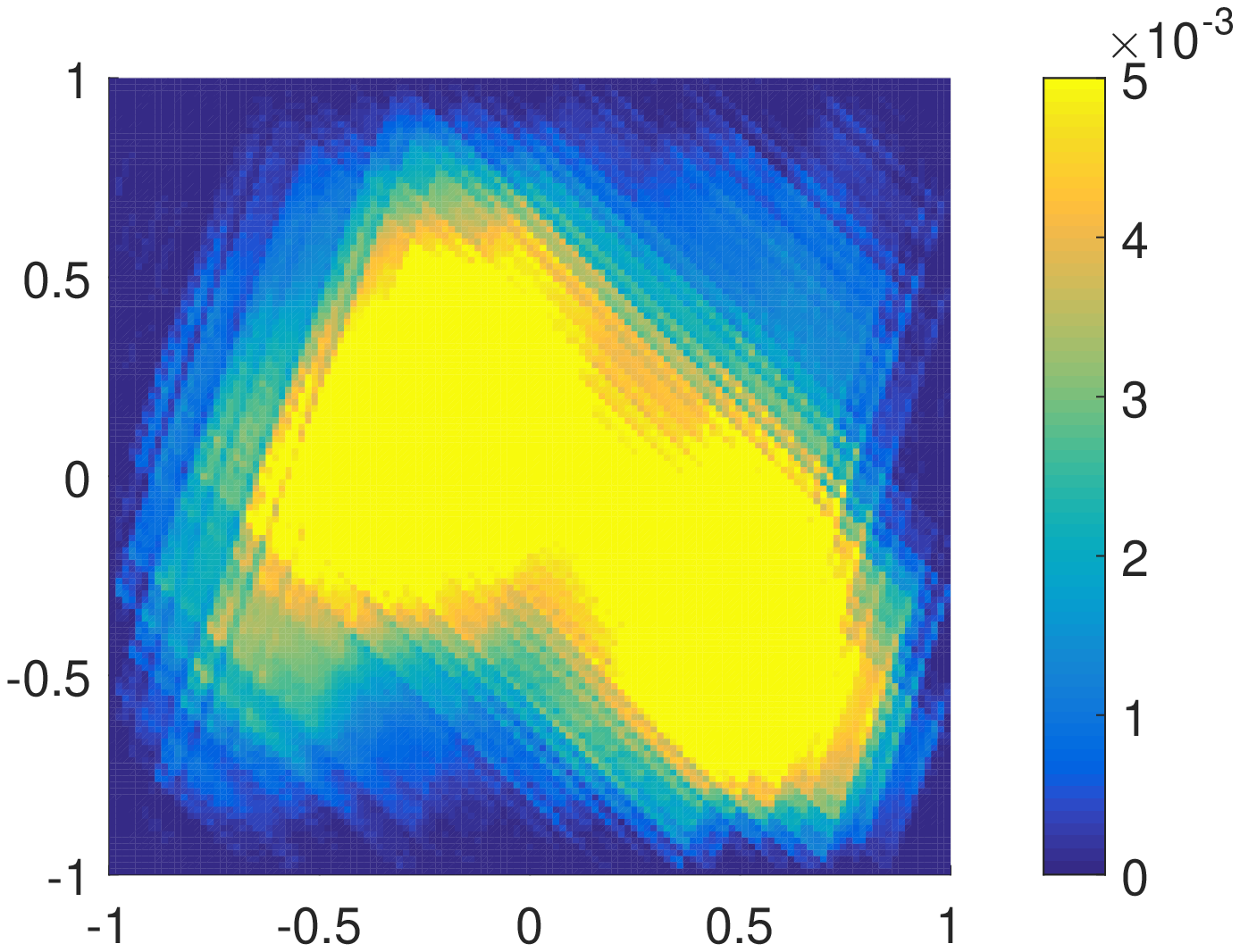}
		} 
		\subfigure[point-wise error for WWP] { 
			\label{Diffusion2d_E4:perr2Fourier}     
			\includegraphics[scale=0.3]{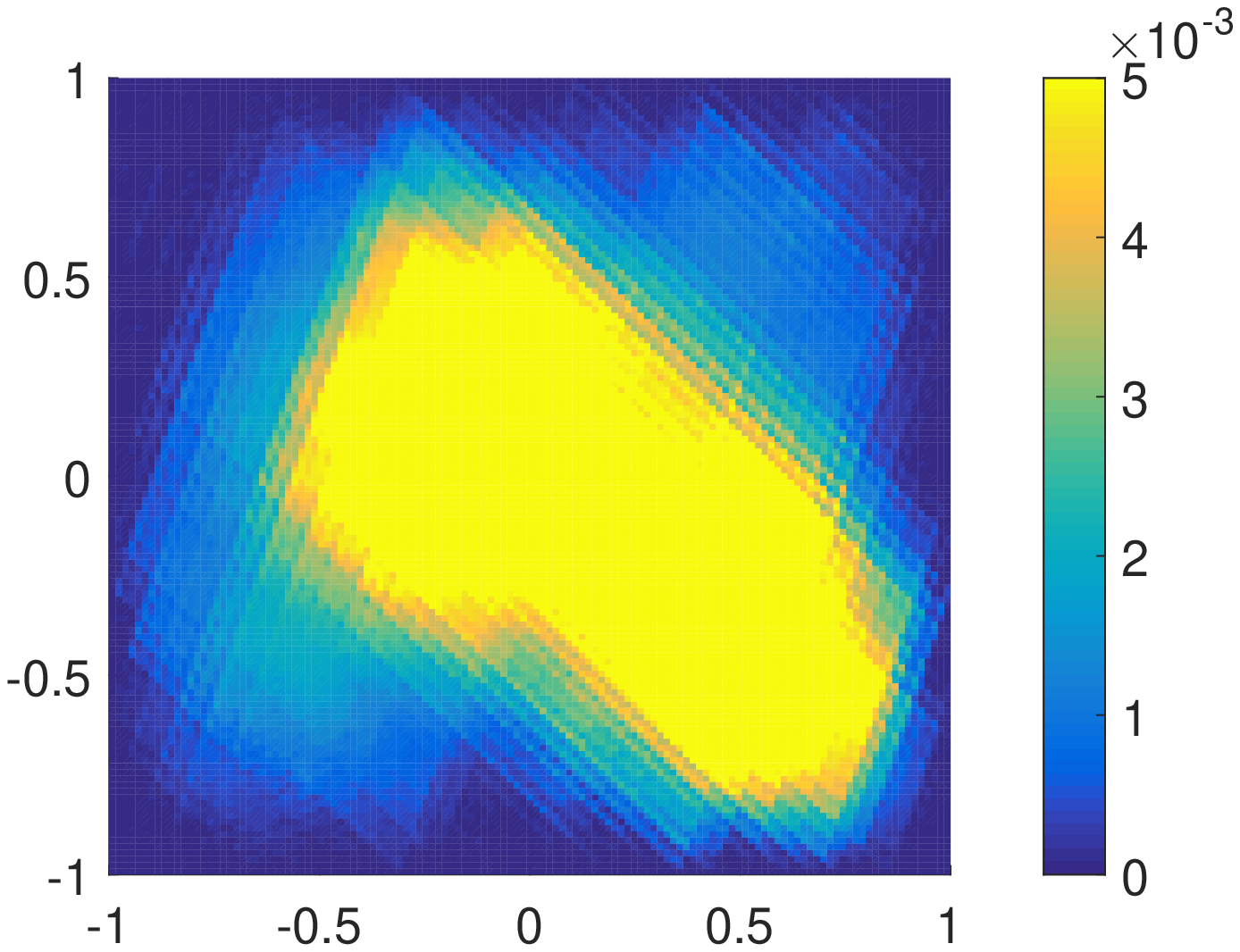}     
		}
		\subfigure[point-wise error for SD$^2$NN2] { 
			\label{Diffusion2d_E4:perr2SD$^2$NN}     
			\includegraphics[scale=0.3]{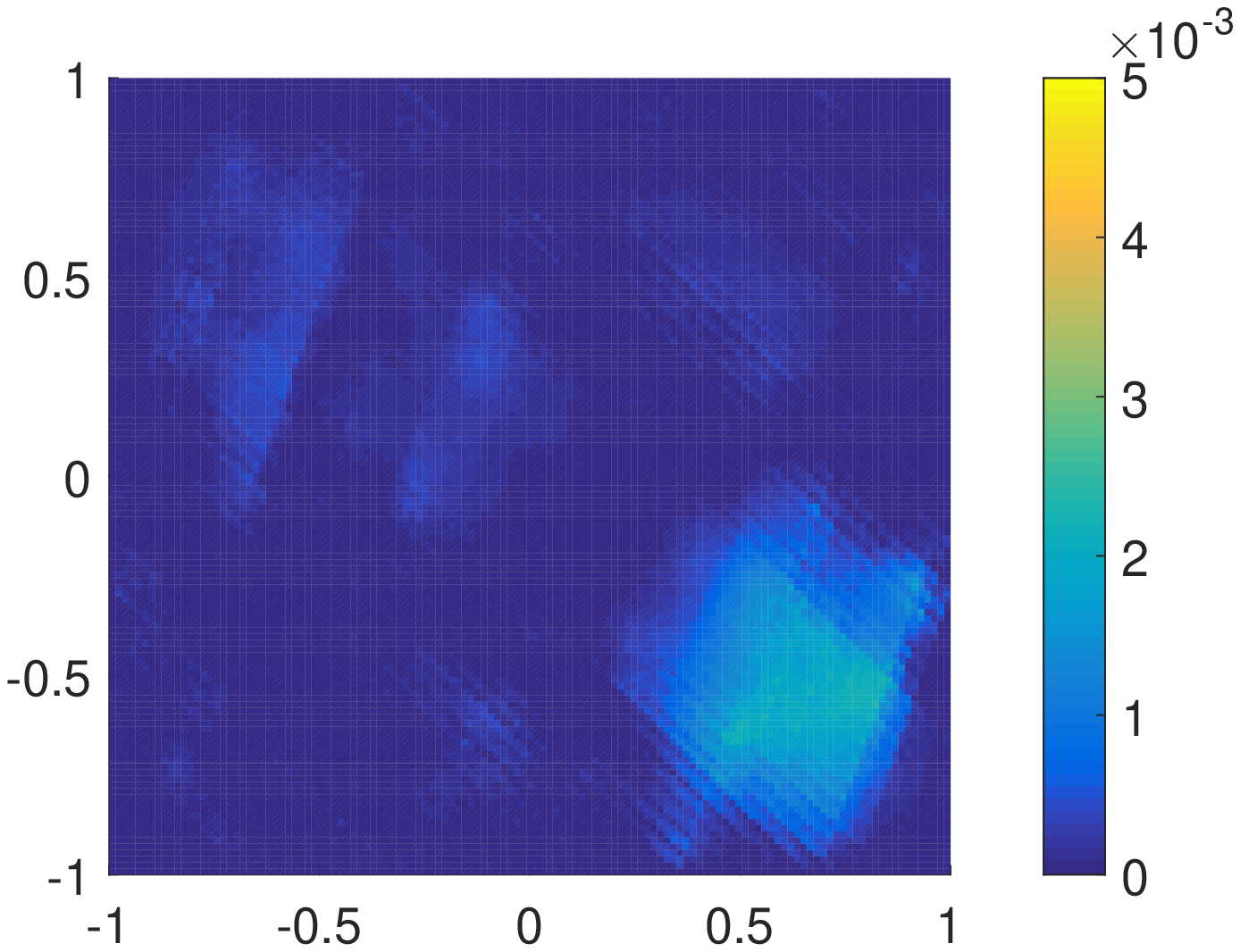}     
		}  
		\subfigure[relative error for Mscale, WWP and SD$^2$NN2] { 
			\label{Diffusion2d_E4:MSE_REL}     
			\includegraphics[scale=0.3]{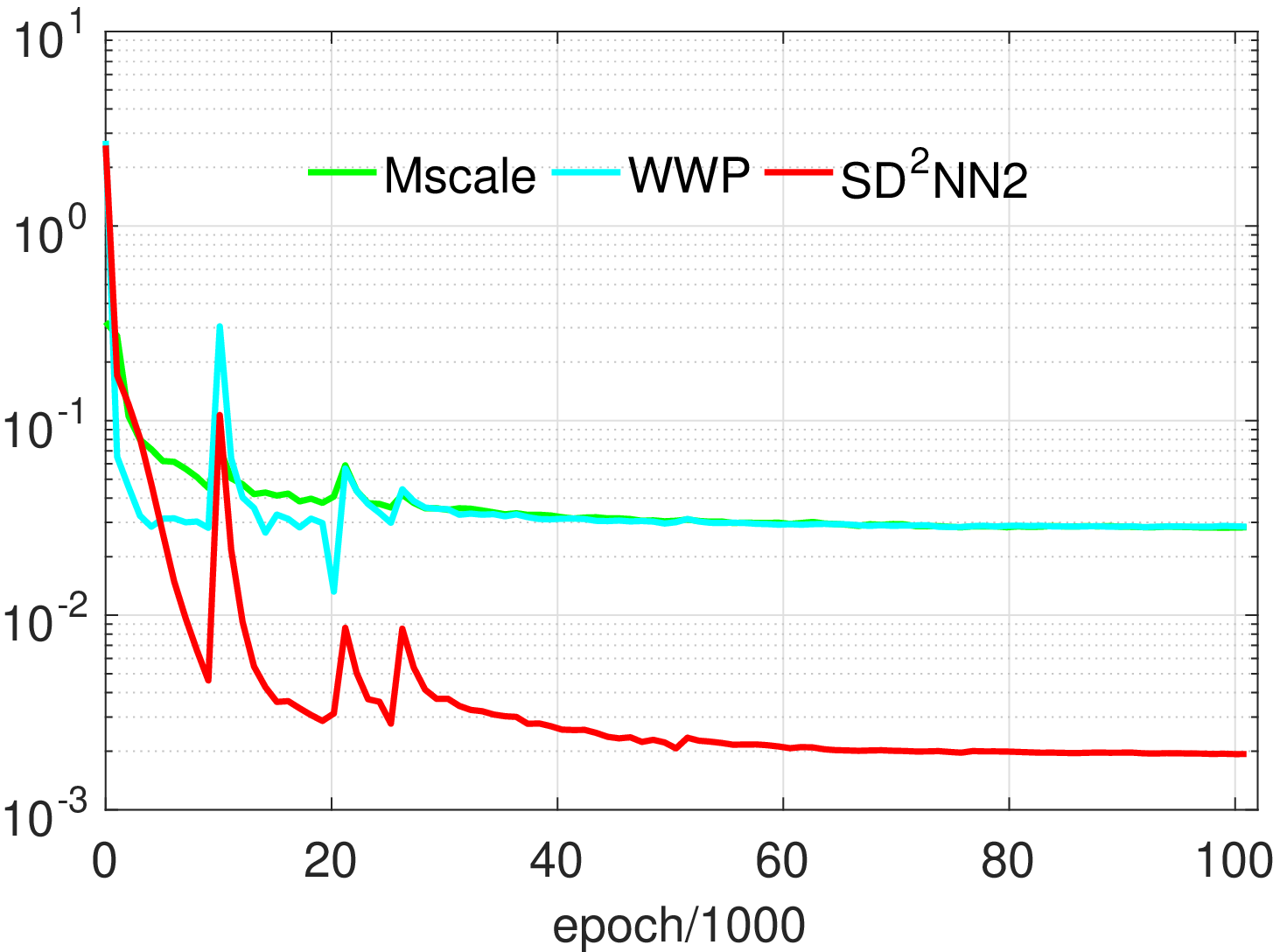}     
		} 
		\subfigure[SD$^2$NN2 coarse solution] {
			\label{Diffusion2d_E4:coarse2}     
			\includegraphics[scale=0.315]{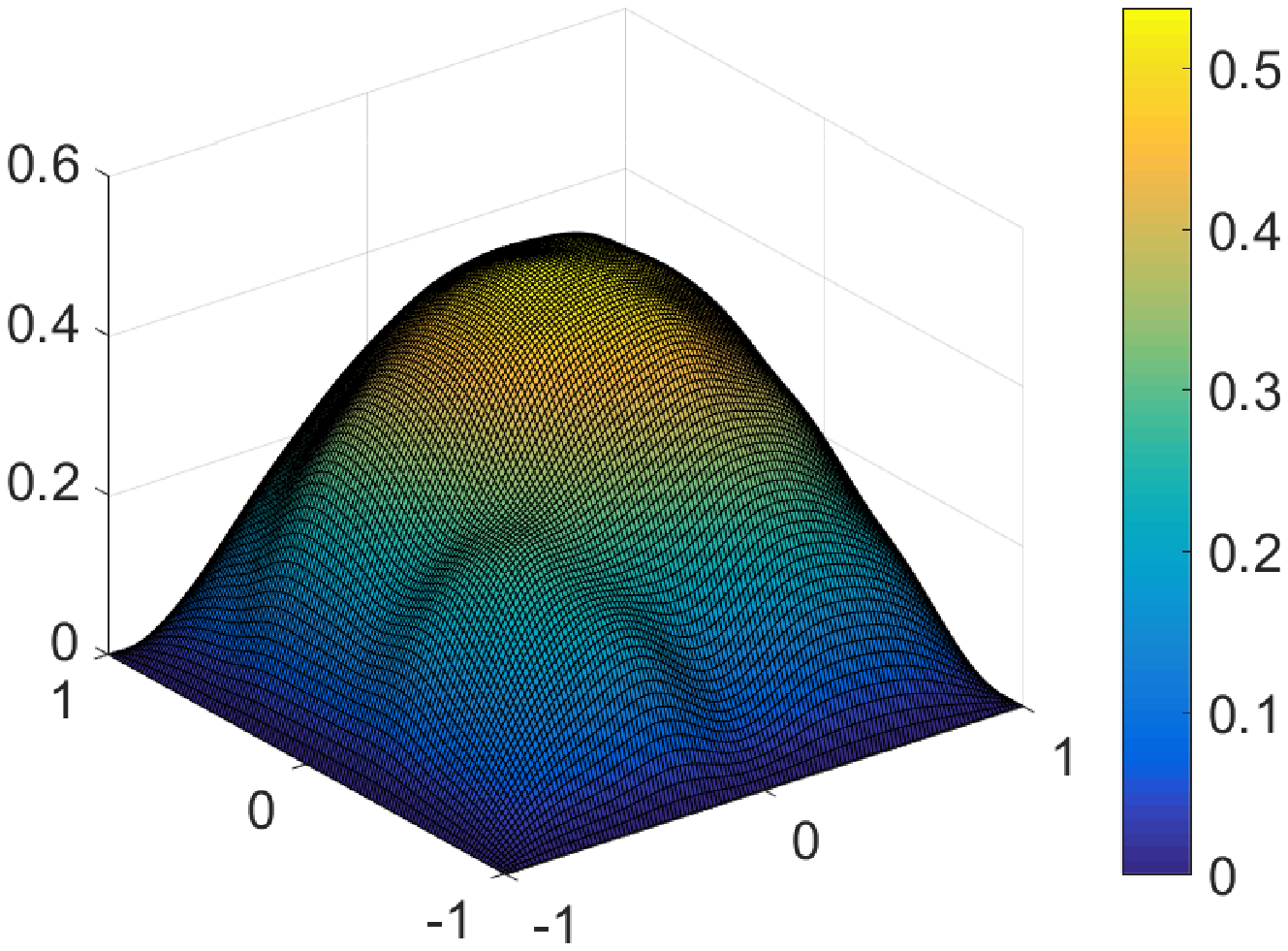}  
		}
		\subfigure[SD$^2$NN2 fine solution] {
			\label{Diffusion2d_E4:fine2}     
			\includegraphics[scale=0.315]{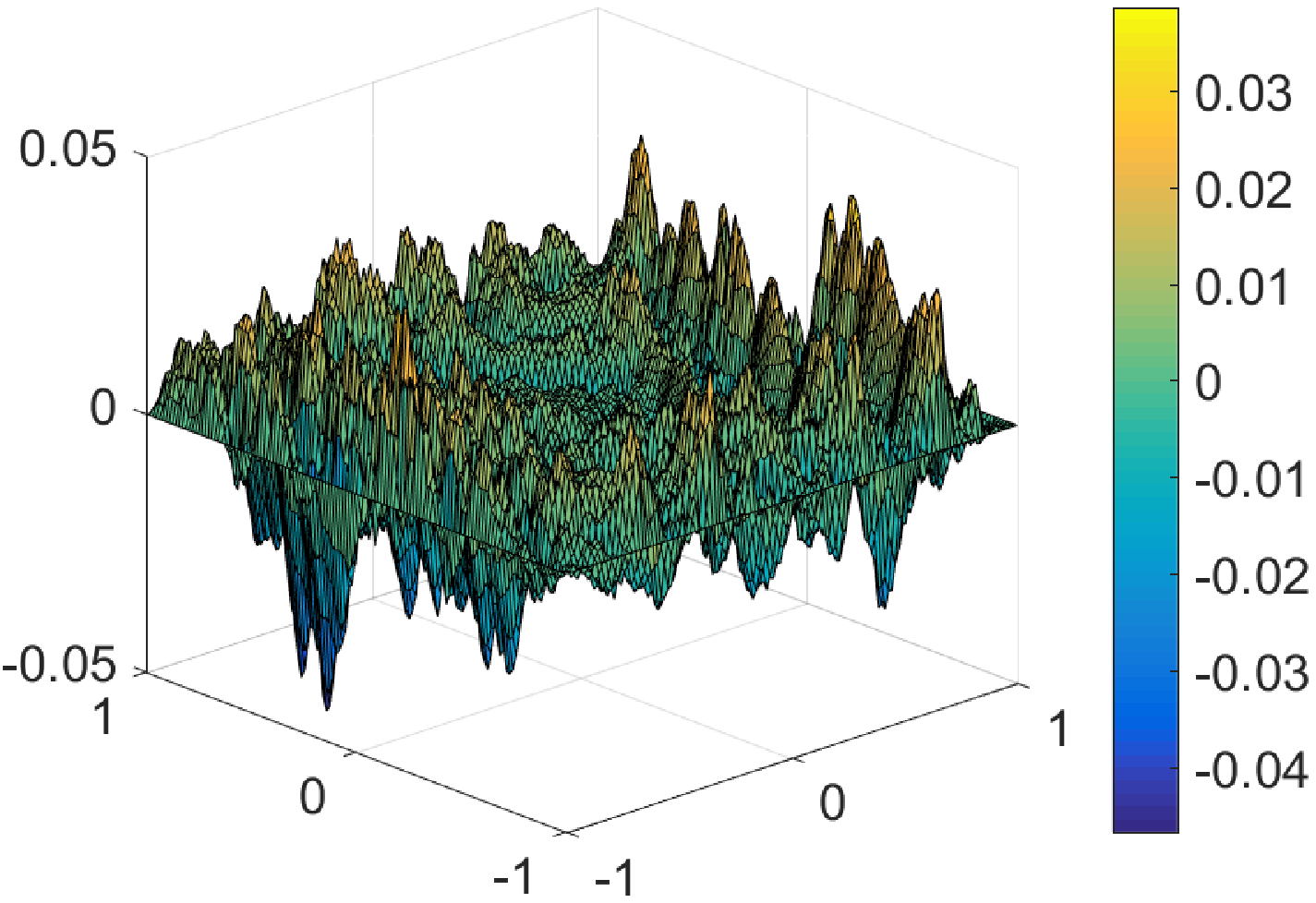}  
		}
		\caption{Testing results for Example \ref{Diffusion2D}} 
		\label{Diffusion2d_E4}         
	\end{figure}
	
	The numerical results in Figures \ref{Diffusion2d_E4:scale} -- \ref{Diffusion2d_E4:MSE_REL} show that the SD$^2$NN2 is still the method of choice in terms of both the point-wise error and the mean square error, and it is much better than Mscale and WWP models. In addition, the results in Figures \ref{Diffusion2d_E4:coarse2} and \ref{Diffusion2d_E4:fine2} show that the submodules of SD$^2$NN2 successfully separate the coarse and fine parts of the solution.

\begin{figure}
		\centering  
		\subfigure[Domain with holes] {
			\label{PB3dPDE:3D_domain}     
			\includegraphics[scale=0.325]{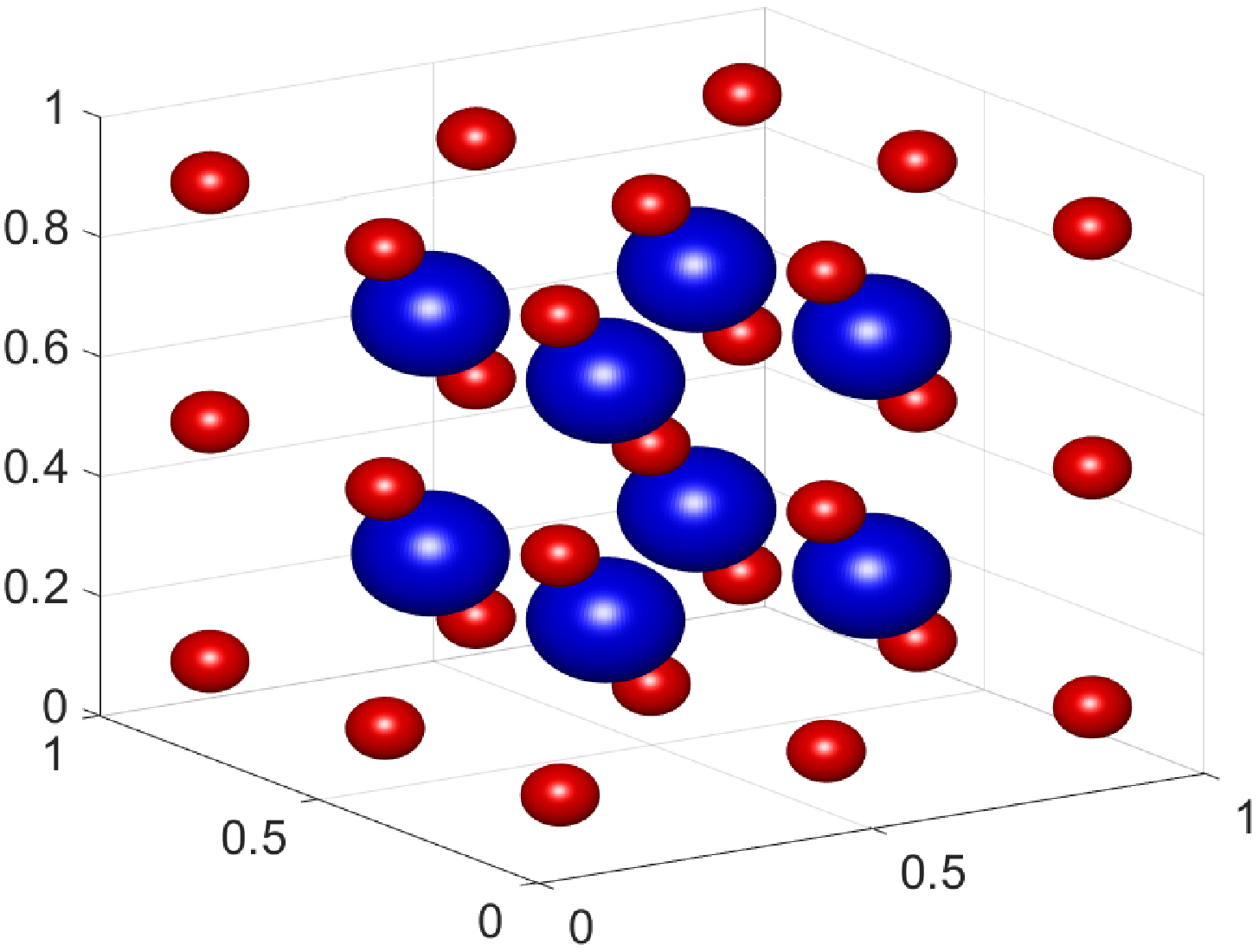} 
		}     
		\subfigure[point-wise error for Mscale] {
			\label{PB3dPDE:perr2scale}     
			\includegraphics[scale=0.325]{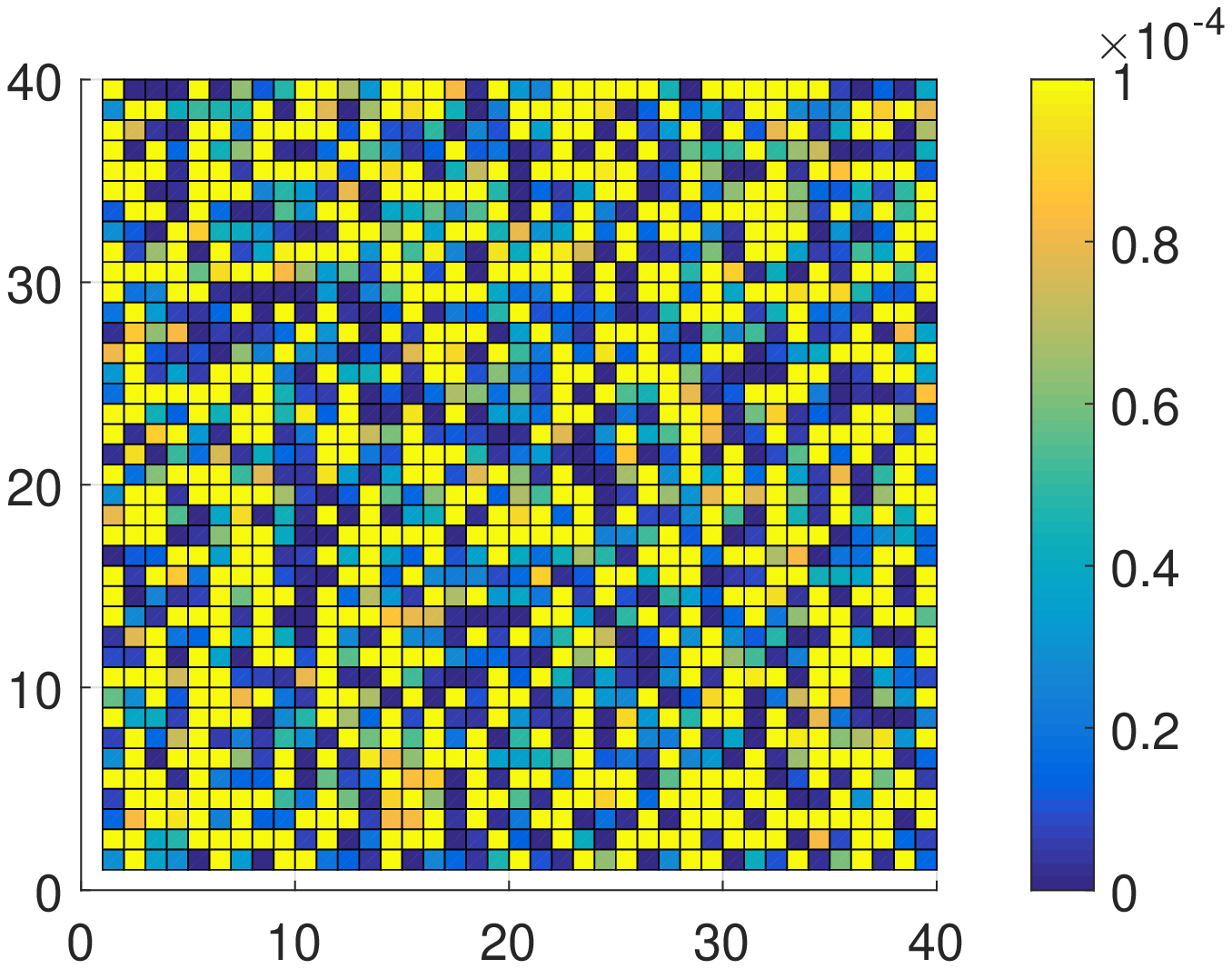} 
		}     
		\subfigure[point-wise error for WWP] { 
			\label{PB3dPDE:perr2Fourier}     
			\includegraphics[scale=0.325]{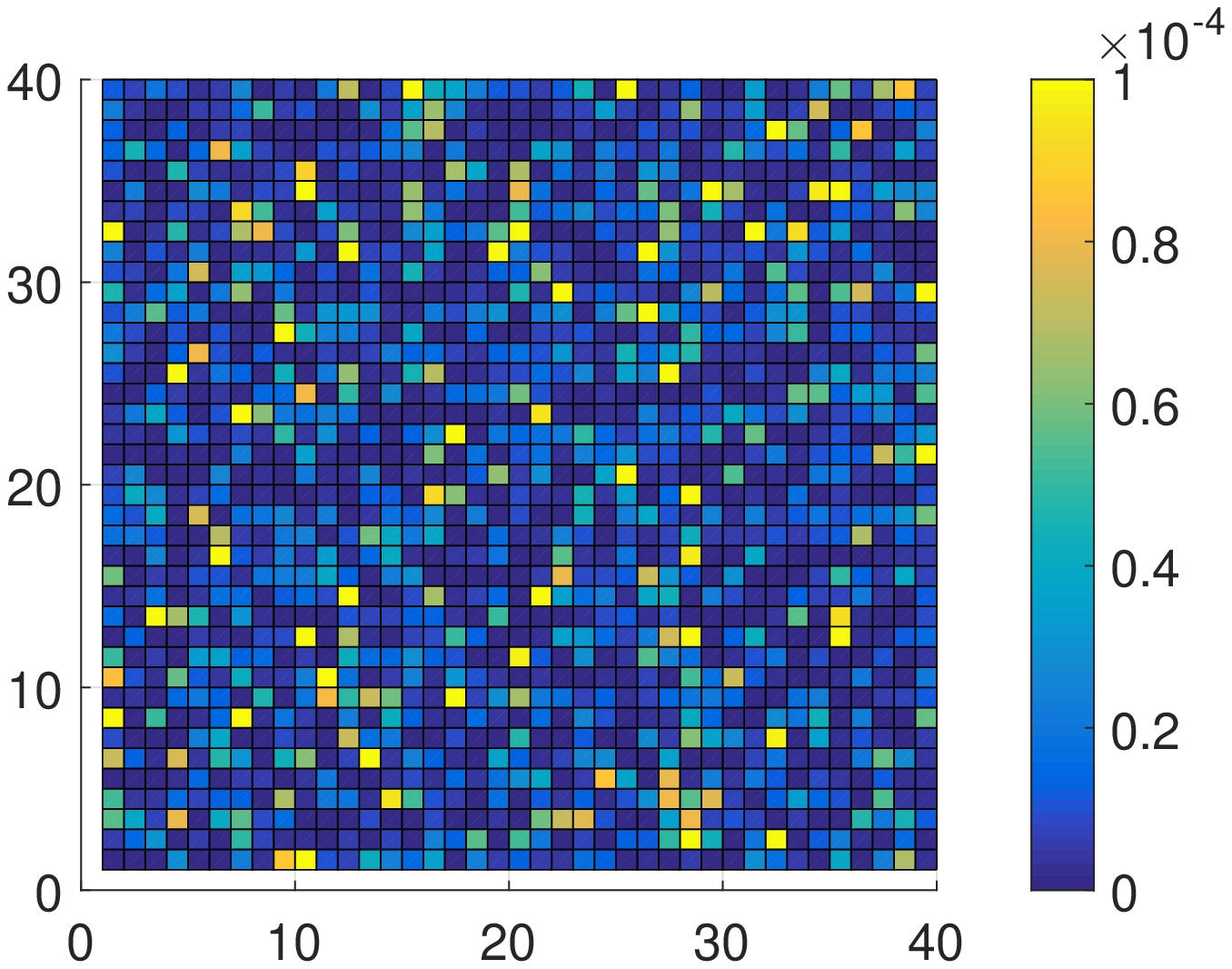}  
		} 
		\subfigure[point-wise error for SD$^2$NN2] { 
			\label{PB3dPDE:perr2SD$^2$NN2}     
			\includegraphics[scale=0.325]{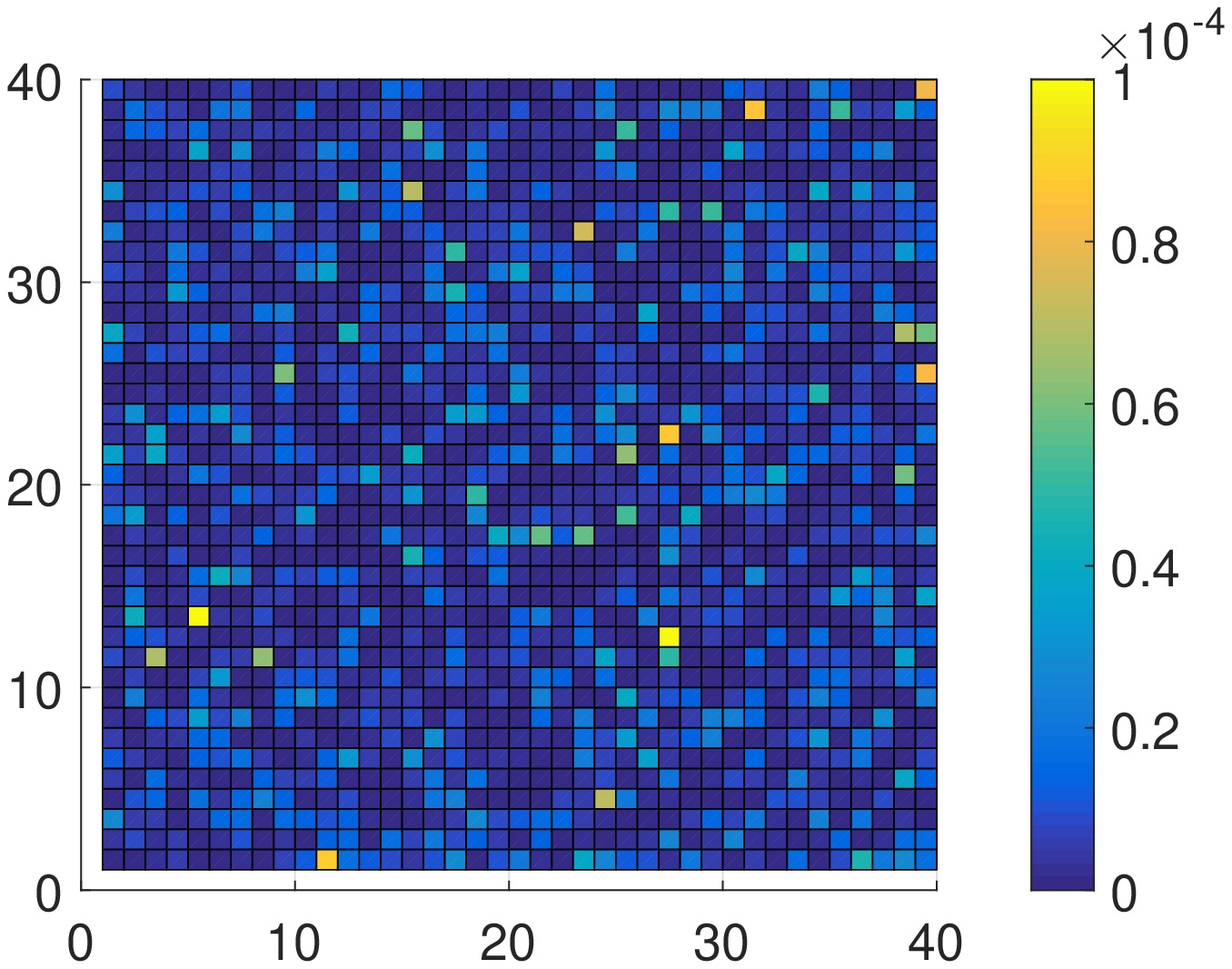}     
		}
		\subfigure[relative error of Mscale, WWP and SD$^2$NN2] { 
			\label{PB3dPDE:mse_rel}     
			\includegraphics[scale=0.325]{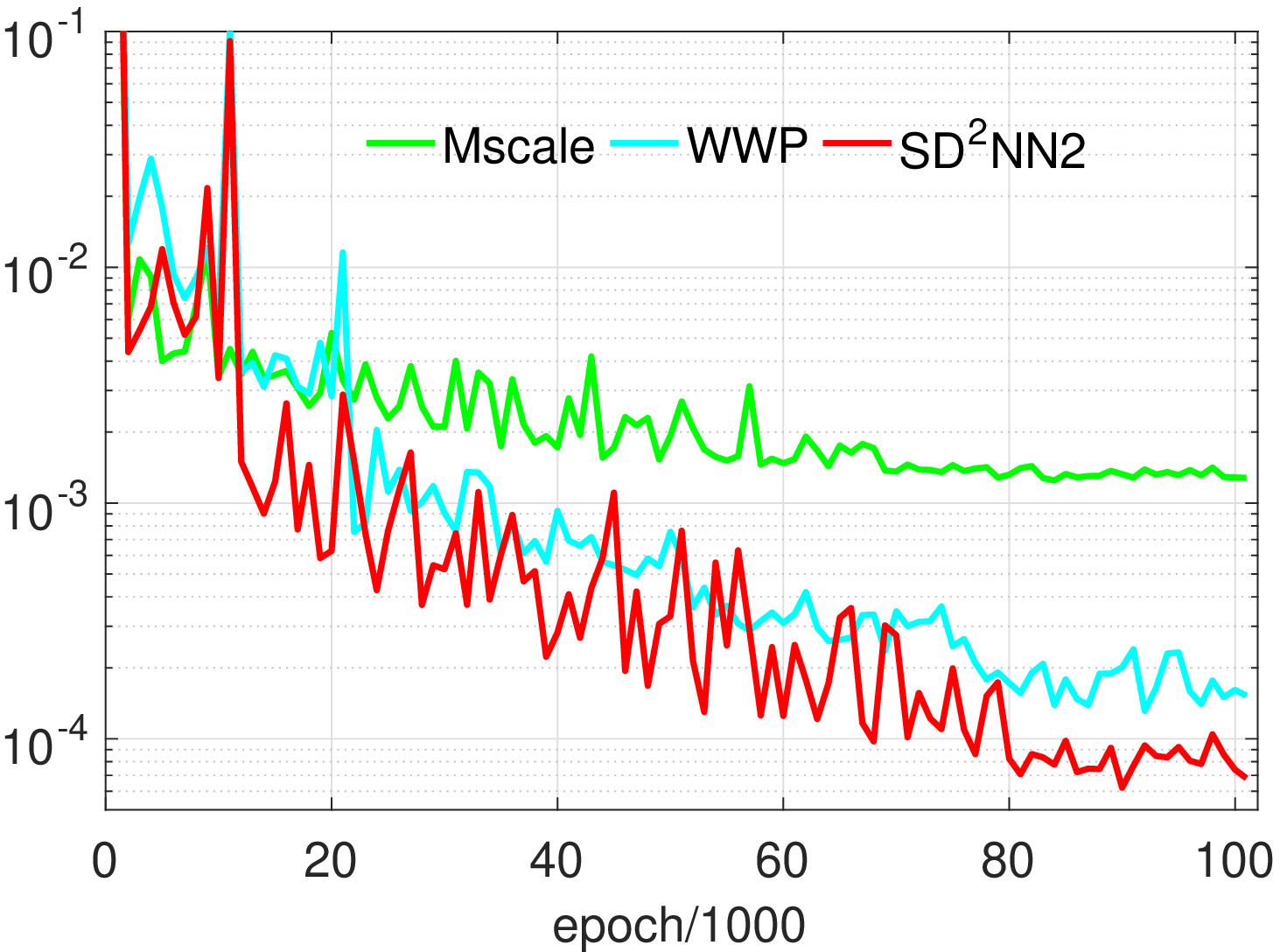}
		}
		\caption{Testing results for Example \ref{Boltzmann3D}. } 
		\label{PB3dPDE}         
	\end{figure}

	\begin{example}\label{Boltzmann3D}
		We now consider the following Poisson-Boltzmann equation with Dirichlet boundary condition,
		\begin{equation}\label{Boltzmann}
		\begin{cases}
		-\textbf{div}(A(\bm{x})\nabla u(\bm{x})) + \kappa(\bm{x})u(\bm{x}) = f(\bm{x}),~~\bm{x}\in \Omega \subset \mathbb{R}^d,\\
		~~~~~~~~~~~~~~~~~~  u(\bm{x}) = g(\bm{x}),  ~~~~~~~~~~~~~~~~~~~\bm{x}\in\partial \Omega
		\end{cases}
		\end{equation}
		where $A_{\epsilon}(\bm{x})$ is the dielectric constant and $\kappa(\bm{x})$ the inverse Debye-Huckel length of an ionic solvent. It is natural to introduce the energy
		\begin{equation}\label{Boltzmann-variational}
		\mathcal{J}(v) = \frac{1}{2}\int_{\Omega} \bigg{(}A|\nabla v|^2+\kappa v^{2}\bigg{)}d\bm{x} -\int_{\Omega}fvd\bm{x}.
		\end{equation}
		and the corresponding variational formulation.
		
		We solve the elliptic equation \eqref{Boltzmann} in the cube $\Omega=[0,1]^3$ with 8 big holes (blue) and 27 smaller holes (red), see Figure \ref{PB3dPDE}(a). We take $\kappa(x_1.x_2,x_3)=\pi^2$,  
		\begin{equation}\label{ExamplePB3d_A}
		A(x_1,x_2,x_3)=0.5\bigg{(}2+\cos(10\pi x_1)\cos(20\pi x_2)\cos(30\pi x_3)\bigg{)},
		\end{equation}
		and impose an exact solution
		\begin{equation}\label{ExamplePB2d_u}
		u(x_1,x_2,x_3)=\sin(\pi x_1)\sin(\pi x_2)\sin(\pi x_3) + 0.05\sin(10\pi x_1)\sin(20\pi x_2)\sin(30\pi x_3).
		\end{equation}
		The exact solution prescribes the corresponding boundary condition and $f(x)$ in \eqref{Boltzmann}.
	\end{example}
	
	In this example, the network sizes and the number of parameter for Mscale, WWP and SD$^2$NN2 are listed in Table \ref{Network_size2Boltzmann_3d} in Appendix \ref{appendixA}, their parameters are comparable. In addition, the balance parameter $\alpha=0.05$ for the fine-part of SD$^2$NN2. All models are trained for 100000 epochs. At each training step, the training data set including 6000 interior points and 1000 boundary points randomly sampled from $\Omega$ and $\partial \Omega$. The testing dataset is 1600 random samples in $\Omega$. Testing results are plotted in Fig.\ref{PB3dPDE}. 
	
	From the results in \ref{PB3dPDE:perr2scale} -- \ref{PB3dPDE:perr2SD$^2$NN2}, the SD$^2$NN2 model still keeps its good performance for the Poisson-Boltzmann equation in 3D perforated domain.

	\section{Conclusion}\label{sec:05}
	
	Deep learning algorithms have demonstrated great potential in scientific computing tasks. However, the efficient solution of multi-scale problems remains to be a big challenge for DNN based numerical methods. In this paper, we combine subspace decomposition ideas from traditional numerical analysis and multi-scale deep neural network, and propose the \emph{SD$^2$NN} method for multi-scale PDEs. This new architecture consists of one low-frequency or normal DNN submodule and one (or several) high-frequency MscaleDNN submodule(s). In addition, we incorporate the trigonometric \emph{SFM} activation function for the SD$^2$NN model. Computational results show that this new method is feasible and efficient for multi-scale problems in 1d, 2d and 3d, and in both regular or perforated domains. In the future, we plan to apply this method to multi-scale PDEs with nonlinearity and/or randomness, as well as operator learning for multi-scale PDEs. 
	
	\section*{Acknowledgements}
	X.A.L and L.Z are partially supported by the National Natural Science Foundation of China (NSFC 11871339, 11861131004). Z.X. is supported by the National Key R\&D Program of China  Grant No. 2019YFA0709503, the Shanghai Sailing Program, the Natural Science Foundation of Shanghai Grant No. 20ZR1429000, the National Natural Science Foundation of China Grant No. 62002221, Shanghai Municipal of Science and Technology Project Grant No. 20JC1419500, Shanghai Municipal of Science and Technology Major Project NO. 2021SHZDZX0102, and the HPC of School of Mathematical Sciences and the Student Innovation Center at Shanghai Jiao Tong University. Thanks to Xin-Liang Liu for his some codes and helpful suggestions.

	\section*{Appendix}\label{appendix2mscale}
	\begin{appendices}
		\section{}\label{appendixA}
		\begin{table}[H]
			\centering
			\caption{ Network sizes and number of parameters for  models in Example \ref{DiffusionEq_1d_01}}
			\label{Network_size2DiffusionEq_1d}
			\begin{tabular}{|l|c|c|}
				\hline
				&Network Size                 &Number of Parameters    \\  \hline
				DNN             &(250, 100, 80, 80, 60)       &44510         \\  \hline
				Mscale          &(250, 100, 80, 80, 60)       &44510  \\  \hline
				WWP             &(125, 100, 80, 80, 60)       &44385  \\  \hline
				\multirow{2}*{SD$^2$NN1}&coarse:(100, 80, 60, 60, 40) &\multirow{2}*{44440} \\
				&fine:(125, 60, 60, 60, 50)   &                     \\  \hline
				\multirow{2}*{SD$^2$NN2}&coarse:(50, 80, 60, 60, 40)  &\multirow{2}*{44315} \\
				&fine:(100, 60, 60, 50, 40)   &                     \\  \hline
				\multirow{2}*{SD$^2$NN3}&coarse:(50, 80, 60, 60, 40)  &\multirow{2}*{44315} \\
				&fine:(100, 60, 60, 50, 40)   &                     \\
				\hline
			\end{tabular}
		\end{table}
		
		\begin{table}[H]
			\centering
			\caption{ Network sizes and number of parameters for  models in Example \ref{Diffusion1D3scale}}
			\label{Network_size2Diffusion_3scale}
			\begin{tabular}{|l|c|c|c|}
				\hline
				&Network Size                     & Para.                 &  Scaling factors\\  \hline
				Mscale          &(350, 300, 200, 200, 100)        & 225450                &    ------\\  \hline
				WWP             &(175, 300, 200, 200, 100)        & 225275                &    ------\\  \hline
				\multirow{3}*{SD$^2$NN1}&coarse:(100, 80, 60, 60, 40)     & \multirow{3}*{207730} &    ------\\
				& mesoscale:(125, 80, 60, 60, 40) &                       & $(30,31,32, \cdots, 69, 70)$              \\
				&fine:(225, 200, 150, 150, 100)   &                       & $(251, 252, 253, \cdots, 360)$       \\  \hline
				\multirow{3}*{SD$^2$NN2}&coarse:(50, 80, 60, 60, 40)      & \multirow{3}*{207680} & $(0.5, 1, 1.5, \cdots,24.5, 25)$     \\
				& mesoscale:(125, 80, 60, 60, 40) &                       & $(30,31,32, \cdots, 69, 70)$            \\
				&fine:(225, 200, 150, 150, 100)   &                       & $(251, 252, 253, \cdots, 360)$       \\  \hline
				\multirow{3}*{SD$^2$NN3}&coarse:(50, 80, 60, 60, 40)      & \multirow{3}*{207680} & $(0.5, 1, 1.5, \cdots,24.5, 25)$    \\
				& mesoscale:(125, 80, 60, 60, 40) &                       & $(30,31,32, \cdots, 69, 70)$   \\
				&fine:(225, 200, 150, 150, 100)   &                       & $(251, 252, 253, \cdots, 360)$       \\  \hline
			\end{tabular}
		\end{table}
		
		\begin{table}[H]
			\centering
			\caption{ Network sizes and parameter of  models for Example \ref{Diffusion2D}}
			\label{Network_size2DiffusionEq_2d}
			\begin{tabular}{|l|c|c|}
				\hline
				&Network Size                     & Para.        \\  \hline
				Mscale          &(250, 200, 200, 100, 100, 80)      & 128330      \\  \hline
				WWP             &(125, 200, 200, 100, 100, 80)      & 128205      \\  \hline
				\multirow{2}*{SD$^2$NN2}&coarse:(50, 100, 80, 80, 80, 60)   & \multirow{2}*{127410}      \\
				&fine:(120, 150, 150, 100, 100, 80) &        \\  \hline
			\end{tabular}
		\end{table}
		
		\begin{table}[H]
			\centering
			\caption{ Network sizes and parameter of different models for Example \ref{Boltzmann3D}}
			\label{Network_size2Boltzmann_3d}
			\begin{tabular}{|l|c|c|}
				\hline
				&Network Size                     & Para.        \\  \hline
				Mscale          &(500, 400, 400, 200, 200, 150)      & 510650      \\  \hline
				WWP             &(250, 400, 400, 200, 200, 150)      & 510400      \\  \hline
				\multirow{2}*{SD$^2$NN2}&coarse:(70, 200, 200, 150, 150, 150)   & \multirow{2}*{508870}      \\
				&fine:(250, 300, 290, 200, 200, 150) &        \\  \hline
			\end{tabular}
		\end{table}
	\end{appendices}
	
\end{document}